\RequirePackage{fix-cm}
\documentclass{article}

\PassOptionsToPackage{numbers,sort&compress}{natbib}
\usepackage[main,final]{neurips_2026}
\renewcommand{\acksection}{\section*{Acknowledgments}}
\setcitestyle{numbers,square,comma}

\usepackage[utf8]{inputenc}
\usepackage[T1]{fontenc}
\usepackage{hyperref}
\usepackage{url}
\usepackage{amsmath}
\usepackage{amsfonts}
\usepackage{amssymb}
\usepackage{graphicx}
\usepackage{wrapfig}
\usepackage{algorithm}
\usepackage{algpseudocode}
\usepackage{float}
\usepackage{array}
\usepackage{multirow}
\usepackage{threeparttable}
\usepackage{nicefrac}
\usepackage{microtype}
\usepackage[table]{xcolor}
\usepackage{tikz}

\usepackage{booktabs}
\usepackage{hyperref}
\usepackage{siunitx}
\usepackage[table]{xcolor} 
\usepackage[most]{tcolorbox}
\usepackage{placeins}
\usepackage{subcaption}

\newcommand{\cmark}{\textcolor{green!70!black}{\ding{51}}}
\newcommand{\xmark}{\textcolor{red}{\ding{55}}}
\definecolor{DegBest}{RGB}{0, 100, 0} 
\definecolor{DegMid}{RGB}{180, 120, 0}
\definecolor{DegWorst}{RGB}{150, 0, 0}

\newcommand{\myparagraph}[1]{\noindent\textbf{#1}}

\usepackage{multirow} 
\usepackage{pifont} 

\hypersetup{colorlinks=true,linkcolor=black,citecolor=black,urlcolor=blue}

\providecommand{\Description}[1]{}

\title{DTA-GT: Direction- and Topology-Aware Graph Transformer for Neural Network Representation Learning%
}

\author{%
  \textbf{Yuxiang Zeng}$^{\dagger}$, \quad \textbf{Kun Xie}$^{\dagger,*}$, \quad \textbf{Yang Wang}$^{\dagger}$, \quad \textbf{Jigang Wen}$^{\ddagger}$, \quad \textbf{Xiaocan Li}$^{\dagger}$, \\
  \textbf{Guangxing Zhang}$^{\S}$, \quad \textbf{Gaogang Xie}$^{\S}$, \quad \textbf{Jiannong Cao}$^{\P}$ \\[0.2em]
  {\normalfont$^{\dagger}$Hunan University \quad $^{\ddagger}$Hunan University of Science and Technology} \\
  {\normalfont$^{\S}$Chinese Academy of Sciences \quad $^{\P}$The Hong Kong Polytechnic University} \\[0.2em]
  {\normalfont\texttt{\{{\hypersetup{urlcolor=black}\href{mailto:zengyuxiang@hnu.edu.cn}{zengyuxiang@hnu.edu.cn}}, xiekun@hnu.edu.cn\}}}
}

\begin{document}

\raggedbottom

\setcounter{footnote}{0}
\maketitle
\begingroup
\renewcommand{\footnotesize}{\fontsize{6}{7}\selectfont}
\setlength{\footnotesep}{6pt}
\makeatletter
\long\def\@makefntext#1{\parindent 1em\noindent\hbox to 1.8em{\hss $\m@th ^{\@thefnmark}$}#1}
\makeatother
\begingroup
\renewcommand{\thefootnote}{\ensuremath{*}}
\footnotetext{Kun Xie is the corresponding author.}
\endgroup
\begingroup
\renewcommand{\thefootnote}{\ensuremath{\dagger}}
\footnotetext{\parbox[t]{\dimexpr\linewidth-1.8em\relax}{%
Hunan University, 410082, Changsha, China\\
College of Computer Science and Electronic Engineering\\
Ministry of Education Key Laboratory for Supercomputing and Artificial Intelligence Converged Computing%
}}
\endgroup
\begingroup
\renewcommand{\thefootnote}{}
\footnotetext{Preprint; subject to revision. The camera-ready version will be authoritative.}
\endgroup
\endgroup
\setcounter{footnote}{0}

\begin{abstract}
  Architecture attribute predictors reduce the cost of neural architecture search by estimating accuracy or latency from a small set of evaluated candidates.
  Neural architectures are operation-labeled directed acyclic graphs (DAGs), where edge directions determine computational roles and execution dependencies.
  Existing graph Transformer predictors incorporate topological and directional information in different ways, but primarily focus on specific mechanisms rather than coordinated encoder-wide modeling of these semantics.
  We introduce \textbf{encoder-wide directional consistency}, a design principle that coordinates the modeling of directed computational semantics across node initialization, global spectral encoding, pairwise interaction, and feature updates.
  To implement this principle, we propose DTA-GT, a Direction- and Topology-Aware Graph Transformer that integrates direction- and topology-aware node initialization, magnetic Laplacian spectral encoding, direction-dual structural attention, and a Direction-Sensitive MoE-FFN.
  Across accuracy and latency prediction on NAS-Bench-101 and NAS-Bench-201, DTA-GT consistently outperforms representative RNN-, CNN-, GNN-, Transformer-, and hybrid predictors under limited-budget settings.
  Ablations and magnetic spectral controls support the contribution of the direction-aware stages and orientation-sensitive spectral information, while mechanistic diagnostics characterize module behavior.
  These results support encoder-wide directional consistency as an effective design principle for architecture DAG representation.

  \noindent\textbf{Code:} \url{https://github.com/ZengYuXiang7/DTA-GT}
  \end{abstract}


\section{Introduction}\label{Introduction}

Deep learning has been widely applied across various fields~\citep{huang2026tools,wang2026libra,Zhang2025_UniGRF,10234360,s23156775,ye2025cross,liu2026palm,shen2026fine}.
Neural architecture design shapes a network's computational semantics and execution dependencies, influencing both task performance and execution efficiency.
Neural Architecture Search (NAS) automates this design process by searching a candidate architecture space for models that meet performance and deployment constraints~\citep{cai2018pathlevel,liu2018pnas,pham2018enas,cai2019proxylessnas,tan2019mnasnet,wu2019fbnet}.
However, evaluating candidates during search is expensive: accuracy evaluation typically requires full network training, and latency evaluation requires hardware measurement or profiling.
A practical way to reduce this cost is to train predictors on a small subset of evaluated architectures and use them to estimate accuracy or latency for unseen candidates~\citep{dudziak2020brpnas,yi2023narformer,yi2023narformerv2,xu2025nnformer}.

\noindent\begin{minipage}{\linewidth}
\setlength{\intextsep}{0pt}
\begin{wrapfigure}[14]{r}{0.46\linewidth}
\centering
\captionsetup{font=footnotesize,skip=3pt,justification=raggedright,singlelinecheck=false,hypcap=true}
\includegraphics[width=\linewidth]{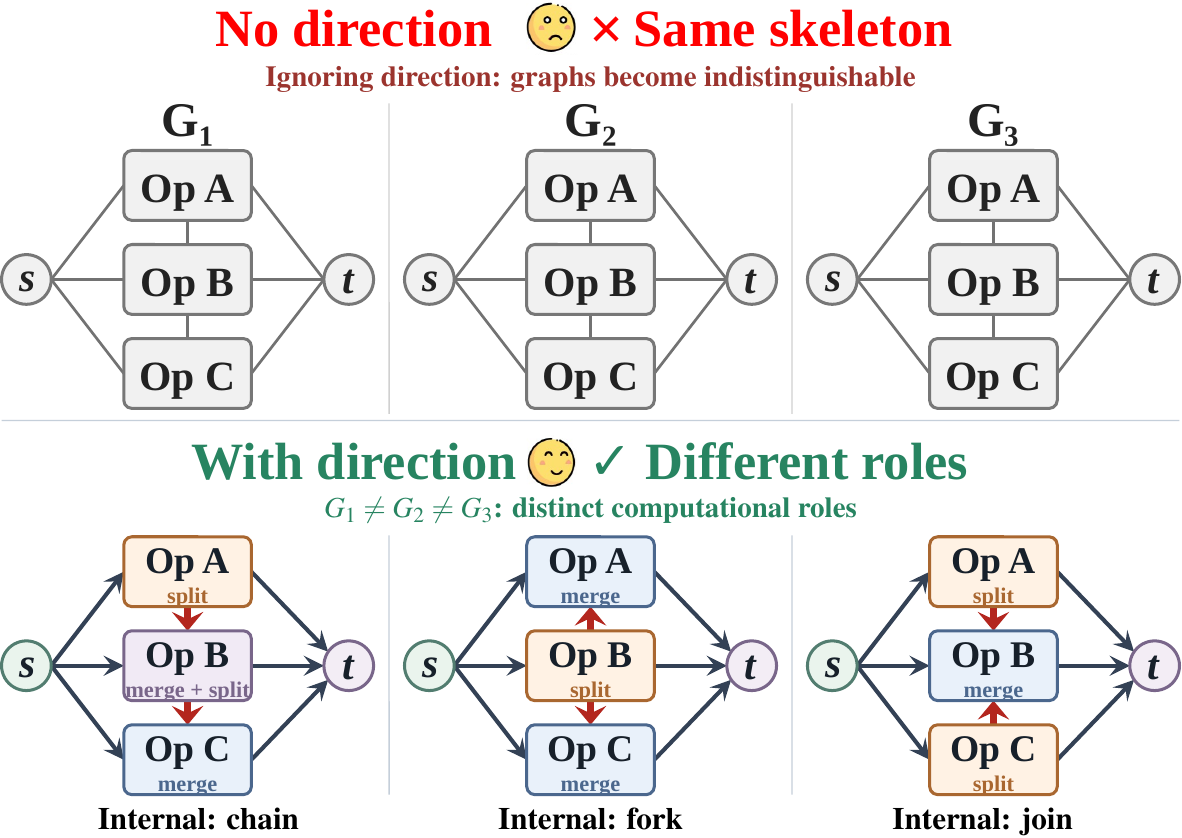}
\caption{Edge directions distinguish operation order and split/merge roles.}
\label{fig:motivation}
\end{wrapfigure}
\noindent
Learning-based predictors rely on architecture representations that capture how operations jointly process information.
Each candidate architecture is represented as a directed acyclic graph (DAG), where nodes represent operations and an edge $u \rightarrow v$ indicates that the output of operation $u$ feeds into operation $v$.
Directed paths extend these dependencies beyond adjacent operations.
The three architectures in Fig.~\ref{fig:motivation} share the same operations and undirected skeleton, but their internal edge directions organize computation differently.
In $G_1$, the internal path follows $A \rightarrow B \rightarrow C$; in $G_2$, $B$ feeds both $A$ and $C$; and in $G_3$, $A$ and $C$ feed into $B$.
These patterns can affect a network's accuracy by changing how features are transformed and combined to form semantic representations.
They can also influence execution latency by shaping computational dependencies: $G_1$ imposes a serial internal path; $G_2$ and $G_3$ permit potential parallelism after and before $B$, respectively, with $B$ waiting for both branches in $G_3$.
These connections provide a structural basis for predicting accuracy and latency from architecture graphs.
Discarding edge directions reduces the three architectures to the same undirected graph, obscuring their different feature flows and execution dependencies.
An effective architecture representation should retain these directed computational semantics across local connections and multi-hop paths.
\end{minipage}

GNN-based predictors exploit graph topology through local message passing and can incorporate edge directions to model dependencies between neighboring operations~\citep{dudziak2020brpnas,wen2020neuralpredictor,ning2020gates,liu2022nnlqp,wei2023npenas}.
However, modeling long-range dependencies often requires repeated message passing, which can lead to over-squashing and, with increasing depth, over-smoothing~\citep{alon2021bottleneck,oono2020graph}.
Pure Transformers enable direct interactions between distant graph elements through global self-attention~\citep{kim2022pure}.
Nevertheless, vanilla self-attention lacks inherent inductive biases for graph topology and edge direction, motivating the introduction of graph-specific structural encodings~\citep{ying2021graphormer}.
In neural architecture prediction, Transformer-based and hybrid methods integrate structural and directional information at different levels of encoder design through mechanisms such as topology-aware encodings, structural attention, and bidirectional aggregation~\citep{lu2021tnasp,lu2023pinat,xu2025nnformer,yi2023narformer,yi2023narformerv2} (see Table~\ref{tab:method_positioning}).
Despite these advances, the broader challenge is how to coordinate these mechanisms so that directed computational semantics are modeled consistently throughout the encoder.
In architecture DAGs, edge direction determines operation roles, multi-hop dependencies, and distinct upstream and downstream contexts.
Because these aspects arise from the same directed computation, a shared design principle should guide how they are modeled during representation construction, interaction, and feature updates.
\begin{table}[t]
\centering
\fontsize{7.0}{7.7}\selectfont
\captionsetup{justification=centering}
\definecolor{RelBest}{RGB}{0,145,70}
\definecolor{RelMid}{RGB}{215,135,0}
\definecolor{RelWorst}{RGB}{205,0,0}
\providecommand{\relstatusfull}[1]{\makebox[1.0em][c]{\smash{\tikz[baseline=-0.67ex]\fill[#1] (0,0) circle (0.74ex);}}}
\providecommand{\relstatushalf}[1]{\makebox[1.0em][c]{\smash{\tikz[baseline=-0.67ex]{\begin{scope}\clip (0,0) circle (0.74ex);\fill[#1] (-0.74ex,-0.74ex) rectangle (0,0.74ex);\end{scope}\draw[#1,line width=0.18ex] (0,0) circle (0.74ex);}}}}
\providecommand{\relstatusopen}[1]{\makebox[1.0em][c]{\smash{\tikz[baseline=-0.67ex]\draw[#1,line width=0.18ex] (0,0) circle (0.74ex);}}}
\providecommand{\relfullcircle}{\relstatusfull{RelBest}}
\providecommand{\relhalfcircle}{\relstatushalf{RelMid}}
\providecommand{\relemptycircle}{\relstatusopen{RelWorst}}
\providecommand{\relhead}[1]{{\centering\fontsize{7.0}{7.4}\selectfont\textbf{#1}\par}}
\makeatletter
\providecommand{\relnumcite}[1]{{\setcitestyle{numbers,square,comma}\def\NAT@sort{\@ne}\citep{#1}}}
\makeatother
\caption{Topology and direction modeling in representative architecture predictors.}
\label{tab:method_positioning}
\setlength{\tabcolsep}{1.5pt}
\renewcommand{\arraystretch}{1.05}
\begin{tabular}{%
@{}>{\centering\arraybackslash}m{0.150\textwidth}
>{\centering\arraybackslash}m{0.110\textwidth}
>{\centering\arraybackslash}m{0.270\textwidth}
>{\centering\arraybackslash}m{0.100\textwidth}
>{\centering\arraybackslash}m{0.100\textwidth}
>{\centering\arraybackslash}m{0.105\textwidth}
>{\centering\arraybackslash}m{0.100\textwidth}@{}}
\toprule
\relhead{Category} &
\relhead{Representative\\Methods} &
\relhead{Modeling\\Paradigm} &
\relhead{Topology \& Direction Encoding} &
\relhead{Directional Features / Relations} &
\relhead{Global Directional Spectral Prior} &
\relhead{Direction-Specific Updates} \\ \midrule

\textbf{\mbox{CNN-based}} &
\relnumcite{xu2021renas} &
Convolutional tensor encoding &
\relhalfcircle &
\relemptycircle &
\relemptycircle &
\relemptycircle \\ \addlinespace[1.5pt]

\textbf{\mbox{RNN-based}} &
\relnumcite{deng2017peephole,luo2018nao,luo2020seminas} &
Recurrent sequence encoding &
\relhalfcircle &
\relemptycircle &
\relemptycircle &
\relemptycircle \\ \addlinespace[1.5pt]

\textbf{\mbox{GNN-based}} &
\relnumcite{dudziak2020brpnas,wen2020neuralpredictor,ning2020gates,liu2022nnlqp} &
GNN-based message passing &
\relhalfcircle &
\relhalfcircle &
\relemptycircle &
\relhalfcircle \\ \addlinespace[1.5pt]

\textbf{\mbox{Transformer-based}} &
\relnumcite{lu2021tnasp,yi2023narformer,lu2023pinat} &
Topology-aware token-attention &
\relhalfcircle &
\relhalfcircle &
\relemptycircle &
\relemptycircle \\ \addlinespace[1.5pt]

\textbf{Hybrid} &
\relnumcite{yi2023narformerv2,xu2025nnformer} &
GNN--Transformer hybrid &
\relhalfcircle &
\relhalfcircle &
\relemptycircle &
\relhalfcircle \\ \midrule

\rowcolor{gray!15}
\textbf{Proposed} &
\textbf{DTA-GT} &
\textbf{Direction- and Topology-Aware Graph Transformer} &
\relfullcircle &
\relfullcircle &
\relfullcircle &
\relfullcircle \\ \bottomrule
\end{tabular}
\par\smallskip
\begin{minipage}{\linewidth}
\centering
\fontsize{6.7}{7.3}\selectfont
\textit{Note:} Symbols summarize the cited representative designs: \relfullcircle{} denotes explicit coverage, \relhalfcircle{} denotes partial or varying coverage, and \relemptycircle{} denotes no explicit coverage.
The comparison is qualitative and concerns representation design rather than predictive performance.
\end{minipage}
\vspace{-2.7em}
\end{table}

Guided by this intuition mentioned above, we ask a central question:
\begingroup
\setlength{\parskip}{0pt}
\setlength{\topsep}{0pt}
\setlength{\partopsep}{0pt}
\begin{center}
\begin{tcolorbox}[
    width=0.94\linewidth,
    colback=white,
    colframe=black!40,
    boxrule=0.5pt,
    arc=1mm,
    left=5pt,
    right=5pt,
    top=6pt,
    bottom=6pt,
    before skip=2pt,
    after skip=2pt,
]
\centering\emph{How can a neural architecture encoder capture DAG topology while consistently modeling the computational semantics of edge direction?}
\end{tcolorbox}
\end{center}
\endgroup

Our key insight is that directionality in architecture DAGs is more than an auxiliary structural cue: it carries computational semantics that should guide representation learning throughout the encoder.
Building on this insight, we propose \textbf{encoder-wide directional consistency} as a unifying design principle that requires node representations, global structural encoding, pairwise interactions, and feature updates to explicitly account for these computational semantics.

Guided by this principle, we develop DTA-GT, a Direction- and Topology-Aware Graph Transformer.
Direction- and topology-aware node initialization captures directed node roles, and magnetic Laplacian spectral encoding provides a global prior sensitive to edge orientation.
Direction-dual structural attention models pairwise dependencies through separate directional relation channels, and a Direction-Sensitive MoE-FFN adaptively integrates self, predecessor, and successor contexts during feature updates.
We evaluate DTA-GT on four prediction settings spanning accuracy and latency prediction over NAS-Bench-101 and NAS-Bench-201. 
Across multiple limited-budget settings and evaluation metrics, DTA-GT consistently improves over representative methods.
Stage-wise ablations assess the contribution of the coordinated direction-aware stages.
Magnetic spectral controls isolate the effect of the spectral prior sensitive to edge orientation.
Mechanistic diagnostics examine whether the internal behaviors align with the intended directed computational semantics.
Our main contributions are:
\begin{itemize}
    \item We propose \textbf{encoder-wide directional consistency} as a unifying design principle that coordinates the modeling of directed computational semantics across the main stages of an architecture encoder.

    \item We introduce DTA-GT, which integrates direction- and topology-aware node initialization, magnetic Laplacian spectral encoding, direction-dual structural attention, and a Direction-Sensitive MoE-FFN.

    \item We conduct a unified evaluation on accuracy and latency prediction over standard NAS benchmarks, with stage-wise ablations, magnetic spectral controls, and mechanistic diagnostics to analyze the contribution and behavior of the proposed direction-aware stages.
\end{itemize}

\section{Related Work}\label{RelatedWork}

Existing architecture predictors include CNN-based, RNN-based, GNN-based, Transformer-based, and GNN--Transformer hybrid models.
These families differ in how they represent topology and use edge directions to model dependencies between operations (see Table~\ref{tab:method_positioning}).

\myparagraph{CNN- and RNN-based methods.}
ReNAS~\citep{xu2021renas} uses a CNN predictor with handcrafted architecture features.
Peephole~\citep{deng2017peephole} and NAO~\citep{luo2018nao} use LSTM encoders to map architecture sequences to vector representations.
SemiNAS~\citep{luo2020seminas} also uses an LSTM encoder and incorporates unlabeled architectures through semi-supervised training.
Nevertheless, these predictors process architectures as tensors or sequences; retaining connectivity information in such inputs does not by itself ensure that graph topology and edge direction are jointly modeled throughout the encoder.

\myparagraph{GNN-based methods.}
BRP-NAS~\citep{dudziak2020brpnas}, NP~\citep{wen2020neuralpredictor}, NPENAS~\citep{wei2023npenas}, GATES~\citep{ning2020gates}, and DNNPerf~\citep{gao2023dnnperf} model neural architectures as graphs.
NNLP~\citep{liu2022nnlqp}, the latency prediction component of the NNLQP system, uses graph-based representations to estimate neural network latency.
Message passing lets these predictors use graph topology directly and can distinguish incoming from outgoing edges within a local neighborhood.
With local message passing, capturing long-range dependencies requires repeated propagation, which can lead to over-squashing and, with increasing depth, over-smoothing~\citep{alon2021bottleneck,oono2020graph}.

\myparagraph{Transformer-based and hybrid methods.}
NAR-Former~\citep{yi2023narformer} encodes operations and their connections as a sequence of tokens.
TNASP~\citep{lu2021tnasp} and PINAT~\citep{lu2023pinat} incorporate graph topology through positional encodings based on Laplacian matrices.
Among hybrid predictors, NAR-Former V2~\citep{yi2023narformerv2} uses adjacency to restrict which nodes attend to one another.
NN-Former~\citep{xu2025nnformer} incorporates directional graph relations into attention and feature aggregation.
However, the Transformer and hybrid predictors summarized in Table~\ref{tab:method_positioning} provide only partial coverage of directional modeling across encoder stages.
Directional cues introduced at one stage do not by themselves ensure that their computational meaning guides the other stages.
The broader design challenge is to coordinate the modeling of operation roles, multi-hop dependencies, and upstream and downstream contexts across representation construction, interaction, and update.
In contrast, guided by our proposed encoder-wide directional consistency, DTA-GT makes directed computational semantics an explicit design requirement across node representations, global structural encoding, pairwise interactions, and feature updates.

\section{Problem Formulation and Model Overview}\label{ProblemFormulation}

    \subsection{Problem Formulation}
    Following established conventions~\citep{liu2022nnlqp,lu2021tnasp,yi2023narformer,yi2023narformerv2,xu2025nnformer}, let $\mathcal{A}$ be the candidate architecture space.
    Each architecture $a\in\mathcal{A}$ is represented as an operation-labeled directed acyclic graph (DAG) $G=(V,E,\tau)$, where each node represents a computational operation, such as a convolutional layer or a pooling layer, directed edges represent data flow between operations, and $\tau(v)$ denotes the operation type of node $v$.
    We adopt the adjacency convention that $\mathbf{A}_{u,v}=1$ indicates a directed edge $u\to v$.

    Each architecture has a scalar target attribute $y\in\mathbb{R}$, such as accuracy or latency.
    Given a measured subset $\mathcal{D}=\{(a_i,y_i)\}_{i=1}^N$ sampled from $\mathcal{A}$, we learn a predictor $f_\vartheta:\mathcal{A}\to\mathbb{R}$ to estimate attributes for remaining unmeasured candidates in the same architecture space, without fully training or profiling each candidate. We decompose $f_\vartheta$ into an encoder $h_\phi$ and a prediction head $p_\psi$: $\mathbf{z}_G=h_\phi(G)$ and $\hat{y}=p_\psi(\mathbf{z}_G)$, with $\vartheta=\{\phi,\psi\}$.
    Our goal is to learn architecture representations that capture operation semantics and DAG topology while consistently modeling directed computational semantics across the encoder.

\subsection{Overall Architecture}\label{sec:overall}
\begin{figure}[t]
    \centering
    \includegraphics[width=\linewidth]{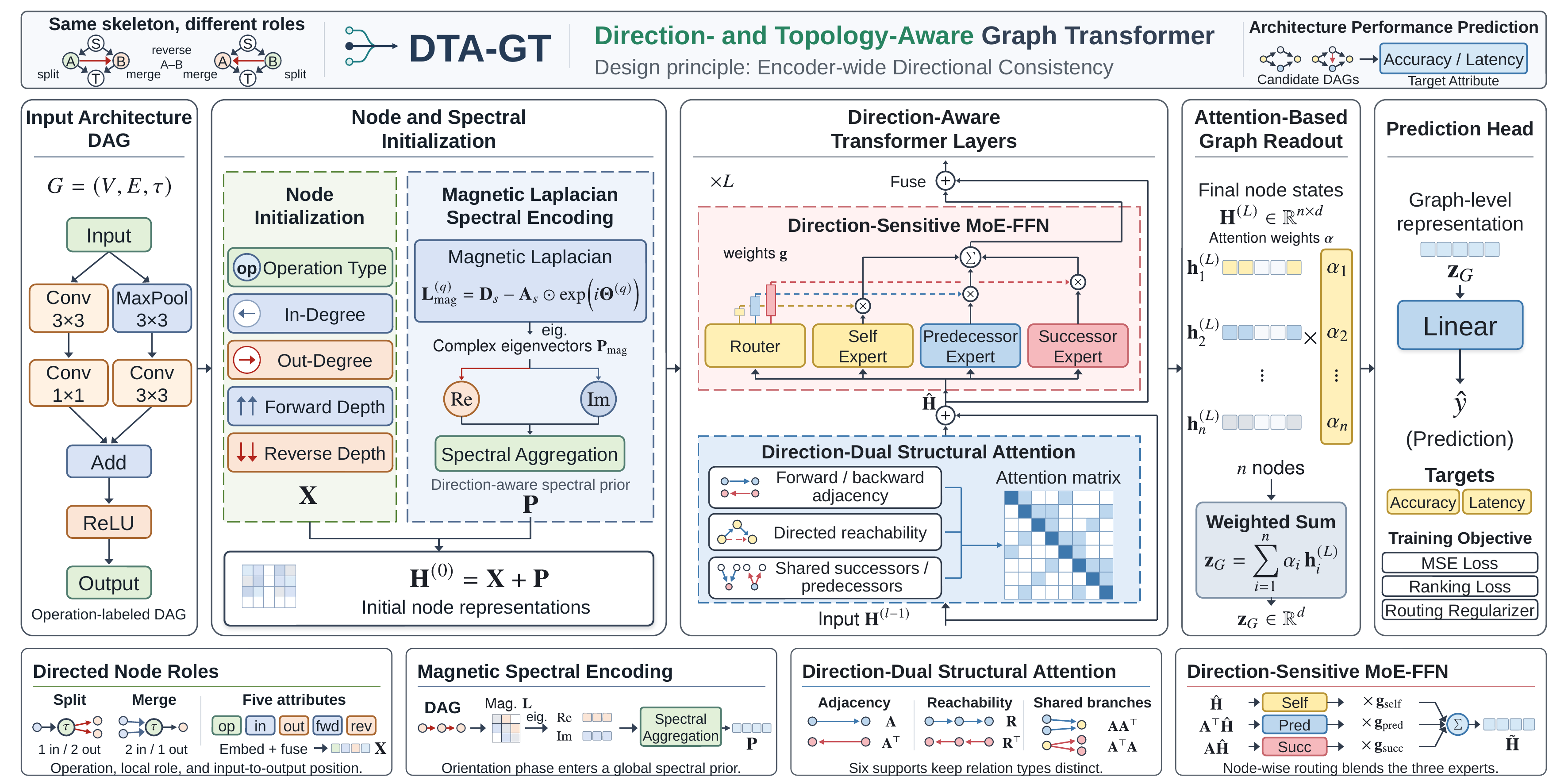}
    \caption{The DTA-GT framework.}
    \Description{Overview of the DTA-GT encoder architecture, including node embedding, direction-aware spectral encoding, structural attention, direction-sensitive expert updates, graph readout, and prediction.}
    \label{fig:framework}
    \vspace{-16pt}
\end{figure}

    Fig.~\ref{fig:framework} illustrates DTA-GT, which maps an architecture DAG $G$ to a graph-level representation $\mathbf{z}_G=h_\phi(G)$ under the encoder-wide directional consistency principle.
    Specifically, the encoder first initializes node states as $\mathbf{H}^{(0)}=\mathrm{NodeEmbed}(G)+\mathrm{DirSpecEnc}(G)$, combining operation/local directed-role features with a global spectral prior sensitive to edge orientation.
    It then applies $L$ residual layers~\citep{he2016resnet}, each consisting of direction-dual structural attention followed by a direction-sensitive MoE-FFN, using $\mathbf{R}_G$ and $\mathbf{A}$ for relation-aware interaction and predecessor/successor-aware updates, respectively.
    Finally, attention-based readout produces $\mathbf{z}_G=\mathrm{AttnReadout}(\mathbf{H}^{(L)})$, which is passed to $p_\psi$ for prediction without altering the original architecture DAG.

\section{Method}\label{Method}
    \subsection{Direction- and Topology-Aware Node Initialization}\label{sec:node_init}

    Before global spectral encoding and attention-based interaction, each node should carry a basic description of its computation and its directed role in the architecture DAG.
    Operation labels specify the computation at each node, but they do not distinguish nodes with the same operation type that appear in different local structures or input-to-output positions.
    Topological-order indices can provide positional cues, but they may introduce ordering artifacts because a DAG can admit multiple valid topological orders~\citep{yi2023narformer,yi2023narformerv2,lu2023pinat}.
    Consequently, we initialize nodes with topological-order-invariant attributes that encode computation, local directed roles, and input-to-output positions.

    \myparagraph{Role-aware attributes.}
    Each node $v_i$ is described by five discrete features: operation label $\tau(v_i)$, indegree $\deg^{-}(v_i)$, outdegree $\deg^{+}(v_i)$, forward depth $\ell_\text{fwd}(v_i)$, and reverse depth $\ell_\text{rev}(v_i)$.
    The forward depth is the maximum length of a directed path from an input node to $v_i$, and the reverse depth is the maximum length of a directed path from $v_i$ to an output node.
    The operation label captures computational semantics, indegree and outdegree characterize local merging and branching roles, and $(\ell_\text{fwd},\ell_\text{rev})$ locates the node along the directed input-to-output flow.
    For example, two nodes can have the same forward depth but different longest path lengths to an output.
    Forward depth alone cannot distinguish these positions, so we also include reverse depth.

    So, we embed, concatenate, and project them intuitively:
    \begin{equation}
    \mathbf{x}_i =
    \big[
    \mathbf{e}_{\tau(v_i)}\mathbf{W}_\text{op}
    \;\|\;
    \mathbf{e}_{\deg^{-}(v_i)}\mathbf{W}_\text{in}
    \;\|\;
    \mathbf{e}_{\deg^{+}(v_i)}\mathbf{W}_\text{out}
    \;\|\;
    \mathbf{e}_{\ell_\text{fwd}(v_i)}\mathbf{W}_\text{fwd}
    \;\|\;
    \mathbf{e}_{\ell_\text{rev}(v_i)}\mathbf{W}_\text{rev}
    \big]\mathbf{W}_\text{fuse},
    \end{equation}
    where $\mathbf{e}_{(\cdot)}$ denotes a one-hot vector, the $\mathbf{W}$ terms are learnable projections, $\|$ denotes concatenation, and stacking all $\mathbf{x}_i$ gives $\mathbf{X}\in\mathbb{R}^{n\times d}$.

    \subsection{Magnetic Laplacian Spectral Encoding}\label{sec:maglap}

The node initialization $\mathbf{X}$ provides operation semantics and local directed roles, but not a global direction-aware spectral prior over the architecture DAG.
Existing predictors usually lack an explicit global direction-aware spectral prior~\citep{dudziak2020brpnas,wen2020neuralpredictor,ning2020gates,liu2022nnlqp,lu2021tnasp,yi2023narformer,lu2023pinat,yi2023narformerv2,xu2025nnformer}, while standard Laplacian encodings capture symmetric connectivity but not the asymmetry between $u\to v$ and $v\to u$.
Inspired by magnetic Laplacian positional encoding~\citep{geisler2023transformers}, we derive $\mathbf{P}$ from the magnetic Laplacian~\citep{furutani2019hermitian,zhang2021magnet}, whose complex phase records edge orientation.

\myparagraph{Construction.}
Given the directed adjacency matrix $\mathbf{A}$, define the symmetrized support $(\mathbf{A}_s)_{u,v}=\frac{1}{2}(\mathbf{A}_{u,v}+\mathbf{A}_{v,u})$, the orientation phase $(\boldsymbol{\Theta}^{(q)})_{u,v}=2\pi q(\mathbf{A}_{u,v}-\mathbf{A}_{v,u})$, and the degree $(\mathbf{D}_s)_{u,u}=\sum\nolimits_{v=1}^{n}(\mathbf{A}_s)_{u,v}$.
The magnetic Laplacian is
\begin{equation}\label{eq:maglap}
\mathbf{L}_{\mathrm{mag}}^{(q)}
= \mathbf{D}_s-\mathbf{A}_s\odot \exp\left(i\boldsymbol{\Theta}^{(q)}\right).
\end{equation}
Since $\mathbf{A}_s$ is symmetric and $\boldsymbol{\Theta}^{(q)}$ is antisymmetric, the magnetic Laplacian is Hermitian.
We derive a direction-aware spectral encoding from the magnetic Laplacian.

\myparagraph{Two-channel spectral features.}
We form the retained eigenvector matrix $\mathbf{P}_{\mathrm{mag}}=[\boldsymbol{\gamma}_1,\ldots,\boldsymbol{\gamma}_k]\in\mathbb{C}^{n\times k}$ from the $k$ low-frequency eigenvectors.
For each node $u$ and spectral component $r$, we separate its real and imaginary parts as $\mathbf{E}_{u,r}=[\operatorname{Re}(\mathbf{P}_{\mathrm{mag}}(u,r))\Vert\operatorname{Im}(\mathbf{P}_{\mathrm{mag}}(u,r))]$, giving $\mathbf{E}\in\mathbb{R}^{n\times k\times 2}$.
The real and imaginary channels provide a real-valued representation of the complex spectral signal, preserving orientation-dependent phase information.

\myparagraph{Spectral aggregation.}
For each node, $\operatorname{SpectralAgg}$ embeds its two-channel spectral components and applies self-attention across the $k$ spectral modes.
The outputs are concatenated and projected to $d$ dimensions, producing the learned node encoding $\mathbf{p}_u$.
Stacking these encodings gives $\mathbf{P}\in\mathbb{R}^{n\times d}$.
The detailed formulation of spectral aggregation is provided in Appendix~\ref{app:maglap_aggregation}.
The initial encoder states are then obtained by adding this global direction-aware spectral encoding to the node initialization:
\begin{equation}\label{eq:init_with_mag}
\mathbf{P}=\operatorname{SpectralAgg}(\mathbf{E}),\qquad \mathbf{H}^{(0)}=\mathbf{X}+\mathbf{P}.
\end{equation}

    \subsection{Direction-Dual Structural Attention}\label{sec:struct_attn}

Standard self-attention~\citep{vaswani2017attention,yi2023narformer} evaluates pairwise interactions from node representations.
The magnetic spectral encoding in Sec.~\ref{sec:maglap} supplies a node-level global direction-aware spectral prior.
Direction-dual structural attention complements this representation with explicit pairwise relation learning, implementing the interaction stage of encoder-wide directional consistency.
It organizes pairwise interactions through six directional relation channels.
Each channel specifies a distinct structural context for query--key interactions, covering local dependencies, directed reachability, or shared-successor and shared-predecessor structures.
Within each channel, content-based attention adaptively weights the admissible keys according to their node representations.

\myparagraph{Direction-dual relation channels.}
We use paired attention channels to distinguish forward from backward dependencies, and to separate node pairs that receive input from the same upstream node from those that feed into the same downstream node.
We instantiate three paired relation types, yielding six directional relation channels: (1) \textbf{local adjacency}, where $\mathbf{A}$ and $\mathbf{A}^{\top}$ indicate direct forward and backward dependencies; (2) \textbf{directed reachability}, where $\mathbf{R}_{\mathrm{fwd}}(u,v)=1$ if $v$ is reachable from $u$, and $\mathbf{R}_{\mathrm{bwd}}=\mathbf{R}_{\mathrm{fwd}}^{\top}$ captures the reverse relation~\citep{luo2023transformers}; and (3) \textbf{shared successors/predecessors}, where $\mathbf{A}\mathbf{A}^{\top}$ captures common successors and $\mathbf{A}^{\top}\mathbf{A}$ captures common predecessors.
Let $\mathbf{R}_G=\{\mathbf{R}^{(r)}\}_{r=1}^{6}$ denote the resulting relation channels.
Together, these channels preserve local connectivity, multi-hop directed reachability, and higher-order branch structures as distinct structural contexts.

\myparagraph{Relation-constrained attention.}
For notational brevity, write the input node states of layer $l$ as $\mathbf{H}=\mathbf{H}^{(l-1)}\in\mathbb{R}^{n\times d}$, and let $d_{\mathrm{head}}=d/M$ for $M$ attention heads.
For the $m$-th head, we compute $\mathbf{Q}_m=\mathbf{H}\mathbf{W}_m^Q$, $\mathbf{K}_m=\mathbf{H}\mathbf{W}_m^K$, and $\mathbf{V}_m=\mathbf{H}\mathbf{W}_m^V$ using projection matrices $\mathbf{W}_m^Q,\mathbf{W}_m^K,\mathbf{W}_m^V\in\mathbb{R}^{d\times d_{\mathrm{head}}}$.
We use a fixed relation-to-head assignment: in the six-head configuration used in our experiments, head $m$ is associated with directional relation channel $\rho(m)=m$.
The assigned relation defines the structural support of that head: we set $\mathbf{B}_m(u,v)=0$ when $\mathbf{R}^{(\rho(m))}(u,v)>0$ or $u=v$, and $\mathbf{B}_m(u,v)=-\infty$ otherwise.
The attention scores and multi-head output are then computed as
\begin{equation}
\mathbf{S}_m=\frac{\mathbf{Q}_m\mathbf{K}_m^\top}{\sqrt{d_{\mathrm{head}}}}+\mathbf{B}_m,
\qquad
\mathrm{StructAttn}(\mathbf{H},\mathbf{R}_G)=
\left[\big\Vert_{m=1}^{M}\mathrm{softmax}(\mathbf{S}_m)\mathbf{V}_m\right]\mathbf{W}^O,
\end{equation}
where $\big\Vert$ denotes concatenation over attention heads, $\mathbf{W}^O\in\mathbb{R}^{Md_{\mathrm{head}}\times d}$, and the softmax is applied row-wise over key nodes.

In each attention head, the relation mask specifies which nodes a query node can attend to.
Within this set, the query--key scores determine the attention weights.
We combine the head outputs to incorporate information from different directional relations into each node representation.

We note that these three relation pairs are a heuristic choice motivated by DAG semantics.
The directional consistency principle can guide additional channel pairs that distinguish opposite directions or shared upstream and downstream contexts.
For example, dependencies at specific directed path lengths can define new pairs of forward and backward channels (Appendix~\ref{app:limitations-impact}).

    \subsection{Direction-Sensitive Mixture-of-Experts Feed-Forward Network}\label{sec:dir_ffn}

After structural attention, a standard position-wise FFN updates each node using a shared transformation.
However, the relative importance of upstream and downstream information can vary across nodes, and this transformation does not explicitly weight the two directional contexts separately.
We therefore replace the standard FFN with a Direction-Sensitive MoE-FFN, which uses three role-specific experts for the node's own state, predecessor context, and successor context.
A node-wise routing gate adaptively weights these experts according to the current node state.
This keeps the feature-update stage direction-aware without assigning a fixed expert to every node role, since the useful predecessor/successor balance can vary across architectures and prediction targets.

\myparagraph{Directional experts.}
Given the post-attention states $\hat{\mathbf{H}}^{(l)}$, the three expert outputs are
\begin{equation}
\mathbf{H}_\text{self}
=
\hat{\mathbf{H}}^{(l)}\mathbf{W}_\text{self},
\qquad
\mathbf{H}_\text{pred}
=
\mathbf{A}^\top\hat{\mathbf{H}}^{(l)}\mathbf{W}_\text{pred},
\qquad
\mathbf{H}_\text{succ}
=
\mathbf{A}\hat{\mathbf{H}}^{(l)}\mathbf{W}_\text{succ}.
\end{equation}
Under the convention $\mathbf{A}_{u,v}=1$ for $u\to v$, $\mathbf{H}_\text{self}$ transforms each node's own state, $\mathbf{A}^{\top}\hat{\mathbf{H}}^{(l)}$ aggregates predecessor contexts, and $\mathbf{A}\hat{\mathbf{H}}^{(l)}$ aggregates successor contexts.
The routing gate produces node-wise expert weights and fuses the expert outputs as
\begin{equation}
\resizebox{0.93\linewidth}{!}{$\displaystyle
[\mathbf{g}_\text{self},\mathbf{g}_\text{pred},\mathbf{g}_\text{succ}]
=
\mathrm{softmax}(\hat{\mathbf{H}}^{(l)}\mathbf{W}_\text{gate}),
\;
\tilde{\mathbf{H}}
=
\mathbf{g}_\text{self}\odot\mathbf{H}_\text{self}
+
\mathbf{g}_\text{pred}\odot\mathbf{H}_\text{pred}
+
\mathbf{g}_\text{succ}\odot\mathbf{H}_\text{succ}
$}
\end{equation}
where $\mathbf{g}_\text{self},\mathbf{g}_\text{pred},\mathbf{g}_\text{succ}\in\mathbb{R}^{n\times 1}$, and the softmax is applied over the three experts.
The layer output is $\mathbf{H}^{(l)}=\hat{\mathbf{H}}^{(l)}+\mathrm{ReLU}(\tilde{\mathbf{H}})\mathbf{W}_\text{proj}$; $\odot$ denotes broadcast multiplication, and all $\mathbf{W}$ terms are learnable projections.

\myparagraph{Routing regularization.}
To discourage overly sharp routing distributions, we use a negative-entropy regularizer $\mathcal{L}_{\mathrm{aux}}=\frac{1}{n}\sum\nolimits_{i=1}^{n}\sum\nolimits_{r\in\{\text{self},\text{pred},\text{succ}\}}g_{i,r}\log(g_{i,r}+\epsilon)$.
It encourages more balanced expert weights without imposing role labels on the router.

    \subsection{Attention-Based Graph Readout}\label{sec:readout}

After direction-aware encoding, the readout summarizes final node states without adding virtual tokens, avoiding artificial topological relations that would alter structural biases and magnetic spectral computation.
This is important because the structural attention channels and magnetic Laplacian are both computed from the original architecture topology; adding a virtual node would introduce extra edges or reachability patterns that are not part of the cell DAG.

\myparagraph{Attention pooling.}
We use attention-weighted pooling, with $\boldsymbol{\alpha}=\mathrm{softmax}\!\left(\mathbf{H}^{(L)}\mathbf{r}\right)$ and $\mathbf{z}_G=\sum_{i=1}^{n}\alpha_i\mathbf{H}^{(L)}_i$; $\mathbf{r}\in\mathbb{R}^{d}$ is a learnable scoring vector, and the softmax is applied over nodes.
This pooling yields a graph-level representation $\mathbf{z}_G\in\mathbb{R}^{d}$ without changing the architecture graph.

    \subsection{Prediction and Training Objective}\label{sec:loss}

\myparagraph{Prediction head}. 
Given the graph-level representation $\mathbf{z}_G$, a linear head predicts the target scalar attribute, such as accuracy or latency, as $\hat{y}=p_\psi(\mathbf{z}_G)=\mathbf{w}^\top\mathbf{z}_G+b$, with $\psi=\{\mathbf{w},b\}$.
We keep the predictor head deliberately simple, so most modeling capacity remains in the direction-aware graph encoder rather than in a task-specific regressor.
Architecture attribute prediction requires both calibrated value estimates and reliable attribute ranking, so we train with point-wise regression, pairwise ranking, and the routing regularizer introduced in Sec.~\ref{sec:dir_ffn}.

\myparagraph{Training objective.}
For a mini-batch of size $B$, the mean squared error (MSE) loss calibrates attribute values as $\mathcal{L}_{\mathrm{mse}}=\frac{1}{B}\sum_{i=1}^{B}(\hat{y}_i-y_i)^2$.
To preserve the relative order of candidate architectures under the target attribute, we use a pairwise ranking loss over sampled non-tied architecture pairs $\mathcal{P}$: $\mathcal{L}_{\mathrm{rank}}=\frac{1}{|\mathcal{P}|}\sum_{(i,j)\in\mathcal{P}}\log\!\big(1+\exp(-\mathrm{sgn}(y_i-y_j)(\hat{y}_i-\hat{y}_j))\big)$.
The final objective is $\mathcal{L}_{\mathrm{total}}=\lambda_{\mathrm{mse}}\mathcal{L}_{\mathrm{mse}}+\lambda_{\mathrm{rank}}\mathcal{L}_{\mathrm{rank}}+\lambda_{\mathrm{aux}}\mathcal{L}_{\mathrm{aux}}$, with $\mathcal{L}_{\mathrm{aux}}$ denoting the negative-entropy routing regularizer for the Direction-Sensitive MoE-FFN; the $\lambda$ terms control the relative weights of the three objectives.

\section{Experiments}\label{Experiment}

\subsection{Experimental Setup}
We evaluate accuracy and latency prediction on NAS-Bench-101~\citep{ying2019nasbench101} and NAS-Bench-201~\citep{dong2020nasbench201}; dataset statistics, sources of latency labels, and splits are provided in Appendix~\ref{app:datasets-splits-budgets}.
We compare with ReNAS~\citep{xu2021renas}, NAO~\citep{luo2018nao}, BRP-NAS~\citep{dudziak2020brpnas}, NNLP~\citep{liu2022nnlqp}, Graphormer~\citep{ying2021graphormer}, TNASP~\citep{lu2021tnasp}, NAR-Former~\citep{yi2023narformer}, PINAT~\citep{lu2023pinat}, GraphTrans~\citep{wu2021graphtrans}, NAR-Former V2~\citep{yi2023narformerv2}, and NN-Former~\citep{xu2025nnformer}, and report Kendall's $\tau$ (KT) and MAPE, where higher $\tau$ and lower MAPE are better.
We evaluate how well predictors trained on a small labeled subset estimate architecture attributes and recover the ranking of the full candidate pool.
Tables~\ref{tab:main_results}--\ref{tab:direction_ablation} report five-run means.
Appendix~\ref{app:complete-statistics} provides the corresponding standard deviations.
Baseline sources, reproduction strategies, hardware, and code release information are provided in Appendix~\ref{app:experimental-details}, and DTA-GT hyperparameters and sensitivity analyses for $k$, $L$, and $d$ are given in Appendix~\ref{app:hyperparameter-settings}.

\subsection{Main Results}
Across the 16 task--budget combinations in Table~\ref{tab:main_results}, DTA-GT achieves the highest Kendall's $\tau$ and the lowest MAPE among the evaluated predictors.
The design differences are summarized in Table~\ref{tab:method_positioning}.
CNN- and RNN-based predictors retain connectivity through tensor or sequence representations without explicitly organizing interactions around directed graph relations.
GNN-based predictors model local neighborhoods, with longer-range dependencies obtained through repeated propagation.
Transformer-based and hybrid predictors incorporate structural and directional information into selected encoder components.
Guided by encoder-wide directional consistency, DTA-GT models directed node roles during initialization, introduces a global spectral prior sensitive to edge orientation, distinguishes directional relations during attention, and adaptively weights upstream and downstream information during feature updates.
These stages jointly describe how operations are positioned, connected, and combined in directed computation, providing architecture representations for both accuracy and latency prediction.

\begin{table}[!t]
    \begingroup
    \def\mainresultsscope{0}
    \def\mainresultsfontsize{\fontsize{7pt}{8pt}\selectfont}
    \def\mainresultstext#1{\textbf{\boldmath #1}}
    \def\mainresultstabcolsep{0.8pt}
    \def\mainresultsarraystretch{0.90}
    \def\samplehead#1#2{\shortstack{\mainresultstext{#1\%}\\[-0.875ex]\mainresultstext{(@#2)}}}
    \def\mainresultsgroupheaderbreak{\\[-0.90ex]}
    \def\mainresultssampleheaderbreak{\\[-0.70ex]}
    \def\methodref#1#2{#1}
    \providecommand{\metricvsep}{\hspace{0.45pt}\vrule width 0.2pt\hspace{0.45pt}}
\makeatletter
\providecommand{\sidehead}[1]{%
    \begingroup
    \setbox0=\hbox{\samplehead{0.02}{100}}%
    \dimen0=\ht0%
    \ifdim\dimen0<\ht\@arstrutbox\dimen0=\ht\@arstrutbox\fi%
    \dimen2=\dp0%
    \ifdim\dimen2<\dp\@arstrutbox\dimen2=\dp\@arstrutbox\fi%
    \dimen0=\dimexpr(\dimen0+\dimen2-\ht\@arstrutbox-\dp\@arstrutbox+\aboverulesep+\belowrulesep+\cmidrulewidth)/2\relax%
    \multirow[c]{2}{*}[-\dimen0]{#1}%
    \endgroup
}
\makeatother
\providecommand{\samplehead}[2]{\shortstack{#1\%\\[-0.2ex](@#2)}}
\providecommand{\methodref}[2]{#1}
\providecommand{\result}[2]{#1}
\providecommand{\mainresultstext}[1]{#1}
\providecommand{\mainresultsuparrow}{\textuparrow}
\providecommand{\mainresultsdownarrow}{\textdownarrow}
\providecommand{\mainbestresult}[2]{\result{\textbf{#1}}{\textbf{#2}}}
\providecommand{\mainresultsscope}{1}
\providecommand{\mainresultsfontsize}{\normalsize}
\providecommand{\mainresultstabcolsep}{2pt}
\providecommand{\mainresultsarraystretch}{0.98}
\providecommand{\mainresultsgroupheaderbreak}{\\}
\providecommand{\mainresultssampleheaderbreak}{\\}
\providecommand{\naomethodref}{\methodref{NAO}{luo2018nao}}
\providecommand{\naoresult}[2]{\result{#1}{#2}}

\ifnum\mainresultsscope<3\relax
    \def\mainresultsmetrichead#1{\multicolumn{4}{c|}{\mainresultstext{#1}}}
    \def\mainresultstoprule{\noalign{\hrule height 1.1pt}}
    \let\mainresultsbottomrule\mainresultstoprule
    \providecommand{\segline}{\cmidrule(lr){1-11}}
    \providecommand{\metricrule}{\cmidrule(lr){4-7}\cmidrule(lr){8-11}}
    \providecommand{\mainresultsheaders}{\sidehead{\mainresultstext{Backbone}} & \sidehead{\mainresultstext{Method}} & \sidehead{\mainresultstext{Publication}}}
    \providecommand{\mainresultsemptyheaders}{& &}
    \providecommand{\mainresultscategoryline}{\specialrule{0.2pt}{0.15ex}{0.15ex}}
    \providecommand{\mainresultsoursentry}{& \mainresultstext{DTA-GT} & \mainresultstext{--}}
    \providecommand{\mainresultscolspec}{@{}>{\centering\arraybackslash}p{\dimexpr0.10\linewidth-2\tabcolsep\relax}|>{\raggedright\arraybackslash}p{\dimexpr0.155\linewidth-2\tabcolsep\relax}|>{\raggedright\arraybackslash}p{\dimexpr0.11\linewidth-2\tabcolsep\relax}|>{\centering\arraybackslash}p{\dimexpr0.078125\linewidth-2\tabcolsep\relax}|>{\centering\arraybackslash}p{\dimexpr0.078125\linewidth-2\tabcolsep\relax}|>{\centering\arraybackslash}p{\dimexpr0.078125\linewidth-2\tabcolsep\relax}|>{\centering\arraybackslash}p{\dimexpr0.078125\linewidth-2\tabcolsep\relax}|>{\centering\arraybackslash}p{\dimexpr0.078125\linewidth-2\tabcolsep\relax}|>{\centering\arraybackslash}p{\dimexpr0.078125\linewidth-2\tabcolsep\relax}|>{\centering\arraybackslash}p{\dimexpr0.078125\linewidth-2\tabcolsep\relax}|>{\centering\arraybackslash}p{\dimexpr0.078125\linewidth-2\tabcolsep\relax}@{}}
    \expandafter\def\csname mainbackbone@xu2021renas\endcsname{CNN}
    \expandafter\def\csname mainbackbone@luo2018nao\endcsname{LSTM}
    \expandafter\def\csname mainbackbone@liu2022nnlqp\endcsname{\multirow[c]{2}{*}{GNN}}
    \expandafter\def\csname mainbackbone@dudziak2020brpnas\endcsname{}
    \expandafter\def\csname mainbackbone@ying2021graphormer\endcsname{\multirow[c]{4}{*}{Transformer}}
    \expandafter\def\csname mainbackbone@lu2021tnasp\endcsname{}
    \expandafter\def\csname mainbackbone@yi2023narformer\endcsname{}
    \expandafter\def\csname mainbackbone@lu2023pinat\endcsname{}
    \expandafter\def\csname mainbackbone@wu2021graphtrans\endcsname{\multirow[c]{3}{*}{Hybrid}}
    \expandafter\def\csname mainbackbone@yi2023narformerv2\endcsname{}
    \expandafter\def\csname mainbackbone@xu2025nnformer\endcsname{}
    \expandafter\def\csname mainpublication@xu2021renas\endcsname{CVPR 2021}
    \expandafter\def\csname mainpublication@luo2018nao\endcsname{NeurIPS 2018}
    \expandafter\def\csname mainpublication@liu2022nnlqp\endcsname{ICPP 2022}
    \expandafter\def\csname mainpublication@dudziak2020brpnas\endcsname{NeurIPS 2020}
    \expandafter\def\csname mainpublication@ying2021graphormer\endcsname{NeurIPS 2021}
    \expandafter\def\csname mainpublication@lu2021tnasp\endcsname{NeurIPS 2021}
    \expandafter\def\csname mainpublication@yi2023narformer\endcsname{CVPR 2023}
    \expandafter\def\csname mainpublication@lu2023pinat\endcsname{AAAI 2023}
    \expandafter\def\csname mainpublication@wu2021graphtrans\endcsname{NeurIPS 2021}
    \expandafter\def\csname mainpublication@yi2023narformerv2\endcsname{NeurIPS 2023}
    \expandafter\def\csname mainpublication@xu2025nnformer\endcsname{CVPR 2025}
    \let\mainresultsbasemethodref\methodref
    \renewcommand{\methodref}[2]{\mainresultstext{\csname mainbackbone@#2\endcsname} & \mainresultstext{\mainresultsbasemethodref{#1}{#2}} & \mainresultstext{\csname mainpublication@#2\endcsname}}
    \ifnum\mainresultsscope=0\relax
        \renewcommand{\naomethodref}{\raisebox{-0.5pt}{\mainresultstext{LSTM}} & \raisebox{-0.5pt}{\mainresultstext{\mainresultsbasemethodref{NAO}{luo2018nao}}} & \raisebox{-0.5pt}{\mainresultstext{NeurIPS 2018}}}
        \renewcommand{\naoresult}[2]{\raisebox{-0.5pt}{\result{#1}{#2}}}
    \fi
    \providecommand{\panelheading}[1]{%
        \multicolumn{11}{@{}l}{\rule{0pt}{1.2em}\hspace{0.35em}\mainresultstext{#1}} \\
        \cmidrule(lr){1-11}}
    \providecommand{\nextpanelheading}[1]{%
        \noalign{\vskip -0.40ex}
        \cmidrule(lr){1-11}
        \multicolumn{11}{@{}l}{\rule{0pt}{0.30em}\hspace{0.35em}\mainresultstext{#1}} \\[-1.05ex]
        \cmidrule(lr){1-11}}
\else
    \def\mainresultsmetrichead#1{\multicolumn{4}{c}{#1}}
    \let\mainresultstoprule\toprule
    \let\mainresultsbottomrule\bottomrule
    \providecommand{\segline}{\midrule}
    \providecommand{\metricrule}{\cmidrule(lr){2-5}\cmidrule(lr){6-9}}
    \providecommand{\mainresultsheaders}{\sidehead{Method}}
    \providecommand{\mainresultsemptyheaders}{}
    \providecommand{\mainresultscategoryline}{}
    \providecommand{\mainresultsoursentry}{DTA-GT}
    \providecommand{\mainresultscolspec}{@{}>{\raggedright\arraybackslash}p{\dimexpr0.18\linewidth-2\tabcolsep\relax}|>{\centering\arraybackslash}p{\dimexpr0.10125\linewidth-2\tabcolsep\relax}|>{\centering\arraybackslash}p{\dimexpr0.10125\linewidth-2\tabcolsep\relax}|>{\centering\arraybackslash}p{\dimexpr0.10125\linewidth-2\tabcolsep\relax}|>{\centering\arraybackslash}p{\dimexpr0.10125\linewidth-2\tabcolsep\relax}|>{\centering\arraybackslash}p{\dimexpr0.10125\linewidth-2\tabcolsep\relax}|>{\centering\arraybackslash}p{\dimexpr0.10125\linewidth-2\tabcolsep\relax}|>{\centering\arraybackslash}p{\dimexpr0.10125\linewidth-2\tabcolsep\relax}|>{\centering\arraybackslash}p{\dimexpr0.10125\linewidth-2\tabcolsep\relax}@{}}
    \providecommand{\panelheading}[1]{%
        \multicolumn{9}{@{}l}{\rule{0pt}{1.2em}\hspace{0.35em}#1} \\
        \cmidrule(lr){1-9}}
    \providecommand{\nextpanelheading}[1]{%
        \noalign{\vskip 2.2pt}
        \cmidrule(lr){1-9}
        \panelheading{#1}}
\fi
\providecommand{\resultssecondpanelheading}[1]{\nextpanelheading{#1}}
\ifnum\mainresultsscope=0\relax
    \providecommand{\resultsthirdpanelheading}[1]{\nextpanelheading{#1}}
\else
    \providecommand{\resultsthirdpanelheading}[1]{\panelheading{#1}}
\fi

\def\MainResultsPanelA{%
    \panelheading{Panel A: Accuracy prediction on NAS-Bench-101}
    \mainresultsheaders & \mainresultsmetrichead{KT \mainresultsuparrow{} by training budget} & \multicolumn{4}{c}{\mainresultstext{MAPE \mainresultsdownarrow{} by training budget}} \mainresultsgroupheaderbreak
    \metricrule
    \mainresultsemptyheaders & \samplehead{0.02}{100} & \samplehead{0.04}{172} & \samplehead{0.1}{424} & \samplehead{1}{4236} & \samplehead{0.02}{100} & \samplehead{0.04}{172} & \samplehead{0.1}{424} & \samplehead{1}{4236} \mainresultssampleheaderbreak
    \segline
    \methodref{ReNAS}{xu2021renas} & \result{0.0963}{0.0579} & \result{0.1240}{0.0453} & \result{0.1596}{0.0471} & \result{0.5141}{0.1137} & \result{2.8409}{2.7629} & \result{0.8343}{0.1109} & \result{0.5456}{0.1241} & \result{0.0660}{0.0338} \\
    \mainresultscategoryline
    \naomethodref & \naoresult{0.3395}{0.0445} & \naoresult{0.4015}{0.0218} & \naoresult{0.4202}{0.0123} & \naoresult{0.4472}{0.0017} & \naoresult{0.0577}{0.0149} & \naoresult{0.0559}{0.0030} & \naoresult{0.0493}{0.0108} & \naoresult{0.0407}{0.0017} \\
    \mainresultscategoryline
    \methodref{NNLP}{liu2022nnlqp} & \result{0.4452}{0.0188} & \result{0.4936}{0.0319} & \result{0.4987}{0.0470} & \result{0.7809}{0.0046} & \result{0.0665}{0.0192} & \result{0.0412}{0.0023} & \result{0.0399}{0.0118} & \result{0.0252}{0.0007} \\
    \methodref{BRP-NAS}{dudziak2020brpnas} & \result{0.4129}{0.0116} & \result{0.4338}{0.0107} & \result{0.4549}{0.0093} & \result{0.7643}{0.0007} & \result{0.0672}{0.0079} & \result{0.0548}{0.0086} & \result{0.0424}{0.0051} & \result{0.0236}{0.0002} \\
    \mainresultscategoryline
    \methodref{Graphormer}{ying2021graphormer} & \result{0.2229}{0.0687} & \result{0.2610}{0.0314} & \result{0.2991}{0.0202} & \result{0.5610}{0.0013} & \result{0.0589}{0.0081} & \result{0.0536}{0.0124} & \result{0.0477}{0.0054} & \result{0.0355}{0.0003} \\
    \methodref{TNASP}{lu2021tnasp} & \result{0.5091}{0.0079} & \result{0.5749}{0.0125} & \result{0.6404}{0.0109} & \result{0.7663}{0.0035} & \result{0.0443}{0.0020} & \result{0.0393}{0.0010} & \result{0.0377}{0.0009} & \result{0.0250}{0.0003} \\
    \methodref{NAR-Former}{yi2023narformer} & \result{0.4228}{0.0048} & \result{0.4986}{0.0120} & \result{0.5699}{0.0097} & \result{0.8181}{0.0012} & \result{0.0538}{0.0055} & \result{0.0423}{0.0011} & \result{0.0406}{0.0017} & \result{0.0257}{0.0005} \\
    \methodref{PINAT}{lu2023pinat} & \result{0.3496}{0.0553} & \result{0.3994}{0.0050} & \result{0.4243}{0.0078} & \result{0.7162}{0.0153} & \result{0.1339}{0.0558} & \result{0.0516}{0.0101} & \result{0.0406}{0.0025} & \result{0.0271}{0.0012} \\
    \mainresultscategoryline
    \methodref{GraphTrans}{wu2021graphtrans} & \result{0.4300}{0.0229} & \result{0.4438}{0.0419} & \result{0.4666}{0.0067} & \result{0.7702}{0.0035} & \result{0.2735}{0.1625} & \result{0.1401}{0.0688} & \result{0.1337}{0.0583} & \result{0.0252}{0.0004} \\
    \methodref{NAR-Former V2}{yi2023narformerv2} & \result{0.5300}{0.0100} & \result{0.5500}{0.0040} & \result{0.6000}{0.0040} & \result{0.8150}{0.0025} & \result{0.0900}{0.0450} & \result{0.0850}{0.0300} & \result{0.0399}{0.0090} & \result{0.0220}{0.0007} \\
    \methodref{NN-Former}{xu2025nnformer} & \result{0.5428}{0.0107} & \result{0.5642}{0.0034} & \result{0.6057}{0.0037} & \result{0.8209}{0.0024} & \result{0.0895}{0.0472} & \result{0.0907}{0.0314} & \result{0.0368}{0.0016} & \result{0.0219}{0.0007} \\
    \mainresultscategoryline
    \rowcolor{gray!10} \mainresultsoursentry & \mainbestresult{0.6954}{0.0105} & \mainbestresult{0.7278}{0.0047} & \mainbestresult{0.7980}{0.0038} & \mainbestresult{0.8740}{0.0001} & \mainbestresult{0.0327}{0.0001} & \mainbestresult{0.0312}{0.0006} & \mainbestresult{0.0283}{0.0004} & \mainbestresult{0.0202}{0.0017} \\
}
\def\MainResultsPanelB{%
    \resultssecondpanelheading{Panel B: Latency prediction on NAS-Bench-101}
    \mainresultsheaders & \mainresultsmetrichead{KT \mainresultsuparrow{} by training budget} & \multicolumn{4}{c}{\mainresultstext{MAPE \mainresultsdownarrow{} by training budget}} \mainresultsgroupheaderbreak
    \metricrule
    \mainresultsemptyheaders & \samplehead{0.02}{100} & \samplehead{0.04}{172} & \samplehead{0.1}{424} & \samplehead{1}{4236} & \samplehead{0.02}{100} & \samplehead{0.04}{172} & \samplehead{0.1}{424} & \samplehead{1}{4236} \mainresultssampleheaderbreak
    \segline
    \methodref{ReNAS}{xu2021renas} & \result{0.5803}{0.0108} & \result{0.7127}{0.0278} & \result{0.8028}{0.0048} & \result{0.9222}{0.0024} & \result{0.2232}{0.0051} & \result{0.1588}{0.0130} & \result{0.1080}{0.0026} & \result{0.0405}{0.0009} \\
    \mainresultscategoryline
    \naomethodref & \naoresult{0.2569}{0.0372} & \naoresult{0.2583}{0.0509} & \naoresult{0.3049}{0.0752} & \naoresult{0.3964}{0.0004} & \naoresult{0.3057}{0.0016} & \naoresult{0.3068}{0.0009} & \naoresult{0.3062}{0.0007} & \naoresult{0.2878}{0.0009} \\
    \mainresultscategoryline
    \methodref{NNLP}{liu2022nnlqp} & \result{0.8056}{0.0180} & \result{0.8379}{0.0034} & \result{0.8685}{0.0014} & \result{0.9190}{0.0024} & \result{0.1064}{0.0103} & \result{0.0858}{0.0019} & \result{0.0695}{0.0009} & \result{0.0417}{0.0013} \\
    \methodref{BRP-NAS}{dudziak2020brpnas} & \result{0.7937}{0.0030} & \result{0.8487}{0.0027} & \result{0.8702}{0.0015} & \result{0.9270}{0.0006} & \result{0.1105}{0.0008} & \result{0.0811}{0.0021} & \result{0.0682}{0.0004} & \result{0.0372}{0.0003} \\
    \mainresultscategoryline
    \methodref{Graphormer}{ying2021graphormer} & \result{0.3582}{0.0079} & \result{0.3711}{0.0094} & \result{0.4084}{0.0176} & \result{0.4467}{0.0007} & \result{0.2862}{0.0076} & \result{0.2715}{0.0015} & \result{0.2630}{0.0025} & \result{0.2617}{0.0013} \\
    \methodref{TNASP}{lu2021tnasp} & \result{0.8024}{0.0082} & \result{0.8325}{0.0056} & \result{0.8734}{0.0034} & \result{0.9394}{0.0009} & \result{0.1110}{0.0046} & \result{0.0897}{0.0033} & \result{0.0647}{0.0015} & \result{0.0311}{0.0005} \\
    \methodref{NAR-Former}{yi2023narformer} & \result{0.8337}{0.0023} & \result{0.8670}{0.0031} & \result{0.8836}{0.0031} & \result{0.9434}{0.0007} & \result{0.0881}{0.0011} & \result{0.0686}{0.0011} & \result{0.0606}{0.0015} & \result{0.0287}{0.0003} \\
    \methodref{PINAT}{lu2023pinat} & \result{0.8317}{0.0119} & \result{0.8621}{0.0006} & \result{0.8817}{0.0018} & \result{0.9389}{0.0005} & \result{0.0860}{0.0050} & \result{0.0704}{0.0004} & \result{0.0599}{0.0011} & \result{0.0315}{0.0003} \\
    \mainresultscategoryline
    \methodref{GraphTrans}{wu2021graphtrans} & \result{0.8182}{0.0148} & \result{0.8619}{0.0026} & \result{0.8862}{0.0011} & \result{0.9365}{0.0007} & \result{0.0949}{0.0085} & \result{0.0710}{0.0008} & \result{0.0592}{0.0005} & \result{0.0329}{0.0004} \\
    \methodref{NAR-Former V2}{yi2023narformerv2} & \result{0.7741}{0.0115} & \result{0.8314}{0.0069} & \result{0.8752}{0.0039} & \result{0.9380}{0.0019} & \result{0.1238}{0.0069} & \result{0.0888}{0.0038} & \result{0.0652}{0.0024} & \result{0.0317}{0.0007} \\
    \methodref{NN-Former}{xu2025nnformer} & \result{0.8385}{0.0011} & \result{0.8694}{0.0022} & \result{0.9082}{0.0006} & \result{0.9462}{0.0006} & \result{0.0907}{0.0005} & \result{0.0662}{0.0011} & \result{0.0513}{0.0004} & \result{0.0286}{0.0003} \\
    \mainresultscategoryline
    \rowcolor{gray!10} \mainresultsoursentry & \mainbestresult{0.8646}{0.0036} & \mainbestresult{0.8930}{0.0012} & \mainbestresult{0.9094}{0.0010} & \mainbestresult{0.9490}{0.0007} & \mainbestresult{0.0758}{0.0009} & \mainbestresult{0.0613}{0.0021} & \mainbestresult{0.0503}{0.0022} & \mainbestresult{0.0271}{0.0010} \\
}
\def\MainResultsPanelC{%
    \resultsthirdpanelheading{Panel C: Accuracy prediction on NAS-Bench-201}
    \mainresultsheaders & \mainresultsmetrichead{KT \mainresultsuparrow{} by training budget} & \multicolumn{4}{c}{\mainresultstext{MAPE \mainresultsdownarrow{} by training budget}} \mainresultsgroupheaderbreak
    \metricrule
    \mainresultsemptyheaders & \samplehead{1}{156} & \samplehead{3}{469} & \samplehead{5}{781} & \samplehead{10}{1563} & \samplehead{1}{156} & \samplehead{3}{469} & \samplehead{5}{781} & \samplehead{10}{1563} \mainresultssampleheaderbreak
    \segline
    \methodref{ReNAS}{xu2021renas} & \result{0.1816}{0.1025} & \result{0.2236}{0.1178} & \result{0.4097}{0.0893} & \result{0.7130}{0.0338} & \result{1.9822}{1.7129} & \result{0.8425}{0.5541} & \result{0.3444}{0.0594} & \result{0.1676}{0.0185} \\
    \mainresultscategoryline
    \naomethodref & \naoresult{0.6254}{0.0108} & \naoresult{0.6436}{0.0170} & \naoresult{0.6445}{0.0071} & \naoresult{0.6580}{0.0100} & \naoresult{0.0478}{0.0026} & \naoresult{0.0454}{0.0048} & \naoresult{0.0399}{0.0057} & \naoresult{0.0283}{0.0016} \\
    \mainresultscategoryline
    \methodref{NNLP}{liu2022nnlqp} & \result{0.7109}{0.0230} & \result{0.7380}{0.0151} & \result{0.7824}{0.0332} & \result{0.8196}{0.0299} & \result{0.0410}{0.0272} & \result{0.0245}{0.0075} & \result{0.0172}{0.0070} & \result{0.0115}{0.0018} \\
    \methodref{BRP-NAS}{dudziak2020brpnas} & \result{0.6953}{0.0112} & \result{0.7177}{0.0120} & \result{0.7575}{0.0113} & \result{0.7768}{0.0089} & \result{0.0629}{0.0249} & \result{0.0506}{0.0115} & \result{0.0211}{0.0025} & \result{0.0201}{0.0034} \\
    \mainresultscategoryline
    \methodref{Graphormer}{ying2021graphormer} & \result{0.7052}{0.0039} & \result{0.7311}{0.0062} & \result{0.7461}{0.0128} & \result{0.7650}{0.0024} & \result{0.0270}{0.0044} & \result{0.0185}{0.0002} & \result{0.0171}{0.0006} & \result{0.0149}{0.0010} \\
    \methodref{TNASP}{lu2021tnasp} & \result{0.7405}{0.0080} & \result{0.7770}{0.0073} & \result{0.8061}{0.0016} & \result{0.8314}{0.0013} & \result{0.0302}{0.0007} & \result{0.0158}{0.0019} & \result{0.0145}{0.0044} & \result{0.0110}{0.0006} \\
    \methodref{NAR-Former}{yi2023narformer} & \result{0.7540}{0.0107} & \result{0.7839}{0.0039} & \result{0.8188}{0.0036} & \result{0.8554}{0.0009} & \result{0.0256}{0.0017} & \result{0.0188}{0.0005} & \result{0.0141}{0.0006} & \result{0.0107}{0.0005} \\
    \methodref{PINAT}{lu2023pinat} & \result{0.7027}{0.0167} & \result{0.7084}{0.0006} & \result{0.7589}{0.0021} & \result{0.7920}{0.0034} & \result{0.0338}{0.0055} & \result{0.0273}{0.0048} & \result{0.0269}{0.0123} & \result{0.0198}{0.0110} \\
    \mainresultscategoryline
    \methodref{GraphTrans}{wu2021graphtrans} & \result{0.6957}{0.0056} & \result{0.6959}{0.0055} & \result{0.7781}{0.0023} & \result{0.8258}{0.0025} & \result{0.0975}{0.0510} & \result{0.0613}{0.0086} & \result{0.0493}{0.0224} & \result{0.0153}{0.0031} \\
    \methodref{NAR-Former V2}{yi2023narformerv2} & \result{0.7418}{0.0049} & \result{0.7432}{0.0086} & \result{0.8004}{0.0052} & \result{0.8391}{0.0055} & \result{0.1363}{0.0511} & \result{0.0441}{0.0102} & \result{0.0340}{0.0074} & \result{0.0256}{0.0062} \\
    \methodref{NN-Former}{xu2025nnformer} & \result{0.7400}{0.0050} & \result{0.7420}{0.0060} & \result{0.8050}{0.0040} & \result{0.8400}{0.0040} & \result{0.1350}{0.0500} & \result{0.0450}{0.0100} & \result{0.0330}{0.0070} & \result{0.0260}{0.0060} \\
    \mainresultscategoryline
    \rowcolor{gray!10} \mainresultsoursentry & \mainbestresult{0.8015}{0.0133} & \mainbestresult{0.8925}{0.0028} & \mainbestresult{0.9138}{0.0010} & \mainbestresult{0.9304}{0.0011} & \mainbestresult{0.0244}{0.0105} & \mainbestresult{0.0125}{0.0007} & \mainbestresult{0.0096}{0.0012} & \mainbestresult{0.0084}{0.0022} \\
}
\def\MainResultsPanelD{%
    \resultssecondpanelheading{Panel D: Latency prediction on NAS-Bench-201}
    \mainresultsheaders & \mainresultsmetrichead{KT \mainresultsuparrow{} by training budget} & \multicolumn{4}{c}{\mainresultstext{MAPE \mainresultsdownarrow{} by training budget}} \mainresultsgroupheaderbreak
    \metricrule
    \mainresultsemptyheaders & \samplehead{1}{156} & \samplehead{3}{469} & \samplehead{5}{781} & \samplehead{10}{1563} & \samplehead{1}{156} & \samplehead{3}{469} & \samplehead{5}{781} & \samplehead{10}{1563} \mainresultssampleheaderbreak
    \segline
    \methodref{ReNAS}{xu2021renas} & \result{0.6166}{0.0032} & \result{0.6506}{0.0070} & \result{0.6725}{0.0005} & \result{0.6863}{0.0004} & \result{0.0844}{0.0009} & \result{0.0777}{0.0009} & \result{0.0707}{0.0004} & \result{0.0640}{0.0006} \\
    \mainresultscategoryline
    \naomethodref & \naoresult{0.6963}{0.0004} & \naoresult{0.6980}{0.0003} & \naoresult{0.6984}{0.0002} & \naoresult{0.7012}{0.0002} & \naoresult{0.0628}{0.0000} & \naoresult{0.0602}{0.0000} & \naoresult{0.0600}{0.0001} & \naoresult{0.0592}{0.0000} \\
    \mainresultscategoryline
    \methodref{NNLP}{liu2022nnlqp} & \result{0.6349}{0.0062} & \result{0.6437}{0.0016} & \result{0.6527}{0.0017} & \result{0.6567}{0.0019} & \result{0.0807}{0.0009} & \result{0.0726}{0.0007} & \result{0.0716}{0.0003} & \result{0.0702}{0.0002} \\
    \methodref{BRP-NAS}{dudziak2020brpnas} & \result{0.6690}{0.0021} & \result{0.6861}{0.0005} & \result{0.6923}{0.0011} & \result{0.6986}{0.0005} & \result{0.0717}{0.0003} & \result{0.0640}{0.0001} & \result{0.0615}{0.0002} & \result{0.0594}{0.0001} \\
    \mainresultscategoryline
    \methodref{Graphormer}{ying2021graphormer} & \result{0.6799}{0.0010} & \result{0.6972}{0.0010} & \result{0.7030}{0.0017} & \result{0.7030}{0.0006} & \result{0.0651}{0.0005} & \result{0.0600}{0.0001} & \result{0.0595}{0.0002} & \result{0.0589}{0.0001} \\
    \methodref{TNASP}{lu2021tnasp} & \result{0.6646}{0.0069} & \result{0.6873}{0.0010} & \result{0.6926}{0.0008} & \result{0.6990}{0.0005} & \result{0.0713}{0.0010} & \result{0.0633}{0.0001} & \result{0.0613}{0.0001} & \result{0.0595}{0.0002} \\
    \methodref{NAR-Former}{yi2023narformer} & \result{0.5916}{0.0637} & \result{0.6014}{0.0393} & \result{0.6275}{0.0101} & \result{0.6522}{0.0027} & \result{0.1309}{0.0070} & \result{0.1299}{0.0096} & \result{0.1229}{0.0001} & \result{0.1226}{0.0001} \\
    \methodref{PINAT}{lu2023pinat} & \result{0.6859}{0.0107} & \result{0.6992}{0.0008} & \result{0.6950}{0.0019} & \result{0.6996}{0.0001} & \result{0.0645}{0.0010} & \result{0.0607}{0.0002} & \result{0.0602}{0.0003} & \result{0.0593}{0.0001} \\
    \mainresultscategoryline
    \methodref{GraphTrans}{wu2021graphtrans} & \result{0.6441}{0.0059} & \result{0.6871}{0.0013} & \result{0.6964}{0.0006} & \result{0.6993}{0.0008} & \result{0.0762}{0.0013} & \result{0.0643}{0.0006} & \result{0.0612}{0.0001} & \result{0.0600}{0.0003} \\
    \methodref{NAR-Former V2}{yi2023narformerv2} & \result{0.6237}{0.0201} & \result{0.6474}{0.0060} & \result{0.6541}{0.0032} & \result{0.6781}{0.0030} & \result{0.0786}{0.0060} & \result{0.0719}{0.0009} & \result{0.0699}{0.0002} & \result{0.0680}{0.0009} \\
    \methodref{NN-Former}{xu2025nnformer} & \result{0.6819}{0.0014} & \result{0.6891}{0.0012} & \result{0.6947}{0.0014} & \result{0.6981}{0.0017} & \result{0.0695}{0.0003} & \result{0.0631}{0.0003} & \result{0.0609}{0.0001} & \result{0.0594}{0.0000} \\
    \mainresultscategoryline
    \rowcolor{gray!10} \mainresultsoursentry & \mainbestresult{0.8375}{0.0048} & \mainbestresult{0.8409}{0.0008} & \mainbestresult{0.8435}{0.0025} & \mainbestresult{0.8492}{0.0003} & \mainbestresult{0.0305}{0.0007} & \mainbestresult{0.0293}{0.0004} & \mainbestresult{0.0280}{0.0010} & \mainbestresult{0.0276}{0.0001} \\
}

\centering
\mainresultsfontsize
\setlength{\tabcolsep}{\mainresultstabcolsep}
\setlength{\arrayrulewidth}{0.2pt}
\renewcommand{\arraystretch}{\mainresultsarraystretch}
\ifcase\mainresultsscope\relax
    \caption{Main results on NAS-Bench-101/201.
    Higher Kendall's $\tau$ (KT) and lower MAPE are better.
    Complete results with std are provided in Appendix~\ref{app:complete-main-results}.}
    \label{tab:main_results}
\or
    \caption{Main results on NAS-Bench-101.
    Higher Kendall's $\tau$ (KT) and lower MAPE are better.
    Complete results with std are provided in Appendix~\ref{app:complete-main-results}.}
    \label{tab:main_results}
\or
    \caption{Main results on NAS-Bench-201.
    Higher Kendall's $\tau$ (KT) and lower MAPE are better.
    Complete results with std are provided in Appendix~\ref{app:complete-main-results}.}
    \label{tab:main_results_nb201}
\or
    \caption{Complete accuracy-prediction results on NAS-Bench-101.
    Each cell places the mean above its standard deviation.}
    \label{tab:main_results_complete_nb101_acc}
\or
    \caption{Complete latency-prediction results on NAS-Bench-101.
    Each cell places the mean above its standard deviation.}
    \label{tab:main_results_complete_nb101_lat}
\or
    \caption{Complete accuracy-prediction results on NAS-Bench-201.
    Each cell places the mean above its standard deviation.}
    \label{tab:main_results_complete_nb201_acc}
\or
    \caption{Complete latency-prediction results on NAS-Bench-201.
    Each cell places the mean above its standard deviation.}
    \label{tab:main_results_complete_nb201_lat}
\fi
\ifcase\mainresultsscope\relax
    \def\mainresultstablecontents{\MainResultsPanelA\MainResultsPanelB\MainResultsPanelC\MainResultsPanelD}
\or
    \def\mainresultstablecontents{\MainResultsPanelA\MainResultsPanelB}
\or
    \def\mainresultstablecontents{\MainResultsPanelC\MainResultsPanelD}
\or
    \def\mainresultstablecontents{\MainResultsPanelA}
\or
    \def\mainresultstablecontents{\MainResultsPanelB}
\or
    \def\mainresultstablecontents{\MainResultsPanelC}
\or
    \def\mainresultstablecontents{\MainResultsPanelD}
\fi
\ifnum\mainresultsscope<3\relax
\edef\mainresultsbasewidth{\the\linewidth}
\else
\edef\mainresultsbasewidth{\the\dimexpr\linewidth*85/100\relax}
\fi
\noindent\makebox[\linewidth][c]{%
\ifnum\mainresultsscope<3\relax
\begin{tabular}{@{}>{\centering\arraybackslash}p{\dimexpr\mainresultsbasewidth*100/1000-2\tabcolsep\relax}|>{\raggedright\arraybackslash}p{\dimexpr\mainresultsbasewidth*155/1000-2\tabcolsep\relax}|>{\raggedright\arraybackslash}p{\dimexpr\mainresultsbasewidth*110/1000-2\tabcolsep\relax}|>{\centering\arraybackslash}p{\dimexpr\mainresultsbasewidth*5/64-2\tabcolsep\relax}|>{\centering\arraybackslash}p{\dimexpr\mainresultsbasewidth*5/64-2\tabcolsep\relax}|>{\centering\arraybackslash}p{\dimexpr\mainresultsbasewidth*5/64-2\tabcolsep\relax}|>{\centering\arraybackslash}p{\dimexpr\mainresultsbasewidth*5/64-2\tabcolsep\relax}|>{\centering\arraybackslash}p{\dimexpr\mainresultsbasewidth*5/64-2\tabcolsep\relax}|>{\centering\arraybackslash}p{\dimexpr\mainresultsbasewidth*5/64-2\tabcolsep\relax}|>{\centering\arraybackslash}p{\dimexpr\mainresultsbasewidth*5/64-2\tabcolsep\relax}|>{\centering\arraybackslash}p{\dimexpr\mainresultsbasewidth*5/64-2\tabcolsep\relax}@{}}
\else
\begin{tabular}{@{}>{\raggedright\arraybackslash}p{\dimexpr\mainresultsbasewidth*20/100-\tabcolsep\relax}*{8}{>{\centering\arraybackslash}p{\dimexpr\mainresultsbasewidth/10-2\tabcolsep\relax}}@{}}
\fi
\mainresultstoprule
\mainresultstablecontents
\mainresultsbottomrule
\end{tabular}%
}

    \endgroup
\end{table}

\begingroup
\setlength{\intextsep}{4pt}
\begin{table}[!t]
    \begingroup
    \providecommand{\yesdir}{\cmark}
\providecommand{\nodir}{\xmark}
\providecommand{\bestpipe}[1]{\textbf{#1}}
\providecommand{\fullpipe}[1]{\cellcolor{gray!20}#1}
\providecommand{\missingresult}{\phantom{0.0000}}
\providecommand{\stagecell}[2]{\makebox[0.32cm][l]{#1:}#2}
\providecommand{\seglinepipe}{\specialrule{\cmidrulewidth}{0pt}{0pt}}

    \centering
    \scriptsize
    \setlength{\tabcolsep}{1.5pt}
    \setlength{\doublerulesep}{0.6pt}
    \renewcommand{\arraystretch}{1.04}
    \setlength{\extrarowheight}{0.15pt}
    \setlength{\aboverulesep}{0pt}
    \setlength{\belowrulesep}{0pt}
    \setlength{\cmidrulesep}{0pt}
    \captionsetup{aboveskip=2pt,belowskip=2pt,justification=raggedright,singlelinecheck=false}
    \caption{Ablation results under low-budget predictor training.
    Crosses indicate the absence of the corresponding component or directional design in each variant.}
    \label{tab:pipeline_consistency_ablation}
    \newlength{\pipelabelwidth}
    \setlength{\pipelabelwidth}{1.85cm}
    \newlength{\pipebasewidth}
    \settowidth{\pipebasewidth}{\bestpipe{0.0000}}
    \newlength{\pipeheaderwidth}
    \newcommand{\pipefitheader}[1]{%
        \settowidth{\pipeheaderwidth}{#1}%
        \setlength{\pipeheaderwidth}{\dimexpr(\pipeheaderwidth-2\tabcolsep)/2\relax}%
        \ifdim\pipeheaderwidth>\pipebasewidth\setlength{\pipebasewidth}{\pipeheaderwidth}\fi%
    }
    \pipefitheader{Directed}
    \pipefitheader{Undirected}
    \pipefitheader{w/o Dir. Feat.}
    \pipefitheader{w/o Spec.}
    \pipefitheader{w/o Attn.}
    \pipefitheader{w/o MoE}
    \pipefitheader{CLS Readout}
    \pipefitheader{MSE Only}
    \resizebox{\linewidth}{!}{%
    \begin{tabular}{>{\raggedright\arraybackslash}p{\pipelabelwidth}|*{2}{>{\centering\arraybackslash}p{\pipebasewidth}}|*{2}{>{\centering\arraybackslash}p{\pipebasewidth}}||*{2}{>{\centering\arraybackslash}p{\pipebasewidth}}|*{2}{>{\centering\arraybackslash}p{\pipebasewidth}}|*{2}{>{\centering\arraybackslash}p{\pipebasewidth}}|*{2}{>{\centering\arraybackslash}p{\pipebasewidth}}||*{2}{>{\centering\arraybackslash}p{\pipebasewidth}}|*{2}{>{\centering\arraybackslash}p{\pipebasewidth}}}
    \noalign{\hrule height 1.1pt}
    {\let\strut\relax\multirow[c]{2}{=}[-\dimexpr(\aboverulesep+\belowrulesep+\cmidrulewidth)/2\relax]{\raggedright Component key}} & \multicolumn{4}{c||}{Overall Directionality} & \multicolumn{8}{c||}{Stage Contributions} & \multicolumn{4}{c}{Readout / Objective} \\
    \cmidrule{2-5}\cmidrule{6-13}\cmidrule{14-17}
     & \multicolumn{2}{c|}{\cellcolor{gray!20}Directed} & \multicolumn{2}{c||}{Undirected} & \multicolumn{2}{c|}{w/o Dir. Feat.} & \multicolumn{2}{c|}{w/o Spec.} & \multicolumn{2}{c|}{w/o Attn.} & \multicolumn{2}{c||}{w/o MoE} & \multicolumn{2}{c|}{CLS Readout} & \multicolumn{2}{c}{MSE Only} \\
    \seglinepipe
    \stagecell{N}{Dir. Node Feat.} & \multicolumn{2}{c|}{\cellcolor{gray!20}\yesdir} & \multicolumn{2}{c||}{\nodir} & \multicolumn{2}{c|}{\nodir} & \multicolumn{2}{c|}{\yesdir} & \multicolumn{2}{c|}{\yesdir} & \multicolumn{2}{c||}{\yesdir} & \multicolumn{2}{c|}{\yesdir} & \multicolumn{2}{c}{\yesdir} \\[-1.2pt]
    \stagecell{P}{Spec. Enc.} & \multicolumn{2}{c|}{\cellcolor{gray!20}\yesdir} & \multicolumn{2}{c||}{\nodir} & \multicolumn{2}{c|}{\yesdir} & \multicolumn{2}{c|}{\nodir} & \multicolumn{2}{c|}{\yesdir} & \multicolumn{2}{c||}{\yesdir} & \multicolumn{2}{c|}{\yesdir} & \multicolumn{2}{c}{\yesdir} \\[-1.2pt]
    \stagecell{A}{Struct. Attn.} & \multicolumn{2}{c|}{\cellcolor{gray!20}\yesdir} & \multicolumn{2}{c||}{\nodir} & \multicolumn{2}{c|}{\yesdir} & \multicolumn{2}{c|}{\yesdir} & \multicolumn{2}{c|}{\nodir} & \multicolumn{2}{c||}{\yesdir} & \multicolumn{2}{c|}{\yesdir} & \multicolumn{2}{c}{\yesdir} \\[-1.2pt]
    \stagecell{F}{MoE-FFN} & \multicolumn{2}{c|}{\cellcolor{gray!20}\yesdir} & \multicolumn{2}{c||}{\nodir} & \multicolumn{2}{c|}{\yesdir} & \multicolumn{2}{c|}{\yesdir} & \multicolumn{2}{c|}{\yesdir} & \multicolumn{2}{c||}{\nodir} & \multicolumn{2}{c|}{\yesdir} & \multicolumn{2}{c}{\yesdir} \\
    \seglinepipe
    Metric & \fullpipe{KT} & \fullpipe{MAPE} & KT & MAPE & KT & MAPE & KT & MAPE & KT & MAPE & KT & MAPE & KT & MAPE & KT & MAPE \\
    \seglinepipe
    NB101 Acc @100 & \fullpipe{\bestpipe{0.6954}} & \fullpipe{\bestpipe{0.0327}} & 0.5882 & 0.0408 & 0.6522 & 0.0361 & 0.6894 & 0.0341 & 0.6552 & 0.0331 & 0.6542 & 0.0331 & 0.6701 & 0.0399 & 0.6720 & 0.0356 \\
    NB101 Lat @100 & \fullpipe{\bestpipe{0.8646}} & \fullpipe{\bestpipe{0.0758}} & 0.8062 & 0.1074 & 0.8482 & 0.0908 & 0.8615 & 0.0806 & 0.8360 & 0.1033 & 0.8495 & 0.0860 & 0.8420 & 0.1203 & 0.8610 & 0.0808 \\
    NB201 Acc @156 & \fullpipe{\bestpipe{0.8015}} & \fullpipe{\bestpipe{0.0244}} & 0.7215 & 0.0617 & 0.7198 & 0.0729 & 0.7452 & 0.0600 & 0.6123 & 0.1037 & 0.7248 & 0.0418 & 0.7535 & 0.0704 & 0.6937 & 0.0686 \\
    NB201 Lat @156 & \fullpipe{\bestpipe{0.8375}} & \fullpipe{\bestpipe{0.0305}} & 0.8178 & 0.0349 & 0.8202 & 0.0326 & 0.8192 & 0.0325 & 0.8227 & 0.0317 & 0.8298 & 0.0313 & 0.8209 & 0.0348 & 0.8201 & 0.0321 \\
    \noalign{\hrule height 1.1pt}
    \end{tabular}%
    }

    \par
    \endgroup
    \vspace{-11pt}
\end{table}
\endgroup
\subsection{Core Design Ablations}\label{sec:ablation}
    Table~\ref{tab:pipeline_consistency_ablation} examines the overall benefit of directional modeling and the contribution of each encoder stage under low-budget settings.
    The undirected control uses symmetric adjacency and removes directional distinctions from node features, attention relations, and gated feature updates.
    Magnetic Laplacian encoding is also omitted (Appendix~\ref{app:undirected-control}).
    The ``w/o Directed Node Features'' variant retains operation embeddings but omits in-degree, out-degree, forward-depth, and reverse-depth features.
    DTA-GT outperforms its undirected counterpart across all reported metrics.
    Removing directed node features or magnetic spectral encoding reduces performance.
    This supports using both node roles and the global pattern of edge directions for prediction.
    Removing structural attention or MoE-FFN also reduces performance.
    This supports distinguishing directed relations during attention and adaptively combining upstream and downstream information during updates.
    The size of these effects varies across tasks.
    Together, these ablations support the value of encoder-wide directional consistency.
    Attention pooling outperforms CLS readout.
    It summarizes the existing nodes without changing graph connectivity, preserving the directional relations used by structural attention and magnetic spectral encoding.
    The MSE-only control also underperforms the proposed training objective.
    MSE fits numerical attribute values, and the ranking term supervises the relative order of candidate architectures.

\subsection{Magnetic Spectral Encoding Analysis}
\begingroup
\setlength{\intextsep}{0pt}
\begin{table}[H]
    \providecommand{\yesdir}{\cmark}
\providecommand{\nodir}{\xmark}
\providecommand{\bestpipe}[1]{\textbf{#1}}
\providecommand{\seglinephase}{\cmidrule(lr){1-11}}

    \centering
    \fontsize{7.25pt}{8.5pt}\selectfont
    \setlength{\tabcolsep}{1.4pt}
    \renewcommand{\arraystretch}{0.86}
    \newcommand{\phasehead}[1]{{\let\strut\relax\multirow[c]{2}{=}[-\dimexpr(\aboverulesep+\belowrulesep+\cmidrulewidth)/2\relax]{#1}}}
    \captionsetup{aboveskip=2pt,belowskip=2pt}
    \caption{Symmetric and magnetic Laplacian spectral encoding controls.
    Bold marks the best value.}
    \label{tab:direction_ablation}
    \begin{tabular}{@{}>{\raggedright\arraybackslash}p{\dimexpr0.24\linewidth-2\tabcolsep\relax}|>{\centering\arraybackslash}p{\dimexpr0.085\linewidth-2\tabcolsep\relax}>{\centering\arraybackslash}p{\dimexpr0.085\linewidth-2\tabcolsep\relax}|*{3}{>{\centering\arraybackslash}p{\dimexpr0.07375\linewidth-2\tabcolsep\relax}>{\centering\arraybackslash}p{\dimexpr0.07375\linewidth-2\tabcolsep\relax}|}>{\centering\arraybackslash}p{\dimexpr0.07375\linewidth-2\tabcolsep\relax}>{\centering\arraybackslash}p{\dimexpr0.07375\linewidth-2\tabcolsep\relax}@{}}
    \noalign{\hrule height 1.1pt}
    \noalign{\vskip 0.35ex}
    \phasehead{Variant} & \phasehead{Sym. Lap.} & \phasehead{Mag. Lap.} & \multicolumn{2}{|c|}{\raisebox{-0.48ex}{NB101 Acc @100}} & \multicolumn{2}{c|}{\raisebox{-0.48ex}{NB101 Lat @100}} & \multicolumn{2}{c|}{\raisebox{-0.48ex}{NB201 Acc @156}} & \multicolumn{2}{c}{\raisebox{-0.48ex}{NB201 Lat @156}} \\
    \cmidrule(lr){4-11}
     & & & KT & MAPE & KT & MAPE & KT & MAPE & KT & MAPE \\
    \seglinephase
    No Directional Encoding & \nodir & \nodir & 0.5882 & 0.0408 & 0.8062 & 0.1074 & 0.7215 & 0.0617 & 0.8178 & 0.0349 \\
    No Direction + Symmetric Lap. & \yesdir & \nodir & 0.5878 & 0.0386 & 0.8071 & 0.1066 & 0.7135 & 0.0485 & 0.8206 & 0.0330 \\
    Directed with Symmetric Lap. & \yesdir & \nodir & 0.6894 & 0.0341 & 0.8615 & 0.0806 & 0.7148 & 0.0792 & 0.8192 & 0.0325 \\
    \rowcolor{gray!20} Directed with Magnetic Lap. & \nodir & \yesdir & \bestpipe{0.6954} & \bestpipe{0.0327} & \bestpipe{0.8646} & \bestpipe{0.0758} & \bestpipe{0.8015} & \bestpipe{0.0244} & \bestpipe{0.8375} & \bestpipe{0.0305} \\
    \noalign{\hrule height 1.1pt}
    \end{tabular}

\end{table}
\endgroup
    Table~\ref{tab:direction_ablation} compares four configurations to examine the role of direction-aware spectral encoding.
    The two undirected variants either omit spectral encoding or use symmetric Laplacian encoding.
    This tests whether global connectivity information helps when direction is ignored.
    The two directed variants retain the same direction-aware node initialization, structural attention, and MoE-FFN, but use symmetric or magnetic Laplacian encoding.
    This isolates the benefit of edge orientation in the spectral prior.
    Symmetric spectral encoding captures connectivity but cannot distinguish different edge orientations on the same undirected skeleton.
    Magnetic spectral encoding makes the global structural context sensitive to edge direction, complementing the directional information used in node representations, attention, and feature updates.
    Its better performance across the evaluated tasks supports extending directional consistency to the global spectral stage, rather than relying only on the other direction-aware components.
    Sensitivity to the magnetic phase parameter $q$ is reported in Appendix~\ref{app:magnetic-phase-analysis} and Table~\ref{tab:hyperparameter-q}.

\subsection{Mechanistic Analysis}
\begin{figure}[H]
    \centering
    \includegraphics[width=\linewidth]{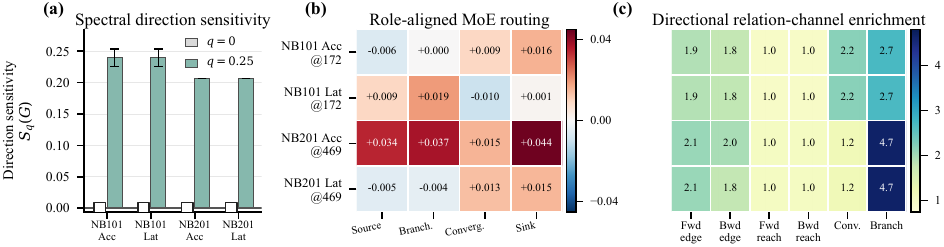}
    \captionsetup{aboveskip=2pt,belowskip=2pt}
    \caption{Mechanistic analysis.
    (a) Magnetic Laplacian sensitivity to edge reversal at $q=0$ and $0.25$.
    (b) Role-aligned MoE gate contrasts (positive: toward aligned context).
    (c) Density-normalized relation attention ($>1$: above the density baseline).}
    \label{fig:mechanism_diagnostics}
    \vspace{-6pt}
\end{figure}

    We analyze the spectral encoding, MoE router, and structural attention to examine whether their internal behavior matches the intended directional design (Figure~\ref{fig:mechanism_diagnostics}).
    Reversing all edges changes the magnetic Laplacian at $q=0.25$ but leaves it unchanged at $q=0$.
    This shows that the spectral operator captures edge orientation even when undirected connectivity is unchanged.
    Routing preferences vary across tasks and node roles, consistent with adapting the balance of upstream and downstream context to different computational roles and prediction targets.
    Local edge and shared-neighbor relations receive more attention relative to their coverage than directed reachability.
    These differences illustrate how attention is distributed across direct, multi-hop, and branch-related dependencies within the supplied relation channels.
    Together, these observations are consistent with the intended roles of global orientation-sensitive encoding, relation-specific interaction, and adaptive directional updates under encoder-wide directional consistency.
    Tables~\ref{tab:main_results}--\ref{tab:direction_ablation} assess the predictive benefits, and Appendix~\ref{app:additional-mechanistic-diagnostics} provides the diagnostic definitions and detailed results.

\subsection{Compact Model Evaluation}
\begingroup
\setlength{\intextsep}{0pt}
\begin{table}[H]
\centering
\caption{Compact model comparison: parameter count, inference time, and prediction quality under the lowest training budgets (100 for NB101 and 156 for NB201).}
\label{tab:compact_model_quality_mean}
\small
\setlength{\tabcolsep}{2pt}
\renewcommand{\arraystretch}{1.10}
\newcommand{\compacthead}[1]{{\let\strut\relax\multirow[c]{2}{*}[-\dimexpr(\aboverulesep+\belowrulesep+\cmidrulewidth)/2\relax]{#1}}}
\begin{tabular}{@{}lcc*{8}{c}@{}}
\toprule
\multicolumn{1}{c}{\compacthead{Model}} & \compacthead{\shortstack{Parameters\\(M)}} & \compacthead{\shortstack{Inference\\(ms/sample)}} & \multicolumn{2}{c}{NB101 Acc} & \multicolumn{2}{c}{NB101 Lat} & \multicolumn{2}{c}{NB201 Acc} & \multicolumn{2}{c}{NB201 Lat} \\
\cmidrule(lr){4-5}\cmidrule(lr){6-7}\cmidrule(lr){8-9}\cmidrule(lr){10-11}
 & & & KT $\uparrow$ & MAPE $\downarrow$ & KT $\uparrow$ & MAPE $\downarrow$ & KT $\uparrow$ & MAPE $\downarrow$ & KT $\uparrow$ & MAPE $\downarrow$ \\
\midrule
Graphormer & 0.61 & 0.050 & 0.2229 & 0.0589 & 0.3582 & 0.2862 & 0.7052 & \textbf{0.0270} & 0.6799 & 0.0651 \\
DTA-GT-S & \textbf{0.36} & \textbf{0.049} & \textbf{0.6603} & \textbf{0.0359} & \textbf{0.8612} & \textbf{0.0777} & \textbf{0.7114} & 0.0486 & \textbf{0.8183} & \textbf{0.0371} \\
\bottomrule
\end{tabular}
\end{table}

\endgroup
To assess whether DTA-GT's performance gains arise from its directional design or simply a larger parameter count, we evaluate a compact variant, DTA-GT-S.
It retains the directional components and uses fewer parameters than Graphormer (Table~\ref{tab:compact_model_quality_mean}).
Even with fewer parameters and comparable inference time, DTA-GT-S generally outperforms Graphormer, showing that these gains do not require a larger parameter budget.
DTA-GT-S preserves encoder-wide directional consistency across node representations, global spectral encoding, attention, and feature updates.
Together with the ablations in Table~\ref{tab:pipeline_consistency_ablation}, these results support the contribution of this coordinated directional design to predictive performance.
Appendix~\ref{app:efficiency} provides the detailed configuration and analysis.

\section{Discussion}\label{sec:discussion}

\myparagraph{Limitations.}
Our evaluation of architecture prediction focuses on NAS-Bench-101 and NAS-Bench-201.
Zero-shot cross-space transfer remains unverified, and adapting to other spaces requires aligned operation vocabularies, node granularity, and dependency semantics.
Our large-DAG profiling covers up to 32,768 nodes (Appendix~\ref{app:large-dag-efficiency}).
Reachability can still become dense, leading to quadratic storage and attention costs; million-node scaling remains unresolved.
The effect of improved predictor ranking on end-to-end NAS remains to be evaluated.

\myparagraph{Future work.}
We will investigate a shared operation-level representation for cross-space pretraining and adaptation, and explore sparse or hierarchical attention with compact reachability representations to support larger computational graphs.
We also plan to integrate DTA-GT into NAS pipelines and evaluate the selected architectures and total search cost under a fixed end-to-end evaluation budget.
Appendix~\ref{app:limitations-impact} elaborates on these directions.

\section{Conclusion}\label{Conclusion}

We formulated \textbf{encoder-wide directional consistency} as an intuitive and unifying design principle for systematically modeling directed computational semantics throughout an architecture encoder.
We developed DTA-GT, a Direction- and Topology-Aware Graph Transformer for neural architecture attribute prediction, to realize this principle.
Experiments on NAS-Bench-101 and NAS-Bench-201 demonstrate consistent gains in accuracy and latency prediction across limited-budget settings.
Ablations and magnetic spectral controls support the contributions of the direction-aware components, and mechanistic diagnostics characterize their learned behavior.


\newpage
\begin{ack}
The work was supported in part by the National Natural Science Foundation of China under Grants~62321166652 and U25B2038, in part by the Zhejiang Provincial Key Research and Development Program under Grant~2026C02A1244, and in part by the Shaoguan Data Industry Research Institute under Grant~26ND0205.
\end{ack}

\section*{Preprint Note}
This is a preprint and is still being revised.
The camera-ready version will be authoritative.

\bibliographystyle{unsrtnat}
\bibliography{bibReferences}

\begin{thebibliography}{44}
\providecommand{\natexlab}[1]{#1}
\providecommand{\url}[1]{\texttt{#1}}
\expandafter\ifx\csname urlstyle\endcsname\relax
  \providecommand{\doi}[1]{doi: #1}\else
  \providecommand{\doi}{doi: \begingroup \urlstyle{rm}\Url}\fi

\bibitem[Huang et~al.(2026)Huang, Shang, Chen, Su, and Chen]{huang2026tools}
Tairan Huang, Siyu Shang, Qiang Chen, Xiu Su, and Yi~Chen.
\newblock Tools as continuous flow for evolving agentic reasoning.
\newblock \emph{arXiv preprint arXiv:2605.07339}, 2026.

\bibitem[Wang et~al.(2026)Wang, Shen, Gao, Wang, Dai, Lu, Zhang, and
  Zheng]{wang2026libra}
Yongcong Wang, Chengchao Shen, Guangwei Gao, Wei Wang, Pengwen Dai, Dianjie Lu,
  Guijuan Zhang, and Zhuoran Zheng.
\newblock {LiBrA-Net}: Lie-algebraic bilateral affine fields for real-time {4K}
  video dehazing.
\newblock \emph{arXiv preprint arXiv:2605.11508}, 2026.

\bibitem[Zhang et~al.(2025)Zhang, Song, Lee, Guo, Wang, Li, Guo, Liu, Lian, and
  Chen]{Zhang2025_UniGRF}
Luankang Zhang, Kenan Song, Yi~Quan Lee, Wei Guo, Hao Wang, Yawen Li, Huifeng
  Guo, Yong Liu, Defu Lian, and Enhong Chen.
\newblock Killing two birds with one stone: Unifying retrieval and ranking with
  a single generative recommendation model.
\newblock In \emph{Proceedings of the 48th International ACM SIGIR Conference
  on Research and Development in Information Retrieval}, pages 2224--2234,
  2025.

\bibitem[Zeng et~al.(2023{\natexlab{a}})Zeng, Li, Xia, Du, Wang, Lian, and
  Xu]{10234360}
Yuxiang Zeng, Yuhui Li, Zhiyu Xia, Zibo Du, Jialin Wang, Ruimin Lian, and
  Jianlong Xu.
\newblock Qoseraser: A data erasable framework for web service qos prediction.
\newblock In \emph{2023 IEEE International Conference on Software Services
  Engineering (SSE)}, pages 89--97, 2023{\natexlab{a}}.
\newblock \doi{10.1109/SSE60056.2023.00022}.

\bibitem[Zeng et~al.(2023{\natexlab{b}})Zeng, Xu, Zhang, Chen, Ling, and
  Wang]{s23156775}
Yuxiang Zeng, Jianlong Xu, Zhuohua Zhang, Caiyi Chen, Qianyu Ling, and Jialin
  Wang.
\newblock {GATCF}: Graph attention collaborative filtering for reliable
  blockchain services selection in {BaaS}.
\newblock \emph{Sensors}, 23\penalty0 (15):\penalty0 6775, 2023{\natexlab{b}}.
\newblock \doi{10.3390/s23156775}.
\newblock URL \url{https://www.mdpi.com/1424-8220/23/15/6775}.

\bibitem[Ye et~al.(2025)Ye, Zhang, Peng, Li, Liu, Wu, and Wong]{ye2025cross}
Yixuan Ye, Yang Zhang, Liang Peng, Rui Li, Cheng Liu, Si~Wu, and Hau-San Wong.
\newblock Cross-view neighborhood contrastive multi-view clustering with view
  mixup feature learning.
\newblock In \emph{2025 IEEE International Conference on Multimedia and Expo
  (ICME)}, pages 1--6. IEEE, 2025.

\bibitem[Liu et~al.(2026)Liu, Zhu, Mo, Li, Cao, Jin, Shen, Li, Yu, Yuan, Ding,
  and Lourentzou]{liu2026palm}
Yuanzhe Liu, Jingyuan Zhu, Yuchen Mo, Gen Li, Xu~Cao, Jin Jin, Yifan Shen,
  Zhengyuan Li, Tianjiao Yu, Wenzhen Yuan, Fangqiang Ding, and Ismini
  Lourentzou.
\newblock {PALM}: Progress-aware policy learning via affordance reasoning for
  long-horizon robotic manipulation.
\newblock \emph{arXiv preprint arXiv:2601.07060}, 2026.

\bibitem[Shen et~al.(2025)Shen, Liu, Zhu, Cao, Zhang, He, Ye, Rehg, and
  Lourentzou]{shen2026fine}
Yifan Shen, Yuanzhe Liu, Jingyuan Zhu, Xu~Cao, Xiaofeng Zhang, Yixiao He,
  Wenming Ye, James~M. Rehg, and Ismini Lourentzou.
\newblock Fine-grained preference optimization improves spatial reasoning in
  {VLMs}.
\newblock In \emph{Advances in Neural Information Processing Systems},
  volume~38, 2025.
\newblock \doi{10.52202/085713-0606}.

\bibitem[Cai et~al.(2018)Cai, Yang, Zhang, Han, and Yu]{cai2018pathlevel}
Han Cai, Jiacheng Yang, Weinan Zhang, Song Han, and Yong Yu.
\newblock Path-level network transformation for efficient architecture search.
\newblock In \emph{International Conference on Machine Learning}, pages
  678--687. PMLR, 2018.

\bibitem[Liu et~al.(2018)Liu, Zoph, Neumann, Shlens, Hua, Li, Fei-Fei, Yuille,
  Huang, and Murphy]{liu2018pnas}
Chenxi Liu, Barret Zoph, Maxim Neumann, Jonathon Shlens, Wei Hua, Li-Jia Li,
  Li~Fei-Fei, Alan Yuille, Jonathan Huang, and Kevin Murphy.
\newblock Progressive neural architecture search.
\newblock In \emph{Proceedings of the European Conference on Computer Vision},
  pages 19--34, 2018.

\bibitem[Pham et~al.(2018)Pham, Guan, Zoph, Le, and Dean]{pham2018enas}
Hieu Pham, Melody Guan, Barret Zoph, Quoc Le, and Jeff Dean.
\newblock Efficient neural architecture search via parameters sharing.
\newblock In \emph{International Conference on Machine Learning}, pages
  4095--4104. PMLR, 2018.

\bibitem[Cai et~al.(2019)Cai, Zhu, and Han]{cai2019proxylessnas}
Han Cai, Ligeng Zhu, and Song Han.
\newblock {ProxylessNAS}: Direct neural architecture search on target task and
  hardware.
\newblock In \emph{International Conference on Learning Representations}, 2019.
\newblock URL \url{https://openreview.net/forum?id=HylVB3AqYm}.

\bibitem[Tan et~al.(2019)Tan, Chen, Pang, Vasudevan, Sandler, Howard, and
  Le]{tan2019mnasnet}
Mingxing Tan, Bo~Chen, Ruoming Pang, Vijay Vasudevan, Mark Sandler, Andrew
  Howard, and Quoc~V Le.
\newblock {MnasNet}: Platform-aware neural architecture search for mobile.
\newblock In \emph{Proceedings of the IEEE/CVF Conference on Computer Vision
  and Pattern Recognition}, pages 2820--2828, 2019.

\bibitem[Wu et~al.(2019)Wu, Dai, Zhang, Wang, Sun, Wu, Tian, Vajda, Jia, and
  Keutzer]{wu2019fbnet}
Bichen Wu, Xiaoliang Dai, Peizhao Zhang, Yanghan Wang, Fei Sun, Yiming Wu,
  Yuandong Tian, Peter Vajda, Yangqing Jia, and Kurt Keutzer.
\newblock {FBNet}: Hardware-aware efficient convnet design via differentiable
  neural architecture search.
\newblock In \emph{Proceedings of the IEEE/CVF Conference on Computer Vision
  and Pattern Recognition}, pages 10734--10742, 2019.

\bibitem[Dudziak et~al.(2020)Dudziak, Chau, Abdelfattah, Lee, Kim, and
  Lane]{dudziak2020brpnas}
Lukasz Dudziak, Thomas Chau, Mohamed Abdelfattah, Royson Lee, Hyeji Kim, and
  Nicholas Lane.
\newblock {BRP-NAS}: Prediction-based {NAS} using {GCNs}.
\newblock \emph{Advances in Neural Information Processing Systems},
  33:\penalty0 10480--10490, 2020.

\bibitem[Yi et~al.(2023{\natexlab{a}})Yi, Zhang, Hu, Wang, and
  Wang]{yi2023narformer}
Yun Yi, Haokui Zhang, Wenze Hu, Nannan Wang, and Xiaoyu Wang.
\newblock {NAR-Former}: Neural architecture representation learning towards
  holistic attributes prediction.
\newblock In \emph{Proceedings of the IEEE/CVF Conference on Computer Vision
  and Pattern Recognition}, pages 7715--7724, 2023{\natexlab{a}}.

\bibitem[Yi et~al.(2023{\natexlab{b}})Yi, Zhang, Xiao, Wang, and
  Wang]{yi2023narformerv2}
Yun Yi, Haokui Zhang, Rong Xiao, Nannan Wang, and Xiaoyu Wang.
\newblock {NAR-Former V2}: Rethinking transformer for universal neural network
  representation learning.
\newblock In A.~Oh, T.~Naumann, A.~Globerson, K.~Saenko, M.~Hardt, and
  S.~Levine, editors, \emph{Advances in Neural Information Processing Systems},
  volume~36, pages 62727--62739. Curran Associates, Inc., 2023{\natexlab{b}}.
\newblock URL
  \url{https://proceedings.neurips.cc/paper_files/paper/2023/file/c60bd92a01804b7df0540ed7ca2f7c05-Paper-Conference.pdf}.

\bibitem[Xu et~al.(2025)Xu, Zhang, Wang, Zeng, and Zhang]{xu2025nnformer}
Ruihan Xu, Haokui Zhang, Yaowei Wang, Wei Zeng, and Shiliang Zhang.
\newblock {NN-Former}: Rethinking graph structure in neural architecture
  representation.
\newblock In \emph{Proceedings of the IEEE/CVF Conference on Computer Vision
  and Pattern Recognition}, pages 10004--10014, 2025.

\bibitem[Wen et~al.(2020)Wen, Liu, Chen, Li, Bender, and
  Kindermans]{wen2020neuralpredictor}
Wei Wen, Hanxiao Liu, Yiran Chen, Hai Li, Gabriel Bender, and Pieter-Jan
  Kindermans.
\newblock Neural predictor for neural architecture search.
\newblock In \emph{European Conference on Computer Vision}, pages 660--676.
  Springer, 2020.

\bibitem[Ning et~al.(2020)Ning, Zheng, Zhao, Wang, and Yang]{ning2020gates}
Xuefei Ning, Yin Zheng, Tianchen Zhao, Yu~Wang, and Huazhong Yang.
\newblock A generic graph-based neural architecture encoding scheme for
  predictor-based {NAS}.
\newblock In \emph{European Conference on Computer Vision}, pages 189--204.
  Springer, 2020.

\bibitem[Liu et~al.(2022)Liu, Shen, Gong, Yu, and Yang]{liu2022nnlqp}
Liang Liu, Mingzhu Shen, Ruihao Gong, Fengwei Yu, and Hailong Yang.
\newblock {NNLQP}: A multi-platform neural network latency query and prediction
  system with an evolving database.
\newblock In \emph{Proceedings of the 51st International Conference on Parallel
  Processing}, pages 1--14, 2022.

\bibitem[Wei et~al.(2022)Wei, Niu, Tang, Wang, Hu, and Liang]{wei2023npenas}
Chen Wei, Chuang Niu, Yiping Tang, Yue Wang, Haihong Hu, and Jimin Liang.
\newblock {NPENAS}: Neural predictor guided evolution for neural architecture
  search.
\newblock \emph{IEEE Transactions on Neural Networks and Learning Systems},
  34\penalty0 (11):\penalty0 8441--8455, 2022.

\bibitem[Alon and Yahav(2021)]{alon2021bottleneck}
Uri Alon and Eran Yahav.
\newblock On the bottleneck of graph neural networks and its practical
  implications.
\newblock In \emph{International Conference on Learning Representations}, 2021.
\newblock URL \url{https://arxiv.org/abs/2006.05205}.

\bibitem[Oono and Suzuki(2020)]{oono2020graph}
Kenta Oono and Taiji Suzuki.
\newblock Graph neural networks exponentially lose expressive power for node
  classification.
\newblock In \emph{International Conference on Learning Representations}, 2020.
\newblock URL \url{https://arxiv.org/abs/1905.10947}.

\bibitem[Kim et~al.(2022)Kim, Nguyen, Min, Cho, Lee, Lee, and
  Hong]{kim2022pure}
Jinwoo Kim, Dat Nguyen, Seonwoo Min, Sungjun Cho, Moontae Lee, Honglak Lee, and
  Seunghoon Hong.
\newblock Pure transformers are powerful graph learners.
\newblock In S.~Koyejo, S.~Mohamed, A.~Agarwal, D.~Belgrave, K.~Cho, and A.~Oh,
  editors, \emph{Advances in Neural Information Processing Systems}, volume~35,
  pages 14582--14595. Curran Associates, Inc., 2022.
\newblock \doi{10.52202/068431-1060}.
\newblock URL
  \url{https://proceedings.neurips.cc/paper_files/paper/2022/file/5d84236751fe6d25dc06db055a3180b0-Paper-Conference.pdf}.

\bibitem[Ying et~al.(2021)Ying, Cai, Luo, Zheng, Ke, He, Shen, and
  Liu]{ying2021graphormer}
Chengxuan Ying, Tianle Cai, Shengjie Luo, Shuxin Zheng, Guolin Ke, Di~He,
  Yanming Shen, and Tie-Yan Liu.
\newblock Do transformers really perform badly for graph representation?
\newblock In M.~Ranzato, A.~Beygelzimer, Y.~Dauphin, P.~S. Liang, and
  J.~Wortman Vaughan, editors, \emph{Advances in Neural Information Processing
  Systems}, volume~34, pages 28877--28888. Curran Associates, Inc., 2021.
\newblock URL
  \url{https://proceedings.neurips.cc/paper_files/paper/2021/file/f1c1592588411002af340cbaedd6fc33-Paper.pdf}.

\bibitem[Lu et~al.(2021)Lu, Li, Tan, Yang, and Liu]{lu2021tnasp}
Shun Lu, Jixiang Li, Jianchao Tan, Sen Yang, and Ji~Liu.
\newblock {TNASP}: A transformer-based {NAS} predictor with a self-evolution
  framework.
\newblock In M.~Ranzato, A.~Beygelzimer, Y.~Dauphin, P.~S. Liang, and
  J.~Wortman Vaughan, editors, \emph{Advances in Neural Information Processing
  Systems}, volume~34, pages 15125--15137. Curran Associates, Inc., 2021.
\newblock URL
  \url{https://proceedings.neurips.cc/paper_files/paper/2021/file/7fa1575cbd7027c9a799983a485c3c2f-Paper.pdf}.

\bibitem[Lu et~al.(2023)Lu, Hu, Wang, Han, Tan, Li, Yang, and Liu]{lu2023pinat}
Shun Lu, Yu~Hu, Peihao Wang, Yan Han, Jianchao Tan, Jixiang Li, Sen Yang, and
  Ji~Liu.
\newblock {PINAT}: A permutation {IN}variance augmented transformer for {NAS}
  predictor.
\newblock \emph{Proceedings of the AAAI Conference on Artificial Intelligence},
  37:\penalty0 8957--8965, Jun. 2023.
\newblock \doi{10.1609/aaai.v37i7.26076}.
\newblock URL \url{https://ojs.aaai.org/index.php/AAAI/article/view/26076}.

\bibitem[Xu et~al.(2021)Xu, Wang, Han, Tang, Jui, Xu, and Xu]{xu2021renas}
Yixing Xu, Yunhe Wang, Kai Han, Yehui Tang, Shangling Jui, Chunjing Xu, and
  Chang Xu.
\newblock {ReNAS}: Relativistic evaluation of neural architecture search.
\newblock In \emph{Proceedings of the IEEE/CVF Conference on Computer Vision
  and Pattern Recognition}, pages 4411--4420, 2021.

\bibitem[Deng et~al.(2017)Deng, Yan, and Lin]{deng2017peephole}
Boyang Deng, Junjie Yan, and Dahua Lin.
\newblock Peephole: Predicting network performance before training.
\newblock \emph{arXiv preprint arXiv:1712.03351}, 2017.

\bibitem[Luo et~al.(2018)Luo, Tian, Qin, Chen, and Liu]{luo2018nao}
Renqian Luo, Fei Tian, Tao Qin, Enhong Chen, and Tie-Yan Liu.
\newblock Neural architecture optimization.
\newblock \emph{Advances in Neural Information Processing Systems}, 31, 2018.

\bibitem[Luo et~al.(2020)Luo, Tan, Wang, Qin, Chen, and Liu]{luo2020seminas}
Renqian Luo, Xu~Tan, Rui Wang, Tao Qin, Enhong Chen, and Tie-Yan Liu.
\newblock Semi-supervised neural architecture search.
\newblock \emph{Advances in Neural Information Processing Systems},
  33:\penalty0 10547--10557, 2020.

\bibitem[Gao et~al.(2023)Gao, Gu, Zhang, Lin, and Yang]{gao2023dnnperf}
Yanjie Gao, Xianyu Gu, Hongyu Zhang, Haoxiang Lin, and Mao Yang.
\newblock Runtime performance prediction for deep learning models with graph
  neural network.
\newblock In \emph{2023 IEEE/ACM 45th International Conference on Software
  Engineering: Software Engineering in Practice}, pages 368--380, 2023.
\newblock \doi{10.1109/ICSE-SEIP58684.2023.00039}.

\bibitem[He et~al.(2016)He, Zhang, Ren, and Sun]{he2016resnet}
Kaiming He, Xiangyu Zhang, Shaoqing Ren, and Jian Sun.
\newblock Deep residual learning for image recognition.
\newblock In \emph{Proceedings of the IEEE Conference on Computer Vision and
  Pattern Recognition}, June 2016.

\bibitem[Geisler et~al.(2023)Geisler, Li, Mankowitz, Cemgil, G{\"u}nnemann, and
  Paduraru]{geisler2023transformers}
Simon Geisler, Yujia Li, Daniel~J Mankowitz, Ali~Taylan Cemgil, Stephan
  G{\"u}nnemann, and Cosmin Paduraru.
\newblock Transformers meet directed graphs.
\newblock In \emph{Proceedings of the 40th International Conference on Machine
  Learning}, volume 202 of \emph{Proceedings of Machine Learning Research},
  pages 11144--11172. PMLR, 2023.
\newblock URL \url{https://proceedings.mlr.press/v202/geisler23a.html}.

\bibitem[Furutani et~al.(2020)Furutani, Shibahara, Akiyama, Hato, and
  Aida]{furutani2019hermitian}
Satoshi Furutani, Toshiki Shibahara, Mitsuaki Akiyama, Kunio Hato, and Masaki
  Aida.
\newblock Graph signal processing for directed graphs based on the hermitian
  laplacian.
\newblock In Ulf Brefeld, Elisa Fromont, Andreas Hotho, Arno Knobbe, Marloes
  Maathuis, and C{\'e}line Robardet, editors, \emph{Machine Learning and
  Knowledge Discovery in Databases}, pages 447--463, Cham, 2020. Springer
  International Publishing.
\newblock ISBN 978-3-030-46150-8.

\bibitem[Zhang et~al.(2021)Zhang, He, Brugnone, Perlmutter, and
  Hirn]{zhang2021magnet}
Xitong Zhang, Yixuan He, Nathan Brugnone, Michael Perlmutter, and Matthew Hirn.
\newblock {MagNet}: A neural network for directed graphs.
\newblock In M.~Ranzato, A.~Beygelzimer, Y.~Dauphin, P.~S. Liang, and
  J.~Wortman Vaughan, editors, \emph{Advances in Neural Information Processing
  Systems}, volume~34, pages 27003--27015. Curran Associates, Inc., 2021.
\newblock URL
  \url{https://proceedings.neurips.cc/paper_files/paper/2021/file/e32084632d369461572832e6582aac36-Paper.pdf}.

\bibitem[Vaswani et~al.(2017)Vaswani, Shazeer, Parmar, Uszkoreit, Jones, Gomez,
  Kaiser, and Polosukhin]{vaswani2017attention}
Ashish Vaswani, Noam Shazeer, Niki Parmar, Jakob Uszkoreit, Llion Jones,
  Aidan~N Gomez, {\L}ukasz Kaiser, and Illia Polosukhin.
\newblock Attention is all you need.
\newblock In I.~Guyon, U.~Von Luxburg, S.~Bengio, H.~Wallach, R.~Fergus,
  S.~Vishwanathan, and R.~Garnett, editors, \emph{Advances in Neural
  Information Processing Systems}, volume~30. Curran Associates, Inc., 2017.
\newblock URL
  \url{https://proceedings.neurips.cc/paper_files/paper/2017/file/3f5ee243547dee91fbd053c1c4a845aa-Paper.pdf}.

\bibitem[Luo et~al.(2023)Luo, Thost, and Shi]{luo2023transformers}
Yuankai Luo, Veronika Thost, and Lei Shi.
\newblock Transformers over directed acyclic graphs.
\newblock In \emph{Advances in Neural Information Processing Systems},
  volume~36, pages 47764--47782, 2023.
\newblock \doi{10.52202/075280-2070}.
\newblock URL
  \url{https://proceedings.neurips.cc/paper_files/paper/2023/hash/94e85561a342de88b559b72c9b29f638-Abstract-Conference.html}.

\bibitem[Ying et~al.(2019)Ying, Klein, Christiansen, Real, Murphy, and
  Hutter]{ying2019nasbench101}
Chris Ying, Aaron Klein, Eric Christiansen, Esteban Real, Kevin Murphy, and
  Frank Hutter.
\newblock {NAS-Bench-101}: Towards reproducible neural architecture search.
\newblock In \emph{International Conference on Machine Learning}, pages
  7105--7114. PMLR, 2019.

\bibitem[Dong and Yang(2020)]{dong2020nasbench201}
Xuanyi Dong and Yi~Yang.
\newblock {NAS-Bench-201}: Extending the scope of reproducible neural
  architecture search.
\newblock In \emph{International Conference on Learning Representations}, 2020.
\newblock URL \url{https://openreview.net/forum?id=HJxyZkBKDr}.

\bibitem[Wu et~al.(2021)Wu, Jain, Wright, Mirhoseini, Gonzalez, and
  Stoica]{wu2021graphtrans}
Zhanghao Wu, Paras Jain, Matthew Wright, Azalia Mirhoseini, Joseph Gonzalez,
  and Ion Stoica.
\newblock Representing long-range context for graph neural networks with global
  attention.
\newblock In M.~Ranzato, A.~Beygelzimer, Y.~Dauphin, P.~S. Liang, and
  J.~Wortman Vaughan, editors, \emph{Advances in Neural Information Processing
  Systems}, volume~34, pages 13266--13279. Curran Associates, Inc., 2021.
\newblock URL
  \url{https://proceedings.neurips.cc/paper_files/paper/2021/file/6e67691b60ed3e4a55935261314dd534-Paper.pdf}.

\bibitem[Freitas et~al.(2021)Freitas, Dong, Neil, and Chau]{freitas2021malnet}
Scott Freitas, Yuxiao Dong, Joshua Neil, and Duen~Horng Chau.
\newblock A large-scale database for graph representation learning.
\newblock In \emph{Proceedings of the Neural Information Processing Systems
  Track on Datasets and Benchmarks}, volume~1, 2021.
\newblock URL
  \url{https://datasets-benchmarks-proceedings.neurips.cc/paper/2021/hash/5fd0b37cd7dbbb00f97ba6ce92bf5add-Abstract-round1.html}.

\bibitem[Wu and Simon(2000)]{wu2000thickrestart}
Kesheng Wu and Horst Simon.
\newblock Thick-restart lanczos method for large symmetric eigenvalue problems.
\newblock \emph{SIAM Journal on Matrix Analysis and Applications}, 22\penalty0
  (2):\penalty0 602--616, 2000.
\newblock \doi{10.1137/S0895479898334605}.

\end{thebibliography}


\clearpage
\appendix

\begingroup
\setlength{\intextsep}{6pt plus 1pt minus 1pt}
\setlength{\textfloatsep}{10pt plus 2pt minus 2pt}
\section{Additional Experimental Details}\label{app:experimental-details}

\subsection{Implementation Platform}
The NAS-Bench-101 and NAS-Bench-201 attribute-prediction experiments are implemented in PyTorch 2.4.0 and conducted on a workstation running Ubuntu 22.04 with an Intel Core i9-13900K CPU, 64 GB RAM, and a single NVIDIA RTX 4090 GPU with 24 GB memory.
The large-DAG scalability experiments are conducted separately on an NVIDIA A800-SXM4-80GB GPU, with the corresponding hardware configuration and timing protocol reported in Appendix~\ref{app:larger-dags}.

\subsection{Datasets, Splits, and Training Budgets}\label{app:datasets-splits-budgets}
NAS-Bench-101~\citep{ying2019nasbench101} contains 423,624 unique cell-based architectures with at most 7 nodes and 9 edges per cell.
NAS-Bench-201~\citep{dong2020nasbench201} contains 15,625 candidate cells with 4 nodes and 6 edges per cell.
For NAS-Bench-201, we represent edge operations as nodes in the predictor input graph, whose encoding topology is fixed across operation assignments.
The NAS-Bench-201 authors trained candidate architectures on CIFAR-10 for 200 epochs and recorded their validation accuracies, alongside latency measurements collected with a batch size of 256.
These measurements serve as labels for our accuracy and latency prediction tasks, with accuracy averaged over the available benchmark runs for each architecture.
NAS-Bench-101 provides accuracy labels but not latency labels; therefore, we measure NAS-Bench-101 latency on an NVIDIA RTX 4090 GPU using batch size 1 to complete the four-task evaluation.
Because latency labels are benchmark- and device-specific, latency magnitudes are compared only within the same benchmark setting.

For each task and training budget, each method is evaluated over five runs using a shared set of five random seeds.
For each seed, architectures are randomly shuffled to form the training and validation partitions.
The partitions vary across seeds, and all methods share the same partitions for a given seed.
The @ notation in the result tables denotes the number of architectures used for predictor fitting, and the validation set contains 1,024 architectures.
Because the predictor serves the candidate screening stage of NAS, its objective is to score and rank the complete architecture pool and identify promising candidates for costly downstream evaluation.
Following the full-pool evaluation setting of NAR-Former~\citep{yi2023narformer}, we evaluate ranking quality over the full benchmark architecture set, including the training and validation architectures and all remaining candidates.
Kendall's $\tau$ and MAPE are both computed over this set.

\subsection{Fair Comparison and Reproducibility}
Within each seed, all methods use the same training and validation partitions, full-set evaluation target, training budgets, and metrics.
When reproducing baselines, we use the official implementations when available and preserve their original model configurations; otherwise, hyperparameters are selected by grid search within the reported or standard ranges.
DTA-GT uses one fixed default configuration across tasks and budgets.
The main comparison and ablation tables report means over five runs; Appendix~\ref{app:complete-statistics} also reports the corresponding standard deviations.

\subsection{Descriptions of Baselines}\label{app:baseline-descriptions}
We compare DTA-GT with representative RNN-, CNN-, GNN-, Transformer-, and hybrid architecture predictors.
NAO uses an LSTM architecture encoder and is included as the RNN-based baseline in Table~\ref{tab:main_results}.
For methods with public source code, we use the released implementation as the primary reference.
For methods for which we did not identify an official public implementation, we reproduce the predictor from the paper description and tune only within the reported or standard ranges under the shared split, budget, and evaluation protocol described above.

\paragraph{RNN- and CNN-based predictors.}
NAO~\citep{luo2018nao} encodes architectures as operation sequences with an encoder--predictor--decoder framework and performs prediction in a continuous architecture embedding space.
ReNAS~\citep{xu2021renas} represents architectures with handcrafted tensor features and trains a relativistic predictor that focuses on ranking architectures rather than only regressing absolute performance.
These baselines provide comparisons with sequence- and tensor-based architecture representations.

\paragraph{GNN-based predictors.}
BRP-NAS~\citep{dudziak2020brpnas} uses GCN-based architecture predictors and emphasizes binary relation prediction for comparing candidate architectures.
NNLP~\citep{liu2022nnlqp} is a multi-platform neural network latency predictor built on graph-structured model representations.
These baselines evaluate local message-passing predictors and latency-oriented graph predictors.

\paragraph{Transformer-based predictors.}
Graphormer~\citep{ying2021graphormer} is a general graph Transformer with structural encodings such as degree, shortest-path, and edge-related biases; we adapt it to architecture DAG prediction.
TNASP~\citep{lu2021tnasp} is a Transformer-based NAS predictor that combines operation embeddings, Laplacian positional encoding, and a self-evolution framework.
NAR-Former~\citep{yi2023narformer} tokenizes operation and topology information into architecture sequences and uses a multi-stage fusion Transformer for holistic attribute prediction.
PINAT~\citep{lu2023pinat} augments a Transformer-like NAS predictor with permutation-invariance modules and Laplacian positional information.
These predictors provide comparisons with alternative Transformer-based representations of architecture structure.

\paragraph{Hybrid predictors.}
GraphTrans~\citep{wu2021graphtrans} combines a GNN module with global Transformer attention to capture long-range graph context.
NAR-Former V2~\citep{yi2023narformerv2} rethinks Transformer-based neural network representation learning for broader architecture-attribute prediction.
NN-Former~\citep{xu2025nnformer} combines topology-aware token mixing and bidirectional graph-isomorphism feed-forward updates to strengthen neural architecture representations.
These hybrid predictors provide comparisons with alternative ways of combining graph structure and attention-based representation learning.

\textbf{Comparison dimensions in Table~\ref{tab:method_positioning}.}
Topology \& Direction Encoding assesses whether the initial node representations explicitly include topological and directional attributes.
Retaining connectivity in the architecture input provides partial coverage when these attributes are not explicitly represented at the node level.
Directional Features / Relations covers dedicated encodings of directional node roles or relations, such as incoming and outgoing edges or predecessor and successor relationships.
Global Directional Spectral Prior denotes a spectral representation derived from the graph as a whole whose construction depends on edge orientation.
Direction-Specific Updates denotes feature propagation or updates that explicitly use directed adjacency or distinguish predecessor and successor contexts.

\begin{table}[t]
\centering
\caption{Public source links and reproduction status for compared baselines.
``Self-reproduced'' denotes paper-based implementation under the shared protocol when no official public code was identified.}
\label{tab:baseline_code_links}
\footnotesize
\setlength{\tabcolsep}{3pt}
\renewcommand{\arraystretch}{1.14}
\begin{tabular}{@{}>{\raggedright\arraybackslash}p{0.15\linewidth}>{\raggedright\arraybackslash}p{0.25\linewidth}>{\raggedright\arraybackslash}p{0.265\linewidth}>{\raggedright\arraybackslash}p{\dimexpr0.335\linewidth-6\tabcolsep\relax}@{}}
\toprule
Method & Public source link & License/availability & Reproduction note \\
\midrule
ReNAS & -- & No official public code identified & Self-reproduced (paper). \\
NAO & \href{https://github.com/renqianluo/NAO}{renqianluo/NAO} & GPL-3.0 & Official code used as reference. \\
BRP-NAS & \href{https://github.com/zheng-ningxin/brp-nas}{zheng-ningxin/brp-nas} & Apache-2.0 & Public code used as reference. \\
NNLP & \href{https://github.com/ModelTC/NNLQP}{ModelTC/NNLQP} & Apache-2.0 & Public code available. \\
Graphormer & \href{https://github.com/microsoft/Graphormer}{microsoft/Graphormer} & MIT & Official code used as reference. \\
TNASP & -- & No official public code identified & Self-reproduced (paper + supp.). \\
NAR-Former & \href{https://github.com/yuny220/NAR-Former}{yuny220/NAR-Former} & Not specified in repository & Official code used as reference. \\
PINAT & \href{https://github.com/ShunLu91/PINAT}{ShunLu91/PINAT} & MIT & Official code used as reference. \\
GraphTrans & \href{https://github.com/ucbrise/graphtrans}{ucbrise/graphtrans} & Apache-2.0 & Official code used as reference. \\
NAR-Former V2 & \href{https://github.com/yuny220/NAR-Former-V2}{yuny220/NAR-Former-V2} & MIT & Official code used as reference. \\
NN-Former & \href{https://github.com/XuRuihan/NNFormer}{XuRuihan/NNFormer} & MIT & Official code used as reference. \\
\bottomrule
\end{tabular}
\end{table}

\subsection{Code and Data Access}\label{app:code-data-access}
Code and experimental scripts are available in the public repository at \url{https://github.com/ZengYuXiang7/DTA-GT}.
The repository contains the proposed DTA-GT implementation, experiment scripts, required benchmark data files, and the measured NAS-Bench-101 latency labels used for the four-task evaluation.
This appendix provides the corresponding dataset, split, budget, hardware, latency-label, and hyperparameter details needed to interpret and reproduce the reported results.

\subsection{Direction-Aware Spectral Encoding Settings}\label{app:maglap-settings}
Due to space constraints, the main text presents the combinatorial magnetic Laplacian in Equation~\eqref{eq:maglap} and omits the numerical details of degree normalization, graph scaling, and eigenvector phase normalization.
Inspired by Geisler et al.~\citep{geisler2023transformers}, we compute spectral features from its degree-normalized counterpart using the effective parameter $q_G=q/\max\{1,\min(m,n)\}$, where $n$ and $m$ denote the numbers of nodes and directed edges.
We use the nominal parameter $q=0.25$.
The same normalized operator and graph-scaled potential are used for the edge-reversal diagnostic in Appendix~\ref{app:magnetic-phase-analysis}.
We retain the first $k=4$ eigenvectors ordered by ascending eigenvalues.

To reduce eigenvector ambiguity, we use the numerical normalization convention of Geisler et al.~\citep{geisler2023transformers}.
We first adjust the sign of each eigenvector so that its largest-magnitude real component is nonnegative.
We then select the reference node with the largest phase in the first eigenvector and rotate each retained eigenvector to set its phase at this node to zero.
The resulting real and imaginary components form the two-channel spectral inputs.
This numerical convention reduces phase ambiguity but does not guarantee invariance under arbitrary basis transformations, particularly for repeated or nearly repeated eigenvalues.

\subsection{Magnetic Spectral Aggregation Details}\label{app:maglap_aggregation}
For each node $u$ and spectral mode $r$, we embed the two-dimensional vector $\mathbf{E}_{u,r}$ containing the real and imaginary parts:
\begin{equation}
\mathbf{Z}_{u,r}=f_{\mathrm{elem}}(\mathbf{E}_{u,r}).
\end{equation}
For each node $u$, the resulting sequence $\mathbf{Z}_{u}=[\mathbf{Z}_{u,1},\ldots,\mathbf{Z}_{u,k}]$ represents its $k$ spectral modes.
We aggregate this sequence with self-attention over the spectral dimension:
\begin{equation}
\widetilde{\mathbf{Z}}_{u}
=
\mathrm{MHA}(\mathbf{Z}_{u},\mathbf{Z}_{u},\mathbf{Z}_{u}).
\end{equation}
The aggregated spectral signal is then flattened across modes and projected to the model dimension:
\begin{equation}
\mathbf{p}_{u}
=
f_{\mathrm{mix}}\!\left(
\operatorname{Concat}_{r=1}^{k}
\widetilde{\mathbf{Z}}_{u,r}
\right).
\end{equation}
Stacking all $\mathbf{p}_u$ gives the spectral encoding $\mathbf{P}\in\mathbb{R}^{n\times d}$, which is added to the node initialization in Eq.~\ref{eq:init_with_mag}.

\subsection{Undirected Control}\label{app:undirected-control}
We construct the undirected counterpart of DTA-GT using a symmetric adjacency matrix.
Node degree and depth features are computed from the graph structure; for each node, the indegree and outdegree inputs have identical values, as do the forward and reverse depth inputs.
Magnetic Laplacian encoding is omitted.
The six directional relation channels are reduced to three for undirected adjacency, reachability, and shared neighbors.
Forward and backward adjacency are merged into one relation, forward and backward reachability into another, and shared successor and shared predecessor relations into a third.
The FFN retains a nonlinear transformation of each node's own state and a GNN aggregation branch over the undirected neighborhood.
A learned gate combines these two representations without separate predecessor and successor aggregation branches.

\section{Additional Mechanistic Diagnostics}\label{app:additional-mechanistic-diagnostics}

These diagnostics examine how the proposed modules encode edge orientation, adapt directional context aggregation across node roles, and distribute attention over complementary structural relations.
The mechanistic diagnostics use 172 training architectures for NAS-Bench-101 and 469 for NAS-Bench-201.
The accompanying phase-parameter experiment uses the lowest training budget.
The mechanistic diagnostics characterize internal module behavior and complement the main results and ablations rather than directly attribute performance gains.
The magnetic phase diagnostic checks operator-level orientation sensitivity, while the MoE and attention diagnostics describe task-dependent routing and attention allocation over predefined relation channels.
The following subsections provide the definitions, numerical results, and per-module views corresponding to Fig.~\ref{fig:mechanism_diagnostics}.

In Fig.~\ref{fig:mechanism_diagnostics}, NB101/NB201 denote NAS-Bench-101/NAS-Bench-201, and Acc/Lat denote accuracy/latency.
The notation @172 or @469 specifies the number of architectures used to train the predictor.
In panel (b), Branch. and Converg. denote branching and converging node roles.
In panel (c), Fwd and Bwd denote forward and backward directions, reach denotes reachability, and Conv. and Branch denote relations defined by shared successors and shared predecessors, respectively.

\subsection{Magnetic Phase Sensitivity and Parameter Analysis}\label{app:magnetic-phase-analysis}

\paragraph{Graph-scaled phase.}
As specified in Appendix~\ref{app:maglap-settings}, the spectral encoder and the edge-reversal diagnostic use the degree-normalized magnetic Laplacian with a graph-scaled phase parameter.
In the reported graph-scaled setting, the shared parameter $q$ is converted to the effective phase parameter $q_G=q/c_G$, where $c_G=\max\{1,\min(m,n)\}$ and $n$ and $m$ are the numbers of nodes and edges in the encoded graph.
For a directed edge $u\to v$, substituting $q_G$ into Eq.~\eqref{eq:maglap} gives conjugate phase factors $e^{i2\pi q_G}$ and $e^{-i2\pi q_G}$, whose separation is
\begin{equation}
\left|e^{i2\pi q_G}-e^{-i2\pi q_G}\right|=2\left|\sin(2\pi q_G)\right|.
\label{eq:magnetic-phase-separation}
\end{equation}
The factors are $+i$ and $-i$ when the effective parameter $q_G=0.25$.
With graph scaling, the nominal setting $q=0.25$ generally gives $q_G\ne0.25$, so phase separation depends on both $q$ and the graph scale $c_G$.

\paragraph{Edge-reversal diagnostic.}
We quantify sensitivity to reversing all edges by
\begin{equation}
S_q(G)=\frac{\|\widetilde{L}_q(A)-\widetilde{L}_q(A^\top)\|_F}{\|\widetilde{L}_q(A)\|_F}.
\label{eq:magnetic-reversal-sensitivity}
\end{equation}
Here, $\widetilde{L}_q$ denotes the normalized magnetic Laplacian constructed with the effective phase $q_G$ defined above.
Larger values indicate stronger sensitivity to edge reversal.
At $q=0$, removing the phase term gives zero edge-reversal sensitivity in all four settings.
At $q=0.25$, the reported mean sensitivity is $0.2398$ on NAS-Bench-101 and $0.2065$ on NAS-Bench-201 for both prediction targets.
The variation across NAS-Bench-101 graphs reflects their different topologies, while the fixed NAS-Bench-201 encoding topology gives a constant response across operation assignments.

\begin{figure}[H]
    \centering
    \includegraphics[width=0.792\linewidth]{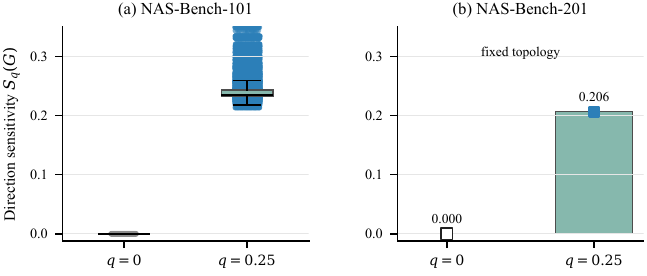}
    \caption{Magnetic spectral response to reversing all edge directions.
    NAS-Bench-101 yields a distribution over topology-diverse cells, whereas NAS-Bench-201 has a fixed encoding topology and therefore produces a constant response across operation assignments.}
    \label{fig:mechanism_magnetic_detail}
\end{figure}

\paragraph{Prediction sensitivity to $q$.}
We also evaluate $q\in\{0,0.125,0.25,0.5\}$ under the lowest training budget of each task, keeping all other parameters at the shared configuration.
Table~\ref{tab:hyperparameter-q} reports means over five runs.
This experiment evaluates the predictive effect of the nominal phase parameter, complementing the operator-level edge-reversal diagnostic above.

\begin{table}[H]
    \centering
    \caption{Effect of the magnetic phase parameter $q$ under the lowest training budget of each task.
    The tabulated $q$ is the shared nominal parameter before graph scaling.
    All other parameters are fixed at the shared configuration.
    Bold indicates the best result for each metric among the tested values.}
    \label{tab:hyperparameter-q}
    \small
    \setlength{\tabcolsep}{5.0pt}
    \begin{tabular}{c cc cc cc cc}
        \toprule
        \multirow[c]{2}{*}[-\dimexpr(\aboverulesep+\belowrulesep+\cmidrulewidth)/2\relax]{$q$} & \multicolumn{2}{c}{NB101 Acc @100}
        & \multicolumn{2}{c}{NB101 Lat @100}
        & \multicolumn{2}{c}{NB201 Acc @156}
        & \multicolumn{2}{c}{NB201 Lat @156} \\
        \cmidrule(lr){2-3}\cmidrule(lr){4-5}\cmidrule(lr){6-7}\cmidrule(lr){8-9}
        & KT $\uparrow$ & MAPE $\downarrow$ & KT $\uparrow$ & MAPE $\downarrow$
        & KT $\uparrow$ & MAPE $\downarrow$ & KT $\uparrow$ & MAPE $\downarrow$ \\
        \midrule
        $0$     & 0.6894 & 0.0341 & 0.8615 & 0.0806 & 0.7148 & 0.0792 & 0.8192 & 0.0325 \\
        $0.125$ & 0.6719 & 0.0330 & 0.8642 & 0.0767 & 0.7343 & 0.0528 & 0.8238 & 0.0325 \\
        $0.25$  & \textbf{0.6954} & \textbf{0.0327} & \textbf{0.8646} & \textbf{0.0758} & \textbf{0.8015} & \textbf{0.0244} & \textbf{0.8375} & \textbf{0.0305} \\
        $0.5$   & 0.6737 & 0.0337 & 0.8633 & 0.0766 & 0.7386 & 0.0461 & 0.8235 & 0.0323 \\
        \bottomrule
    \end{tabular}
\end{table}

The shared setting $q=0.25$ achieves the highest Kendall's $\tau$ and the lowest MAPE in all four evaluated tasks among the tested values.
These results support its predictive effectiveness within the evaluated settings, while the phase-separation expression above makes its graph-dependent directional effect explicit.

\subsection{Detailed MoE Routing by Node Role}
We group nodes by directed structural role and summarize the learned weights of the self, predecessor, and successor experts.
For source and branching nodes, the role-aligned contrast is $g_{\mathrm{succ}}-g_{\mathrm{pred}}$; for converging and sink nodes, it is $g_{\mathrm{pred}}-g_{\mathrm{succ}}$.
A positive contrast indicates greater weight on the context aligned with that role.

\begin{table}[H]
    \centering
    \caption{Mean role-aligned MoE routing contrasts.
    Positive values indicate greater successor-context weight for source and branching nodes and greater predecessor-context weight for converging and sink nodes.}
    \label{tab:moe-role-contrasts}
    \small
    \setlength{\tabcolsep}{8pt}
    \begin{tabular}{lrrrr}
        \toprule
        Setting & Source & Branching & Converging & Sink \\
        \midrule
        NB101 Acc @172 & $-0.0063$ & $+0.0002$ & $+0.0093$ & $+0.0156$ \\
        NB101 Lat @172 & $+0.0094$ & $+0.0192$ & $-0.0096$ & $+0.0014$ \\
        NB201 Acc @469 & $+0.0342$ & $+0.0371$ & $+0.0152$ & $+0.0439$ \\
        NB201 Lat @469 & $-0.0051$ & $-0.0039$ & $+0.0134$ & $+0.0149$ \\
        \bottomrule
    \end{tabular}
\end{table}

NAS-Bench-201 accuracy prediction exhibits positive role-aligned contrasts for all four roles.
The contrasts are $0.0342$, $0.0371$, $0.0152$, and $0.0439$ for source, branching, converging, and sink nodes, respectively.
Across the remaining settings, the signs and magnitudes vary with the prediction target and node role.
These patterns illustrate how the learned router adapts the balance between predecessor and successor context across tasks and structural roles.

\begin{figure}[H]
    \centering
    \includegraphics[width=0.9\linewidth]{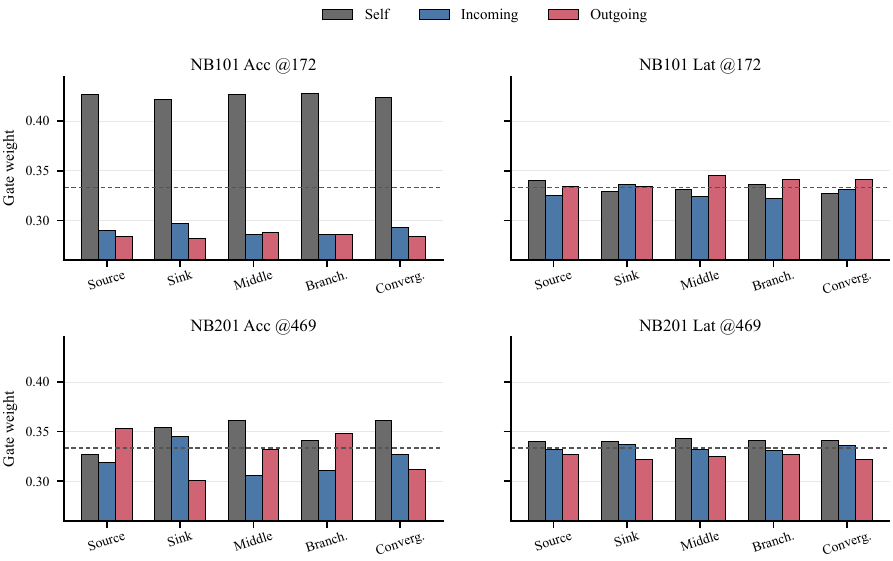}
    \caption{Learned MoE gate weights by directed node role across the four tasks.
    Incoming/outgoing denote predecessor/successor contexts; dashed lines mark uniform gate weights of $1/3$.}
    \label{fig:mechanism_moe_gate_weights}
\end{figure}

\subsection{Detailed Relation-Channel Statistics}
For each directional relation $r$, Figure~\ref{fig:mechanism_attention_detail} reports its density, the attention mass assigned to the node pairs covered by that relation, and density-normalized enrichment.
Relation density is the fraction of off-diagonal node pairs in the encoded graphs covered by the binary relation indicator $M_r$, and attention mass is the fraction of off-diagonal attention assigned to those pairs.
Enrichment is attention mass divided by relation density, averaged across evaluated runs, layers, and heads.
An enrichment value of $1$ corresponds to attention allocation proportional to relation coverage, placing local edges, directed reachability, and shared-neighbor relations on a common coverage-adjusted scale.

Across the four tasks, the six relation channels exhibit distinct enrichment patterns.
On NAS-Bench-101, forward and backward edge relations have enrichment values of approximately $1.89$--$1.91$ and $1.78$--$1.81$, respectively, while the directed-reachability relations are close to $0.98$.
The shared-successor and shared-predecessor relations reach approximately $2.20$ and $2.70$.
On NAS-Bench-201, the edge relations remain enriched, directed reachability is close to $1.04$, and the shared-predecessor relation reaches $4.67$.
These statistics characterize how the learned attention weights distribute mass over the local, multi-hop, and branch-related structural supports supplied to the six directional channels.

\begin{figure}[!htbp]
    \centering
    \includegraphics[width=0.9\linewidth]{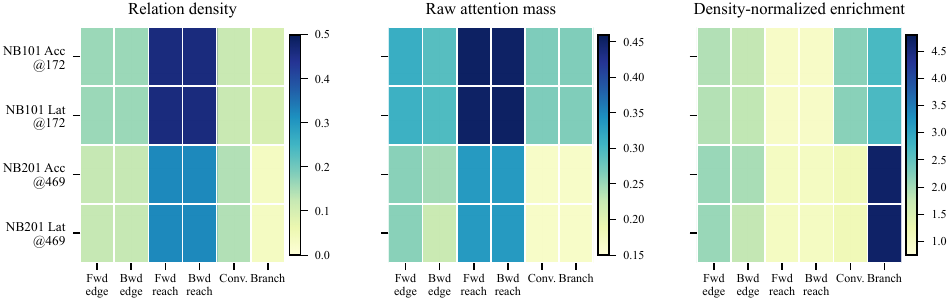}
    \caption{Attention allocation over the six directional relation channels across the four tasks.
    Relation density is the fraction of valid node pairs covered by a relation, attention mass is the fraction of attention assigned to those pairs, and enrichment is their ratio.}
    \label{fig:mechanism_attention_detail}
\end{figure}

\FloatBarrier

\section{Complexity Analysis}\label{app:complexity-analysis}

Let $n$ and $m$ denote the numbers of nodes and directed edges, respectively.
Let $d$ denote the hidden dimension, $k$ the number of retained spectral modes, and $L$ the number of encoder layers.
We treat the number of attention heads and the six directional relation channels as constants.

\subsection{Time Complexity}

Degrees, directed depths, and a sparse magnetic Laplacian can be constructed in $O(n+m)$ time from a sparse DAG representation.
A dense construction of reachability and common-neighbor relations has $O(n^3)$ worst-case time and $O(n^2)$ storage, and a full dense Hermitian eigendecomposition has the same worst-case time and storage orders.
These graph-dependent quantities are computed once and reused during subsequent encoder evaluations.

Aggregating $k$ spectral modes per node costs $O(nkd^2+nk^2d)$, which becomes $O(nd^2)$ for fixed $k$.
Each encoder layer uses $O(nd^2)$ work for feature projections and $O(n^2d)$ work for dense masked attention.
When adjacency is represented densely, the directional context updates remain within the same $O(n^2d+nd^2)$ per-layer bound.
The resulting encoder-evaluation cost is
\begin{equation}
T_{\mathrm{eval}}
=O\!\left(nkd^2+nk^2d+L(n^2d+nd^2)\right).
\end{equation}
The first-use cost adds graph-dependent construction to encoder evaluation:
\begin{equation}
T_{\mathrm{first}}
=T_{\mathrm{relations}}+T_{\mathrm{spectral}}+T_{\mathrm{features}}+T_{\mathrm{eval}}.
\end{equation}
With fixed $k$ and dense preprocessing, the conservative first-use bound is $O(n^3+L(n^2d+nd^2))$.

\subsection{Space Complexity}

Dense relation masks require $O(n^2)$ storage, retained eigenvectors require $O(nk)$, node states require $O(nd)$, and encoder parameters require $O(Ld^2)$.
For inference at batch size one, these terms give
\begin{equation}
M_{\mathrm{inference}}=O(n^2+nk+nd+Ld^2).
\end{equation}
A conventional backward pass that retains per-layer attention and node activations adds $O(L(n^2+nd))$ storage.
The measured GPU memory in Appendix~\ref{app:large-dag-efficiency} complements these asymptotic bounds for the profiled implementation.

\subsection{Design Choices for Practical Scalability}\label{app:scalability-design}

The analysis above identifies two scale-dependent costs: graph-dependent construction at first use and dense pairwise storage and computation during encoder evaluation.
DTA-GT addresses these costs through three complementary execution choices that preserve the same direction-aware representation.

\paragraph{Sparse retained-mode spectral computation.}
The large-graph spectral path represents the magnetic Laplacian sparsely and uses thick-restart Lanczos to compute only the retained eigenpairs through sparse matrix-vector products in a compact Krylov subspace.
The solver computes the spectral modes needed by the encoder without computing the full set of eigenvectors.
Its runtime depends on graph sparsity, solver tolerance, and the number of iterations needed for convergence.

\paragraph{Reusable graph-dependent preprocessing.}
The retained eigenpairs and directional relation masks depend only on the architecture graph, so they are constructed once and reused across encoder layers, training epochs, and repeated predictions of the same architecture.
This design separates construction of graph inputs from model evaluation and amortizes structural and spectral construction whenever an architecture is revisited.

\paragraph{Topology- and support-aware relation handling.}
Deployment-oriented architecture design balances predictive quality against computational cost, making sparse-to-moderately-dense connectivity a practically important regime for support-aware execution.
The practical motivation for support-aware relation handling is that computational graphs exhibit structured, architecture-dependent connectivity rather than a single density regime.

ResNet-style architectures introduce selected shortcut edges, whereas DenseNet-style blocks create a denser support pattern by connecting each layer to all preceding layers within a block.
Deriving relation supports from the observed topology preserves this variation and provides the structural basis for sparse execution whenever the graph admits it.

The six directional relations expose their supported node pairs explicitly, allowing relation construction to exploit known graph structure.
Let $S$ denote the total number of admissible query--key pairs across the six heads, including the self-pairs allowed in each head.
A support-indexed attention implementation evaluates only stored supported pairs, yielding an $O(Sd)$ attention term, while sparse predecessor and successor aggregation costs $O(md)$.
The explicit relation supports therefore provide the execution interface for support-proportional computation as graph scale increases.

Appendix~\ref{app:efficiency} reports predictor resource costs on NAS-Bench-201.
Appendix~\ref{app:larger-dags-accuracy} evaluates classification on MalNet-derived DAGs, and Appendix~\ref{app:large-dag-efficiency} profiles computational costs on real and generated DAGs.
The staged large-DAG profiles separately report costs of constructing graph inputs, model-forward latency, and peak forward GPU memory.

\section{Computational Efficiency}
\label{app:efficiency}

We construct DTA-GT-S by reducing the hidden dimension to 60 and the encoder depth to four while retaining the direction-aware design.
DTA-GT-S contains 0.36M parameters, compared with 0.61M for Graphormer and 6.99M for DTA-GT.
We compare their prediction quality under the same lowest-budget setting of each task.
Prediction results are averaged over five runs.

\begin{table}[H]
\centering
\caption{Parameter counts and inference measurements for the compact-model comparison.
DTA-GT-S retains the direction-aware design with reduced width and depth.}
\label{tab:compact_model_capacity}
\small
\setlength{\tabcolsep}{12pt}
\renewcommand{\arraystretch}{1.10}
\begin{tabular}{lrr}
\toprule
Model & Parameters (M) & Inference (ms/sample) \\
\midrule
Graphormer & 0.61 & 0.050 \\
DTA-GT-S ($d=60$, $L=4$) & \textbf{0.36} & \textbf{0.049} \\
\bottomrule
\end{tabular}
\end{table}

Table~\ref{tab:compact_model_quality_mean} in the main text reports the corresponding comparison of prediction quality.

\begin{table}[H]
\centering
\caption{Prediction quality of Graphormer and DTA-GT-S under the lowest training budget of each task.}
\label{tab:compact_model_quality_appendix}
\small
\setlength{\tabcolsep}{5.5pt}
\renewcommand{\arraystretch}{1.10}
\begin{tabular}{@{}lrrrr@{}}
\toprule
\multirow[c]{2}{*}[-\dimexpr(\aboverulesep+\belowrulesep+\cmidrulewidth)/2\relax]{Task} & \multicolumn{2}{c}{Graphormer} & \multicolumn{2}{c}{DTA-GT-S} \\
\cmidrule(lr){2-3}\cmidrule(lr){4-5}
 & KT $\uparrow$ & MAPE $\downarrow$ & KT $\uparrow$ & MAPE $\downarrow$ \\
\midrule
NB101 Acc @100 & 0.2229 & 0.0589 & \textbf{0.6603} & \textbf{0.0359} \\
NB101 Lat @100 & 0.3582 & 0.2862 & \textbf{0.8612} & \textbf{0.0777} \\
NB201 Acc @156 & 0.7052 & \textbf{0.0270} & \textbf{0.7114} & 0.0486 \\
NB201 Lat @156 & 0.6799 & 0.0651 & \textbf{0.8183} & \textbf{0.0371} \\
\bottomrule
\end{tabular}
\end{table}

DTA-GT-S uses $19.4\times$ fewer parameters than DTA-GT and remains smaller than Graphormer.
It achieves higher Kendall's $\tau$ than Graphormer in all four settings and lower MAPE in three of them.
Graphormer obtains the lower MAPE on NAS-Bench-201 accuracy prediction, while DTA-GT-S retains the higher ranking correlation in that setting.
The recorded inference times are $0.049$ ms per sample for DTA-GT-S and $0.050$ ms per sample for Graphormer.
These results show that the direction-aware design remains effective at a compact parameter scale.

We report parameter count, inference latency, peak GPU memory usage, and training time to characterize the computational cost of representative architecture predictors on NAS-Bench-201.
Latency is measured in milliseconds per architecture sample, and training time denotes the wall-clock time required to complete one training run.
For each method, latency, memory, and training time are summarized over four NAS-Bench-201 training budgets: 156, 469, 781, and 1563 training samples.
When a metric varies across budgets, we report its average value to obtain a compact method-level efficiency profile; for range-valued entries, the midpoint is used for aggregation.
The parameter count is reported once since it is independent of the training budget.

\begin{table}[H]
\centering
\caption{
Comparison of computational efficiency among architecture predictors.
Memory denotes peak GPU usage.
Latency, memory, and training time are summarized over four NAS-Bench-201 training budgets: 156, 469, 781, and 1563 training samples.
For range-valued entries, the midpoint is used.
}
\label{tab:app_efficiency}
\small
\setlength{\tabcolsep}{3pt}
\renewcommand{\arraystretch}{1.08}
\begin{tabular}{@{}lrrrr@{}}
\toprule
Method & Params (M) $\downarrow$ & Avg. latency (ms/sample) $\downarrow$ & Avg. memory (GB) $\downarrow$ & Avg. training time (h) $\downarrow$ \\
\midrule
NAO & 0.12 & 0.040 & 0.030 & 0.072 \\
ReNAS & 1.04 & 0.050 & 0.048 & 0.077 \\
NNLP & 0.64 & 0.050 & 0.040 & 0.083 \\
BRP-NAS & 1.09 & 0.050 & 0.062 & 0.089 \\
Graphormer & 0.61 & 0.050 & 0.055 & 0.101 \\
TNASP & 0.61 & 0.050 & 0.055 & 0.101 \\
GraphTrans & 0.93 & 0.053 & 0.073 & 0.123 \\
\textbf{DTA-GT} & \textbf{6.99} & \textbf{0.121} & \textbf{0.320} & \textbf{0.180} \\
NAR-Former V2 & 8.58 & 0.087 & 0.355 & 0.188 \\
NN-Former & 4.93 & 0.120 & 0.215 & 0.256 \\
NAR-Former & 4.82 & 0.135 & 0.180 & 0.284 \\
PINAT & 0.19 & 0.210 & 0.037 & 0.366 \\
\bottomrule
\end{tabular}
\end{table}

DTA-GT uses 6.99M parameters and has an average inference latency of 0.121 ms per sample under this NAS-Bench-201 profile.
Its average training time is 0.180 h, compared with 0.188 h for NAR-Former V2, 0.256 h for NN-Former, and 0.284 h for NAR-Former.
This resource profile complements the predictive results by quantifying the computational cost of the direction-aware design.

\section{Design Rationale and Parameter Sensitivity}
\label{app:design-sensitivity}
\label{app:hyperparameter-settings}

DTA-GT uses three structurally defined update paths for the node's own state, predecessor context, and successor context.
The number of experts follows this decomposition of directed computational context.
A node-dependent router learns adaptive mixing weights over the three contexts from the current node representation.
This allows each node to combine its local state with upstream and downstream information according to the prediction task.

The four directional stages provide complementary information throughout the encoder.
Node initialization represents local directed roles, and magnetic spectral encoding supplies a global orientation-sensitive prior.
Structural attention organizes interactions through directed node-pair relations.
The MoE-FFN adapts each node update to its self, predecessor, and successor contexts.
Together, these stages coordinate the modeling of directed computational semantics across representation construction, interaction, and feature updates.

Source and branching nodes distribute information along downstream paths, whereas converging and sink nodes summarize upstream computation.
These roles motivate separate predecessor and successor contexts, with the learned gate adapting their relative contributions at each node.
The routing diagnostics in Appendix~\ref{app:additional-mechanistic-diagnostics} examine how these weights vary across node roles and prediction targets.

The routing regularizer $\mathcal{L}_{\mathrm{aux}}$ in Section~\ref{sec:dir_ffn} is the negative entropy of the routing distribution; minimizing it with a positive auxiliary weight encourages higher entropy and discourages overly concentrated expert assignments.

Baseline sources and reproduction strategies are described in Appendix~\ref{app:experimental-details}.
The encoder blocks use standard layer normalization.
DTA-GT is optimized with AdamW using a batch size of 128, a learning rate of $1\times 10^{-4}$, and weight decay of $0.01$ for at most 3000 epochs.
For both accuracy and latency prediction, early stopping monitors validation Kendall's $\tau$ with a patience of 100, and evaluation uses the checkpoint with the highest validation $\tau$.
The learning rate is linearly warmed up over the first 10\% of training steps and then follows a cosine decay schedule.
The default DTA-GT model uses embedding dimension $d=180$, backbone depth $L=10$, and loss weights $\lambda_{\mathrm{mse}}=1$, $\lambda_{\mathrm{rank}}=0.2$, and $\lambda_{\mathrm{aux}}=0.001$.
We chose the default configuration in advance.
We used the model sizes and optimizer settings of NAR-Former~\citep{yi2023narformer}, NAR-Former V2~\citep{yi2023narformerv2}, and NN-Former~\citep{xu2025nnformer} as references.
We use the same configuration across all tasks and budgets without task- or budget-specific tuning.

The shared configuration is $(q,k,L,d)=(0.25,4,10,180)$ for both benchmarks, both prediction targets, and all reported NAS training budgets.
The $q$ sensitivity experiment is reported alongside the magnetic mechanism analysis in Appendix~\ref{app:magnetic-phase-analysis} and Table~\ref{tab:hyperparameter-q}.
That experiment and the $k$, $L$, and $d$ experiments below assess sensitivity around the configuration fixed a priori; we retain the same defaults for the main comparisons without task- or budget-specific parameter selection.
Each experiment changes one parameter while keeping the remaining settings fixed.
We evaluate $k\in\{1,2,4\}$, $L\in\{4,7,10,12\}$, and $d\in\{150,180,210\}$ under the lowest training budget of each task.
The tables below report means over five runs.
Bold indicates the best result for each metric among the tested values.

\subsection{Retained Spectral Modes}
After decomposing the magnetic Laplacian, we order the eigenvectors by ascending eigenvalue and retain the first $k$ low-frequency modes, giving $\mathbf{P}_{\mathrm{mag}}\in\mathbb{C}^{n\times k}$.
Each retained mode contributes real and imaginary channels that preserve its orientation-dependent phase information, and the spectral aggregation module combines the resulting $k$ components into the node representation.
The parameter $k$ therefore controls the resolution of the global spectral basis together with its storage and aggregation cost.
We use $k=4$ as a compact basis for the evaluated architecture graphs and compare it with $k=1$ and $k=2$ below.

\begin{table}[H]
    \centering
    \caption{Effect of the number of retained magnetic spectral modes $k$ under the lowest training budget of each task.
    All other parameters are fixed at the shared configuration.
    Bold indicates the best result for each metric among the tested values.}
    \label{tab:hyperparameter-k}
    \small
    \setlength{\tabcolsep}{5.0pt}
    \begin{tabular}{c cc cc cc cc}
        \toprule
        \multirow[c]{2}{*}[-\dimexpr(\aboverulesep+\belowrulesep+\cmidrulewidth)/2\relax]{$k$} & \multicolumn{2}{c}{NB101 Acc @100}
        & \multicolumn{2}{c}{NB101 Lat @100}
        & \multicolumn{2}{c}{NB201 Acc @156}
        & \multicolumn{2}{c}{NB201 Lat @156} \\
        \cmidrule(lr){2-3}\cmidrule(lr){4-5}\cmidrule(lr){6-7}\cmidrule(lr){8-9}
        & KT $\uparrow$ & MAPE $\downarrow$ & KT $\uparrow$ & MAPE $\downarrow$
        & KT $\uparrow$ & MAPE $\downarrow$ & KT $\uparrow$ & MAPE $\downarrow$ \\
        \midrule
        $1$ & 0.6785 & 0.0337 & 0.8606 & 0.0795 & 0.7383 & 0.0481 & 0.8159 & 0.0343 \\
        $2$ & 0.6831 & 0.0335 & 0.8625 & 0.0781 & 0.7406 & 0.0429 & 0.8199 & 0.0353 \\
        $4$ & \textbf{0.6954} & \textbf{0.0327} & \textbf{0.8646} & \textbf{0.0758} & \textbf{0.8015} & \textbf{0.0244} & \textbf{0.8375} & \textbf{0.0305} \\
        \bottomrule
    \end{tabular}
\end{table}

Retaining four spectral modes achieves the highest Kendall's $\tau$ and the lowest MAPE in all four evaluated tasks among the tested values of $k$.
For NAS-Bench-201 accuracy prediction, increasing $k$ from $1$ to $4$ changes Kendall's $\tau$ from $0.7383$ to $0.8015$ and MAPE from $0.0481$ to $0.0244$.
These results support the four-mode representation as the shared compact spectral basis for the evaluated benchmarks and prediction targets.

\subsection{Model Depth and Width}
The encoder depth $L$ and hidden dimension $d$ form the two standard model-scaling axes considered in our controlled sweeps.
The former changes the number of stacked directional encoder blocks, while the latter changes the width of node states and learnable projections.
Both parameters affect representational capacity and computation, and we vary them one at a time while keeping the remaining configuration fixed.

\paragraph{Depth.}
Each DTA-GT encoder layer composes direction-dual structural attention with the node-adaptive self, predecessor, and successor context update.
The depth $L$ therefore determines how many times pairwise structural interaction and direction-sensitive feature updating are repeated before graph readout.
Increasing $L$ expands the sequence of directional transformations as well as the corresponding computation.
We use $L=10$ in the shared configuration and compare it with $L\in\{4,7,12\}$ below.

\begin{table}[H]
    \centering
    \caption{Effect of encoder depth $L$ under the lowest training budget of each task.
    All other parameters are fixed at the shared configuration.
    Bold indicates the best result for each metric among the tested values.}
    \label{tab:hyperparameter-L}
    \small
    \setlength{\tabcolsep}{5.0pt}
    \begin{tabular}{c cc cc cc cc}
        \toprule
        \multirow[c]{2}{*}[-\dimexpr(\aboverulesep+\belowrulesep+\cmidrulewidth)/2\relax]{$L$} & \multicolumn{2}{c}{NB101 Acc @100}
        & \multicolumn{2}{c}{NB101 Lat @100}
        & \multicolumn{2}{c}{NB201 Acc @156}
        & \multicolumn{2}{c}{NB201 Lat @156} \\
        \cmidrule(lr){2-3}\cmidrule(lr){4-5}\cmidrule(lr){6-7}\cmidrule(lr){8-9}
        & KT $\uparrow$ & MAPE $\downarrow$ & KT $\uparrow$ & MAPE $\downarrow$
        & KT $\uparrow$ & MAPE $\downarrow$ & KT $\uparrow$ & MAPE $\downarrow$ \\
        \midrule
        $4$  & 0.6662 & 0.0409 & 0.8623 & 0.0803 & 0.7303 & 0.0645 & 0.8219 & 0.0346 \\
        $7$  & 0.6731 & 0.0368 & 0.8629 & 0.0796 & 0.7420 & 0.0449 & 0.8238 & 0.0335 \\
        $10$ & \textbf{0.6954} & \textbf{0.0327} & \textbf{0.8646} & \textbf{0.0758} & \textbf{0.8015} & \textbf{0.0244} & \textbf{0.8375} & \textbf{0.0305} \\
        $12$ & 0.6765 & 0.0333 & 0.8603 & 0.0827 & 0.7387 & 0.0437 & 0.8226 & 0.0332 \\
        \bottomrule
    \end{tabular}
\end{table}

Increasing the encoder depth from $4$ to $10$ improves Kendall's $\tau$ and reduces MAPE in all four evaluated tasks.
At $L=12$, Kendall's $\tau$ decreases and MAPE increases in all four tasks relative to $L=10$, showing that additional depth does not continue the improvement observed up to ten layers.
The shared ten-layer configuration therefore gives the best predictive results among the evaluated depths across both benchmarks and prediction targets.

\paragraph{Width.}
The hidden dimension $d$ sets the width of the node states and the attention and directional-context projections throughout the encoder.
We use $d=180$ in the shared configuration and compare it with $d=150$ and $d=210$ below.

\begin{table}[H]
    \centering
    \caption{Effect of hidden dimension $d$ under the lowest training budget of each task.
    All other parameters are fixed at the shared configuration.
    Bold indicates the best result for each metric among the tested values.}
    \label{tab:hyperparameter-d}
    \small
    \setlength{\tabcolsep}{5.0pt}
    \begin{tabular}{c cc cc cc cc}
        \toprule
        \multirow[c]{2}{*}[-\dimexpr(\aboverulesep+\belowrulesep+\cmidrulewidth)/2\relax]{$d$} & \multicolumn{2}{c}{NB101 Acc @100}
        & \multicolumn{2}{c}{NB101 Lat @100}
        & \multicolumn{2}{c}{NB201 Acc @156}
        & \multicolumn{2}{c}{NB201 Lat @156} \\
        \cmidrule(lr){2-3}\cmidrule(lr){4-5}\cmidrule(lr){6-7}\cmidrule(lr){8-9}
        & KT $\uparrow$ & MAPE $\downarrow$ & KT $\uparrow$ & MAPE $\downarrow$
        & KT $\uparrow$ & MAPE $\downarrow$ & KT $\uparrow$ & MAPE $\downarrow$ \\
        \midrule
        $150$ & 0.6793 & 0.0337 & 0.8643 & 0.0769 & 0.7430 & 0.0486 & 0.8244 & 0.0336 \\
        $180$ & \textbf{0.6954} & \textbf{0.0327} & \textbf{0.8646} & \textbf{0.0758} & \textbf{0.8015} & \textbf{0.0244} & \textbf{0.8375} & \textbf{0.0305} \\
        $210$ & 0.6734 & 0.0332 & 0.8583 & 0.0804 & 0.7514 & 0.0444 & 0.8185 & 0.0337 \\
        \bottomrule
    \end{tabular}
\end{table}

The shared width $d=180$ achieves the highest Kendall's $\tau$ and the lowest MAPE in all four evaluated tasks.
Increasing the width to $d=210$ lowers Kendall's $\tau$ and increases MAPE in all four tasks relative to $d=180$.
Across the evaluated settings, $d=180$ therefore provides the strongest overall predictive performance among the tested widths.

\section{Evaluation on Larger DAGs}\label{app:larger-dags}

\subsection{Predictive Accuracy on MalNet-Derived DAGs}\label{app:larger-dags-accuracy}
We further evaluate the direction-aware encoder on larger DAGs derived from the five-class MalNet-Tiny graph-classification dataset~\citep{freitas2021malnet}.
The derived graphs contain 5--4,993 nodes, with a mean of 1,407 nodes.
This experiment examines the effectiveness of the directional design on larger graphs under limited training supervision.

We contract each strongly connected component into a node to obtain a condensation DAG.
Each node receives a shared operation token, so the compared predictors must derive their graph representations from the structure of the resulting DAG.
All compared methods use the same derived graphs and data partitions.

We use a stratified 5\%/5\%/90\% training/validation/test split and fit each model on the 5\% training partition.
All classification metrics are computed on the held-out 90\% test partition.
The five classes contain equal numbers of graphs, and stratification preserves their proportions in each partition.
We compare DTA-GT with BRP-NAS, Graphormer, and NN-Former and report accuracy, macro-F1, macro-precision, and macro-recall.
Table~\ref{tab:malnet_accuracy_mean} reports the mean over five runs.

\begin{table}[H]
\centering
\caption{Classification results on DAGs derived from MalNet-Tiny using a stratified 5\%/5\%/90\% training/validation/test split.
All methods use the same derived graphs and partitions.
Values are mean percentages.}
\label{tab:malnet_accuracy_mean}
\setlength{\tabcolsep}{7pt}
\renewcommand{\arraystretch}{1.12}
\begin{tabular}{lrrrr}
\toprule
Model & Accuracy & Macro-F1 & Macro-Precision & Macro-Recall \\
\midrule
BRP-NAS & 57.0 & 53.1 & 56.9 & 57.0 \\
Graphormer & 69.5 & 67.9 & 70.9 & 69.5 \\
NN-Former & 59.3 & 58.8 & 62.4 & 59.3 \\
\textbf{DTA-GT} & \textbf{74.1} & \textbf{73.8} & \textbf{74.8} & \textbf{74.1} \\
\bottomrule
\end{tabular}
\end{table}

DTA-GT achieves 74.1\% accuracy and 73.8\% macro-F1, leading the compared methods on all four metrics.
Its accuracy exceeds Graphormer by 4.6 percentage points and BRP-NAS by 17.1 percentage points.
With a shared node token, these results support the effectiveness of the direction- and topology-aware representation on substantially larger DAGs.

\subsection{Computational Efficiency on Large DAGs}\label{app:large-dag-efficiency}
We profile representative MalNet-derived DAGs and generated DAGs on an NVIDIA A800-SXM4 GPU with 80 GB memory.
The real-graph profiles range from 16 nodes and 25 edges to 3,416 nodes and 9,215 edges, while the generated profiles range from 8 nodes and 16 edges to 32,768 nodes and 65,536 edges.
The real graphs have median edge density below 0.3\%.
All methods use identical graphs and reported model settings with batch size one.
The real-DAG comparison reports the graph preprocessing required by each method followed by one forward pass through the model.
For generated DAGs, we report each method's measured prediction latency.
The DTA-GT values include only the model forward pass and exclude graph-input construction.
For the largest generated DAG, we also report a separate breakdown of the first-use cost, including graph-input construction.
The real-DAG DTA-GT stage profile uses ten warm-up and thirty timed measurements for construction of graph inputs, followed by thirty warm-up and five hundred timed measurements for model forward.
The independently instrumented profile on the largest generated DAG uses one warm-up and five timed measurements for each construction stage, followed by thirty warm-up and fifty timed measurements for model forward.

The full-size DTA-GT predictor uses a hidden dimension of $d=180$ in the main NAS prediction experiments.
For the large-DAG latency and memory measurements in this section, we use $d=192$ with ten encoder layers, six attention heads, four retained spectral modes, and a nominal phase parameter of $q=0.25$.
Graph-dependent structural masks and spectral features are constructed once per graph and shared by all encoder layers.
For larger graphs, we represent the magnetic Laplacian in sparse CSR form and compute only the retained eigenpairs using thick-restart Lanczos~\citep{wu2000thickrestart}; smaller graphs use dense eigendecomposition.
We report construction of structural masks, magnetic spectral construction, and model forward separately in Table~\ref{tab:large_dag_dtagt_breakdown}.

\begin{table}[H]
\centering
\caption{DTA-GT stage latencies on representative MalNet-derived DAGs. Graph inputs are constructed before model forward. Columns identify node and edge counts, and entries report mean latency in milliseconds.}
\label{tab:large_dag_dtagt_breakdown}
\small
\setlength{\tabcolsep}{3pt}
\renewcommand{\arraystretch}{1.10}
\begin{tabular}{@{}lrrrrr@{}}
\toprule
Component & 16--25 & 88--180 & 362--859 & 1,427--3,844 & 3,416--9,215 \\
\midrule
Construction of structural masks & 0.39 & 0.39 & 0.39 & 1.25 & 11.46 \\
Magnetic spectral construction & 0.64 & 2.89 & 6.15 & 7.33 & 7.48 \\
Model forward & 5.19 & 5.30 & 5.52 & 12.13 & 40.01 \\
\bottomrule
\end{tabular}
\end{table}

On the 3,416-node DAG, construction of structural masks, spectral construction, and model forward take 11.46 ms, 7.48 ms, and 40.01 ms, respectively.
The sum of these reported stages is 58.95 ms.
For repeated predictions of the same architecture, the constructed graph inputs can be reused and only the model-forward stage is executed again.

Table~\ref{tab:large_dag_real_latency} compares end-to-end method-specific preprocessing and inference latency on representative MalNet-derived DAGs.
On the largest profiled real DAG, DTA-GT requires 58.95 ms, compared with 122.56 ms for PINAT and 161.20 ms for NN-Former.
BRP-NAS requires 4.24 ms on the same graph, showing the lower computational cost of the GCN predictor in this profile.

\begin{table}[H]
\centering
\caption{End-to-end latency on representative MalNet-derived DAGs. Each method includes its required graph preprocessing and one inference pass. Values are mean milliseconds per graph on the A800, and columns identify node and edge counts.}
\label{tab:large_dag_real_latency}
\small
\setlength{\tabcolsep}{3pt}
\renewcommand{\arraystretch}{1.10}
\begin{tabular}{@{}lrrrrr@{}}
\toprule
Method & 16--25 & 88--180 & 362--859 & 1,427--3,844 & 3,416--9,215 \\
\midrule
BRP-NAS & \textbf{0.55} & \textbf{0.55} & \textbf{0.55} & \textbf{1.35} & \textbf{4.24} \\
PINAT & 10.74 & 10.65 & 10.52 & 29.97 & 122.56 \\
NAR-Former & 43.50 & 238.11 & 1,339.62 & 11,612.44 & 55,393.56 \\
NAR-Former V2 & 8.58 & 30.27 & 119.67 & 449.76 & 1,185.62 \\
NN-Former & 7.29 & 7.27 & 7.44 & 22.71 & 161.20 \\
\textbf{DTA-GT} & 6.22 & 8.58 & 12.06 & \underline{20.71} & \underline{58.95} \\
\bottomrule
\end{tabular}
\end{table}

Table~\ref{tab:large_dag_generated_latency} extends the profile to generated DAGs with up to 32,768 nodes and 65,536 edges.
DTA-GT completes model forward after construction of graph inputs on the largest generated DAG in 3,125.12 ms and reaches 42.249 GiB of peak GPU memory.
PINAT and NAR-Former V2 record 9,684.47 ms and 12,036.15 ms, respectively, on their profiled prediction paths.
NAR-Former and NN-Former encounter out-of-memory failures at this graph size under their reported settings, while BRP-NAS records 313.80 ms.

\begin{table}[H]
\centering
\caption{Recorded prediction-path latency on generated DAGs.
The DTA-GT row reports model forward after construction of graph inputs; Table~\ref{tab:large_dag_generated_breakdown} reports its complete first-use stage sum on the largest graph.
Values are mean milliseconds per graph on the A800.
OOM denotes an out-of-memory failure under the reported settings.}
\label{tab:large_dag_generated_latency}
\small
\setlength{\tabcolsep}{3pt}
\renewcommand{\arraystretch}{1.10}
\begin{tabular}{@{}lrrrrr@{}}
\toprule
Method & 8--16 & 64--128 & 512--1,024 & 4,096--8,192 & 32,768--65,536 \\
\midrule
BRP-NAS & 0.56 & 0.54 & 0.58 & 6.08 & 313.80 \\
PINAT & 10.66 & 10.55 & 10.58 & 191.52 & 9,684.47 \\
NAR-Former & 28.23 & 166.74 & 2,248.00 & 78,731.93 & OOM \\
NAR-Former V2 & 6.01 & 23.96 & 161.23 & 1,298.03 & 12,036.15 \\
NN-Former & 7.43 & 7.24 & 7.76 & 268.80 & OOM \\
DTA-GT (forward) & 5.20 & 5.21 & 5.41 & 61.08 & 3,125.12 \\
\bottomrule
\end{tabular}
\end{table}

Table~\ref{tab:large_dag_generated_breakdown} independently profiles four DTA-GT stages on the largest generated DAG.
Model forward after construction of graph inputs takes 3,125.12 ms and reaches 42.249 GiB of peak GPU memory.
The independently measured means for feature preprocessing, spectral construction, construction of structural masks, and model forward sum to 25,601.083 ms for first use.
The stage breakdown shows that graph-dependent preprocessing is a substantial part of the first-use cost at this scale.

\begin{table}[H]
\centering
\caption{Staged DTA-GT profile for the generated DAG with 32,768 nodes and 65,536 edges. Stage means are measured independently, and the first-use value is their arithmetic sum. Peak GPU memory refers to model forward.}
\label{tab:large_dag_generated_breakdown}
\setlength{\tabcolsep}{10pt}
\renewcommand{\arraystretch}{1.12}
\begin{tabular}{lr}
\toprule
Stage or resource & Recorded value \\
\midrule
Feature preprocessing & 14,616.178 ms \\
Magnetic spectral construction & 9.923 ms \\
Construction of structural masks & 7,849.865 ms \\
Model forward & 3,125.117 ms \\
First-use stage sum & 25,601.083 ms \\
Peak forward GPU memory & 42.249 GiB \\
\bottomrule
\end{tabular}
\end{table}

For generated DAGs containing a spanning directed chain, we additionally evaluate a topology-specific relation-construction procedure motivated by the support-aware design discussed in Appendix~\ref{app:scalability-design}.
On the 32,768-node graph, this procedure constructs the six masks in 33.340 ms, and the resulting dense Boolean masks occupy 6 GiB.
This result separates time required to construct relations from relation storage and provides measured evidence for topology-aware relation handling.

The MalNet-derived DAG experiments show that DTA-GT achieves the highest accuracy and macro-F1 among the compared predictors, supporting the effectiveness of the complete encoder on the evaluated larger graphs.
These results assess the combined directional design at the tested graph scales and do not isolate the predictive contribution of magnetic spectral encoding.
As graph size increases, graph scaling reduces the orientation phase per edge, which may weaken the spectral directional signal in some large-graph regimes.
BRP-NAS has lower latency in the large-graph profiles.
The staged A800 measurements support the intended roles of retained-mode spectral computation, reusable graph-dependent preprocessing, and topology-aware relation construction within the evaluated range.
The reported scaling evidence covers graphs with up to 32,768 nodes; sparse or hierarchical attention and compact reachability representations remain important directions for substantially larger computational graphs.

\section{Extended Discussion}\label{app:limitations-impact}

\paragraph{Cross-space representation and adaptation.}
A node in a cell-based CNN space may denote a primitive convolution or pooling operation, whereas a node in another space may denote an entire Transformer layer or Mamba block.
These different computational granularities motivate a canonical operation-level representation for cross-space adaptation.
Composite blocks could be decomposed into shared semantic units or mapped to a common operation vocabulary with aligned edge definitions and target protocols.
For spaces with compatible operation definitions and dependency semantics, future work can investigate reusing DTA-GT's directional encoder and graph preprocessing, with new vocabulary entries and embeddings for additional operation types.
This would enable studies of cross-space pretraining followed by adaptation on a small evaluated subset from the target space.

\paragraph{Task adaptation and NAS evaluation.}
For a new DAG-level task with compatible atomic units and semantics of directed dependencies, adapting the prediction head and fine-tuning the encoder on a small labeled subset provides a natural extension.
This would allow the same direction-aware representation principle to be studied under different graph-level targets.
A complementary direction is to integrate the predictor into a NAS search loop and evaluate how candidate ranking affects architecture selection under a fixed end-to-end evaluation budget.

\paragraph{Relation-channel design and extension.}
The three paired relation families in Section~\ref{sec:struct_attn} are heuristic choices motivated by DAG semantics, providing a concrete instantiation of direction-dual structural attention.
They capture local dependencies, multi-hop reachability, and shared predecessor/successor structure.
Additional directional information, such as hop-specific dependencies or directed motif patterns, can be represented by further ordered or role-specific relation supports, with a corresponding extension of the relation-to-head assignment.
The relation set can therefore be tailored to the directional semantics available in a target graph domain.
Our experiments evaluate the six-channel instantiation; the predictive value and computational cost of expanded relation sets remain to be studied.

\paragraph{Computational scaling.}
Directed reachability can grow quadratically even when the original adjacency is sparse.
For example, a directed chain has $n-1$ edges but $n(n-1)/2$ reachable ordered pairs.
Dense relation masks therefore require quadratic storage, and dense masked attention incurs quadratic pairwise computation for a fixed hidden dimension.
The A800 profiles in Appendix~\ref{app:large-dag-efficiency} demonstrate execution on generated DAGs with up to 32,768 nodes.
Extending graph Transformer predictors to substantially larger computational graphs calls for sparse or hierarchical attention together with compact representations of directed reachability.
Combining these mechanisms with reusable spectral preprocessing could extend the encoder to larger computational graphs while retaining its direction-aware representation.
Future evaluations should measure prediction quality, construction of graph inputs on first use, latency of the model's forward pass, and GPU memory under matched graph and hardware settings.

\section{Complete Results with Standard Deviations}\label{app:complete-statistics}

This section reports five-run means and standard deviations for the main comparisons, ablations, and spectral controls in Tables~\ref{tab:main_results}--\ref{tab:direction_ablation}.

\subsection{NAS-Bench-101 and NAS-Bench-201 Prediction}\label{app:complete-main-results}
The complete results for all four prediction tasks are reported as means and standard deviations in Tables~\ref{tab:main_results_complete_nb101_acc}--\ref{tab:main_results_complete_nb201_lat}.

\begin{table}[H]
    \begingroup
    \def\mainresultsscope{3}
    \def\result#1#2{\shortstack{#1\\[-0.30ex]{\scriptsize $\pm$ #2}}}
    \def\mainresultsfontsize{\small}
    \def\samplehead#1#2{\shortstack{#1\%\\[-0.2ex]{\footnotesize(@#2)}}}
    \def\mainresultstabcolsep{2.2pt}
    \def\mainresultsarraystretch{1.05}
    \def\panelheading#1{}
    \def\resultssecondpanelheading#1{\panelheading{#1}}
    
    \endgroup
\end{table}

\begin{table}[H]
    \begingroup
    \def\mainresultsscope{4}
    \def\result#1#2{\shortstack{#1\\[-0.30ex]{\scriptsize $\pm$ #2}}}
    \def\mainresultsfontsize{\small}
    \def\samplehead#1#2{\shortstack{#1\%\\[-0.2ex]{\footnotesize(@#2)}}}
    \def\mainresultstabcolsep{2.2pt}
    \def\mainresultsarraystretch{1.05}
    \def\panelheading#1{}
    \def\resultssecondpanelheading#1{\panelheading{#1}}
    
    \endgroup
\end{table}

\begin{table}[H]
    \begingroup
    \def\mainresultsscope{5}
    \def\result#1#2{\shortstack{#1\\[-0.30ex]{\scriptsize $\pm$ #2}}}
    \def\mainresultsfontsize{\small}
    \def\samplehead#1#2{\shortstack{#1\%\\[-0.2ex]{\footnotesize(@#2)}}}
    \def\mainresultstabcolsep{2.2pt}
    \def\mainresultsarraystretch{1.05}
    \def\panelheading#1{}
    \def\resultssecondpanelheading#1{\panelheading{#1}}
    
    \endgroup
\end{table}

\begin{table}[H]
    \begingroup
    \def\mainresultsscope{6}
    \def\result#1#2{\shortstack{#1\\[-0.30ex]{\scriptsize $\pm$ #2}}}
    \def\mainresultsfontsize{\small}
    \def\samplehead#1#2{\shortstack{#1\%\\[-0.2ex]{\footnotesize(@#2)}}}
    \def\mainresultstabcolsep{2.2pt}
    \def\mainresultsarraystretch{1.05}
    \def\panelheading#1{}
    \def\resultssecondpanelheading#1{\panelheading{#1}}
    
    \endgroup
\end{table}

\subsection{Ablation Analysis}
Table~\ref{tab:ablation_complete} reports the uncertainty estimates corresponding to the mean-only ablation results in Table~\ref{tab:pipeline_consistency_ablation}.

\providecommand{\meanstd}[2]{\shortstack{#1\\[-0.30ex]{\scriptsize $\pm$ #2}}}
\providecommand{\bestmeanstd}[2]{\shortstack{\textbf{#1}\\[-0.30ex]{\scriptsize $\pm$ \textbf{#2}}}}

\begin{table}[H]
\centering
\caption{Complete ablation results under the lowest training budget of each task.
Values are mean $\pm$ standard deviation.}
\label{tab:ablation_complete}
\small
\setlength{\tabcolsep}{2.2pt}
\renewcommand{\arraystretch}{1.05}
\begin{tabular}{@{}l*{8}{c}@{}}
\toprule
\multirow[c]{3}{*}[-\dimexpr\aboverulesep+\belowrulesep+\cmidrulewidth\relax]{Variant} & \multicolumn{4}{c}{NAS-Bench-101} & \multicolumn{4}{c}{NAS-Bench-201} \\
\cmidrule(lr){2-5}\cmidrule(lr){6-9}
& \multicolumn{2}{c}{Accuracy} & \multicolumn{2}{c}{Latency} & \multicolumn{2}{c}{Accuracy} & \multicolumn{2}{c}{Latency} \\
\cmidrule(lr){2-3}\cmidrule(lr){4-5}\cmidrule(lr){6-7}\cmidrule(lr){8-9}
& KT $\uparrow$ & MAPE $\downarrow$ & KT $\uparrow$ & MAPE $\downarrow$ & KT $\uparrow$ & MAPE $\downarrow$ & KT $\uparrow$ & MAPE $\downarrow$ \\
\midrule
\rowcolor{gray!20} Directed DTA-GT & \bestmeanstd{0.6954}{0.0105} & \bestmeanstd{0.0327}{0.0001} & \bestmeanstd{0.8646}{0.0036} & \bestmeanstd{0.0758}{0.0009} & \bestmeanstd{0.8015}{0.0133} & \bestmeanstd{0.0244}{0.0105} & \bestmeanstd{0.8375}{0.0048} & \bestmeanstd{0.0305}{0.0007} \\
Undirected & \meanstd{0.5882}{0.0063} & \meanstd{0.0408}{0.0016} & \meanstd{0.8062}{0.0085} & \meanstd{0.1074}{0.0014} & \meanstd{0.7215}{0.0110} & \meanstd{0.0617}{0.0253} & \meanstd{0.8178}{0.0103} & \meanstd{0.0349}{0.0031} \\
w/o Directed Node Features & \meanstd{0.6522}{0.0236} & \meanstd{0.0361}{0.0019} & \meanstd{0.8482}{0.0056} & \meanstd{0.0908}{0.0058} & \meanstd{0.7198}{0.0205} & \meanstd{0.0729}{0.0140} & \meanstd{0.8202}{0.0050} & \meanstd{0.0326}{0.0016} \\
w/o Spec. Enc. & \meanstd{0.6894}{0.0016} & \meanstd{0.0341}{0.0015} & \meanstd{0.8615}{0.0008} & \meanstd{0.0806}{0.0031} & \meanstd{0.7452}{0.0089} & \meanstd{0.0600}{0.0055} & \meanstd{0.8192}{0.0003} & \meanstd{0.0325}{0.0009} \\
w/o Struct. Attn. & \meanstd{0.6552}{0.0144} & \meanstd{0.0331}{0.0007} & \meanstd{0.8360}{0.0202} & \meanstd{0.1033}{0.0017} & \meanstd{0.6123}{0.0339} & \meanstd{0.1037}{0.0087} & \meanstd{0.8227}{0.0033} & \meanstd{0.0317}{0.0029} \\
w/o MoE-FFN & \meanstd{0.6542}{0.0095} & \meanstd{0.0331}{0.0007} & \meanstd{0.8495}{0.0020} & \meanstd{0.0860}{0.0022} & \meanstd{0.7248}{0.0137} & \meanstd{0.0418}{0.0062} & \meanstd{0.8298}{0.0063} & \meanstd{0.0313}{0.0024} \\
CLS Readout & \meanstd{0.6701}{0.0101} & \meanstd{0.0399}{0.0048} & \meanstd{0.8420}{0.0047} & \meanstd{0.1203}{0.0096} & \meanstd{0.7535}{0.0141} & \meanstd{0.0704}{0.0103} & \meanstd{0.8209}{0.0030} & \meanstd{0.0348}{0.0012} \\
MSE Only & \meanstd{0.6720}{0.0109} & \meanstd{0.0356}{0.0022} & \meanstd{0.8610}{0.0056} & \meanstd{0.0808}{0.0060} & \meanstd{0.6937}{0.0065} & \meanstd{0.0686}{0.0006} & \meanstd{0.8201}{0.0021} & \meanstd{0.0321}{0.0007} \\
\bottomrule
\end{tabular}
\end{table}

\subsection{Magnetic Spectral Controls}
Table~\ref{tab:direction_ablation_complete} provides the available repeated-run statistics corresponding to the mean-only controls in Table~\ref{tab:direction_ablation}.
The ``Directed with Symmetric Lap.'' row reports the $q=0$ results from the five-run phase-parameter sensitivity experiment.

\providecommand{\meanstd}[2]{\shortstack{#1\\[-0.30ex]{\scriptsize $\pm$ #2}}}
\providecommand{\bestmeanstd}[2]{\shortstack{\textbf{#1}\\[-0.30ex]{\scriptsize $\pm$ \textbf{#2}}}}

\begin{table}[H]
\centering
\caption{Complete symmetric- and magnetic-Laplacian control results under the lowest training budget of each task.
Cells report mean $\pm$ standard deviation.}
\label{tab:direction_ablation_complete}
\small
\setlength{\tabcolsep}{2.2pt}
\renewcommand{\arraystretch}{1.08}
\begin{tabular}{@{}l*{8}{c}@{}}
\toprule
\multirow[c]{3}{*}[-\dimexpr\aboverulesep+\belowrulesep+\cmidrulewidth\relax]{Variant} & \multicolumn{4}{c}{NAS-Bench-101 @100} & \multicolumn{4}{c}{NAS-Bench-201 @156} \\
\cmidrule(lr){2-5}\cmidrule(lr){6-9}
& \multicolumn{2}{c}{Accuracy} & \multicolumn{2}{c}{Latency} & \multicolumn{2}{c}{Accuracy} & \multicolumn{2}{c}{Latency} \\
\cmidrule(lr){2-3}\cmidrule(lr){4-5}\cmidrule(lr){6-7}\cmidrule(lr){8-9}
& KT $\uparrow$ & MAPE $\downarrow$ & KT $\uparrow$ & MAPE $\downarrow$ & KT $\uparrow$ & MAPE $\downarrow$ & KT $\uparrow$ & MAPE $\downarrow$ \\
\midrule
No Directional Encoding & \meanstd{0.5882}{0.0063} & \meanstd{0.0408}{0.0016} & \meanstd{0.8062}{0.0085} & \meanstd{0.1074}{0.0014} & \meanstd{0.7215}{0.0110} & \meanstd{0.0617}{0.0253} & \meanstd{0.8178}{0.0103} & \meanstd{0.0349}{0.0031} \\
No Direction + Symmetric Lap. & \meanstd{0.5878}{0.0090} & \meanstd{0.0386}{0.0018} & \meanstd{0.8071}{0.0169} & \meanstd{0.1066}{0.0082} & \meanstd{0.7135}{0.0472} & \meanstd{0.0485}{0.0123} & \meanstd{0.8206}{0.0192} & \meanstd{0.0330}{0.0055} \\
Directed with Symmetric Lap. & \meanstd{0.6894}{0.0161} & \meanstd{0.0341}{0.0008} & \meanstd{0.8615}{0.0051} & \meanstd{0.0806}{0.0014} & \meanstd{0.7148}{0.0159} & \meanstd{0.0792}{0.0101} & \meanstd{0.8192}{0.0096} & \meanstd{0.0325}{0.0042} \\
Directed with Magnetic Lap. & \bestmeanstd{0.6954}{0.0105} & \bestmeanstd{0.0327}{0.0001} & \bestmeanstd{0.8646}{0.0036} & \bestmeanstd{0.0758}{0.0009} & \bestmeanstd{0.8015}{0.0133} & \bestmeanstd{0.0244}{0.0105} & \bestmeanstd{0.8375}{0.0048} & \bestmeanstd{0.0305}{0.0007} \\
\bottomrule
\end{tabular}
\end{table}

\endgroup

\newpage
\section*{NeurIPS Paper Checklist}

\begin{enumerate}

\item {\bf Claims}
    \item[] Question: Do the main claims made in the abstract and introduction accurately reflect the paper's contributions and scope?
    \item[] Answer: \answerYes{}
    \item[] Justification: The claims are scoped to the architecture-prediction settings evaluated in Section~\ref{Experiment}.

\item {\bf Limitations}
    \item[] Question: Does the paper discuss the limitations of the work performed by the authors?
    \item[] Answer: \answerYes{}
    \item[] Justification: Limitations are discussed in Section~\ref{sec:discussion} and Appendix~\ref{app:limitations-impact}.

\item {\bf Theory assumptions and proofs}
    \item[] Question: For each theoretical result, does the paper provide the full set of assumptions and a complete (and correct) proof?
    \item[] Answer: \answerNA{}
    \item[] Justification: The paper does not state formal theorems or lemmas.

\item {\bf Experimental result reproducibility}
    \item[] Question: Does the paper fully disclose all the information needed to reproduce the main experimental results of the paper to the extent that it affects the main claims and/or conclusions of the paper (regardless of whether the code and data are provided or not)?
    \item[] Answer: \answerYes{}
    \item[] Justification: Datasets, per-seed shared splits, training settings, and reproduction details are provided in Appendices~\ref{app:experimental-details}--\ref{app:hyperparameter-settings}.

\item {\bf Open access to data and code}
    \item[] Question: Does the paper provide open access to the data and code, with sufficient instructions to faithfully reproduce the main experimental results, as described in supplemental material?
    \item[] Answer: \answerYes{}
    \item[] Justification: The public code and data repository is linked in Appendix~\ref{app:code-data-access}.

\item {\bf Experimental setting/details}
    \item[] Question: Does the paper specify all the training and test details (e.g., data splits, hyperparameters, how they were chosen, type of optimizer) necessary to understand the results?
    \item[] Answer: \answerYes{}
    \item[] Justification: Training and evaluation settings are provided in Section~\ref{Experiment} and Appendices~\ref{app:experimental-details} and \ref{app:hyperparameter-settings}.

\item {\bf Experiment statistical significance}
    \item[] Question: Does the paper report error bars suitably and correctly defined or other appropriate information about the statistical significance of the experiments?
    \item[] Answer: \answerYes{}
    \item[] Justification: Appendix~\ref{app:complete-statistics} reports five-run means and standard deviations for Tables~\ref{tab:main_results}--\ref{tab:direction_ablation}, using the repeated-run protocol in Appendix~\ref{app:datasets-splits-budgets}.

\item {\bf Experiments compute resources}
    \item[] Question: For each experiment, does the paper provide sufficient information on the computer resources (type of compute workers, memory, time of execution) needed to reproduce the experiments?
    \item[] Answer: \answerNo{}
    \item[] Justification: Hardware and selected resource profiles are reported in Appendices~\ref{app:experimental-details}, \ref{app:efficiency}, and \ref{app:larger-dags}, but complete per-run and total compute accounting is not provided.

\item {\bf Code of ethics}
    \item[] Question: Does the research conducted in the paper conform, in every respect, with the NeurIPS Code of Ethics \url{https://neurips.cc/public/EthicsGuidelines}?
    \item[] Answer: \answerYes{}
    \item[] Justification: The experiments use public benchmarks and involve no research with human subjects or personal data.

\item {\bf Broader impacts}
    \item[] Question: Does the paper discuss both potential positive societal impacts and negative societal impacts of the work performed?
    \item[] Answer: \answerNA{}
    \item[] Justification: This is foundational graph-representation research on architecture benchmarks and MalNet-derived DAGs, with no direct societal deployment.

\item {\bf Safeguards}
    \item[] Question: Does the paper describe safeguards that have been put in place for responsible release of data or models that have a high risk for misuse (e.g., pre-trained language models, image generators, or scraped datasets)?
    \item[] Answer: \answerNA{}
    \item[] Justification: The work does not release assets with a high risk of misuse.

\item {\bf Licenses for existing assets}
    \item[] Question: Are the creators or original owners of assets (e.g., code, data, models), used in the paper, properly credited and are the license and terms of use explicitly mentioned and properly respected?
    \item[] Answer: \answerYes{}
    \item[] Justification: Existing assets are credited and used under applicable licenses or permissions; Table~\ref{tab:baseline_code_links} reports code sources and license or availability information.
    \item[] Guidelines:
    \begin{itemize}
        \item The answer \answerNA{} means that the paper does not use existing assets.
        \item The authors should cite the original paper that produced the code package or dataset.
        \item The authors should state which version of the asset is used and, if possible, include a URL.
        \item The name of the license (e.g., CC-BY 4.0) should be included for each asset.
        \item For scraped data from a particular source (e.g., website), the copyright and terms of service of that source should be provided.
        \item If assets are released, the license, copyright information, and terms of use in the package should be provided. For popular datasets, \href{https://paperswithcode.com/datasets}{\nolinkurl{paperswithcode.com/datasets}} has curated licenses for some datasets. Their licensing guide can help determine the license of a dataset.
        \item For existing datasets that are re-packaged, both the original license and the license of the derived asset (if it has changed) should be provided.
        \item If this information is not available online, the authors are encouraged to reach out to the asset's creators.
    \end{itemize}

\item {\bf New assets}
    \item[] Question: Are new assets introduced in the paper well documented and is the documentation provided alongside the assets?
    \item[] Answer: \answerYes{}
    \item[] Justification: Code and latency-label assets are linked in Appendix~\ref{app:code-data-access}, with reproduction details in Appendices~\ref{app:experimental-details}--\ref{app:hyperparameter-settings}.

\item {\bf Crowdsourcing and research with human subjects}
    \item[] Question: For crowdsourcing experiments and research with human subjects, does the paper include the full text of instructions given to participants and screenshots, if applicable, as well as details about compensation (if any)?
    \item[] Answer: \answerNA{}
    \item[] Justification: The work involves no crowdsourcing or research with human subjects.

\item {\bf Institutional review board (IRB) approvals or equivalent for research with human subjects}
    \item[] Question: Does the paper describe potential risks incurred by study participants, whether such risks were disclosed to the subjects, and whether Institutional Review Board (IRB) approvals (or an equivalent approval/review based on the requirements of your country or institution) were obtained?
    \item[] Answer: \answerNA{}
    \item[] Justification: No research with human subjects is conducted.

\item {\bf Declaration of LLM usage}
    \item[] Question: Does the paper describe the usage of LLMs if it is an important, original, or non-standard component of the core methods in this research? Note that if the LLM is used only for writing, editing, or formatting purposes and does \emph{not} impact the core methodology, scientific rigor, or originality of the research, declaration is not required.
    \item[] Answer: \answerNA{}
    \item[] Justification: LLMs are not an important, original, or non-standard component of the core method.

\end{enumerate}

\end{document}